\documentclass{ieeetj}

\usepackage{cite}
\usepackage{amsmath,amssymb,amsfonts}
\usepackage{algorithmic}
\usepackage{graphicx,color}
\usepackage{textcomp}
\usepackage{xcolor}
\usepackage[hidelinks]{hyperref}
\usepackage{algorithm,algorithmic}
\def\BibTeX{{\rm B\kern-.05em{\sc i\kern-.025em b}\kern-.08em
    T\kern-.1667em\lower.7ex\hbox{E}\kern-.125emX}}
\AtBeginDocument{\definecolor{tmlcncolor}{cmyk}{0.93,0.59,0.15,0.02}\definecolor{NavyBlue}{RGB}{0,86,125}}

\def\OJlogo{\vspace{-4pt}$<$Society logo(s) and publication title will appear here.$>$}
\def\seclogo{\vspace{10pt}$<$Society logo(s) and publication title will appear here.$>$}

\makeatletter
\def\@affil{%
\ifnum\affcount>\z@%
   \tempcount=\z@%
   \loop%
   \ifnum\affcount>\z@%
     \advance\tempcount\@ne%
         {\afffont\csname affil\romannumeral\the\tempcount\endcsname\par}%
    \advance\affcount\m@ne%
  \repeat%
\fi%
}
\makeatother
\usepackage{lipsum}

\usepackage{xcolor}

\usepackage[nolist,nohyperlinks]{acronym}
\acrodef{fmcw}[FMCW]{Frequency Modulated Continuous Wave}
\acrodef{fft}[FFT]{Fast Fourier Transform}
\acrodef{if}[IF]{Intermediate Frequency}
\acrodef{aoa}[AoA]{Angle(s) of Arrival}
\acrodef{imu}[IMU]{Inertial Measurement Unit}
\acrodef{fov}[FoV]{Field of View}
\acrodef{aop}[AOP]{Antenna on Package}
\acrodef{ins}[INS]{Inertial Navigation System}
\acrodef{ains}[AINS]{Aided Inertial Navigation System}
\acrodef{cbf}[CBF]{Control Barrier Function}
\acrodef{swap}[SWaP]{Size, Weight, and Power}
\acrodef{uav}[UAV]{Uncrewed Aerial Vehicles}
\acrodef{soc}[SoC]{System on a Chip}
\acrodef{ekf}[EKF]{Extended Kalman Filter}
\acrodef{ukf}[UKF]{Unscented Kalman Filter}
\acrodef{msckf}[MSCKF]{Multi-State Constraint Kalman Filter}
\acrodef{rmse}[RMSE]{Root Mean Squared Error}
\acrodef{ape}[APE]{Absolute Pose Error}
\acrodef{rpe}[RPE]{Relative Pose Error}
\acrodef{adc}[ADC]{Analog-to-Digital Converter}

\acrodef{gnss}[GNSS]{Global Navigation Satellite System}
\acrodef{ransac}[RANSAC]{Random Sample Consensus}
\acrodef{slam}[SLAM]{Simultaneous Localization and Mapping}
\acrodef{ptp}[PTP]{Precision Time Protocol}
\acrodef{nmea}[NMEA]{National Marine Electronics Association}

\acrodef{iekf}[IEKF]{Iterated Extended Kalman Filter}
\acrodef{gnc}[GNC]{Graduated Non-Convexity}

\acrodef{snr}[SNR]{Signal-to-Noise Ratio}

\ifCLASSOPTIONcompsoc
    \usepackage[caption=false, font=normalsize, labelfont=sf, textfont=sf]{subfig}
\else
\usepackage[caption=false, font=footnotesize]{subfig}
\fi

\usepackage[capitalise]{cleveref}

\usepackage{bm}

\usepackage{multirow}
\usepackage{makecell}
\usepackage{booktabs}

\usepackage[T1]{fontenc}
\usepackage{siunitx}
\DeclareSIUnit\bins{bins}
\DeclareSIUnit{\sqrthertz}{\sqrt{\si{\hertz}}}

\usepackage[super]{nth}

\newcommand{\mat}[1]{\mathbf{#1}}
\newcommand{\rot}[3]{\mat{#1}_\mathtt{#2}^\mathtt{#3}}
\newcommand{\scalar}[3]{{#1}_\mathtt{#2}^\mathtt{#3}}
\renewcommand{\vec}[3]{\smash[t]{\bm{#1}}_\mathtt{#2}^\mathtt{#3}}
\newcommand{\coord}[1]{$\{\mathtt{#1}\}$}

\def\hlfon{1} % 1/0 for highlight on/off
\def\hlf#1{%
  \ifnum\hlfon=1
    \textcolor{blue}{#1}%
  \else
    \textcolor{black}{#1}%
  \fi
}
\def\todo#1{%
  \ifnum\hlfon=1
    \textcolor{red}{??#1}%
  \else
    #1%
  \fi
}

\newcommand{\spacingFigureCaption}{\vspace{-1.5em}}

\newcommand{\corridorBase}{Corridor}
\newcommand{\gymBase}{Gym}
\newcommand{\mineBase}{Mine}
\newcommand{\forestBase}{Forest}
\newcommand{\basementBase}{Basement}

\newcommand{\corridor}{\emph{\corridorBase{}}}
\newcommand{\gym}{\emph{\gymBase{}}}
\newcommand{\mine}{\emph{\mineBase{}}}
\newcommand{\forest}{\emph{\forestBase{}}}
\newcommand{\basement}{\emph{\basementBase{}}}

\newcommand{\Corridor}{\textbf{\corridorBase{}}}
\newcommand{\Gym}{\textbf{\gymBase{}}}
\newcommand{\Mine}{\textbf{\mineBase{}}}
\newcommand{\Forest}{\textbf{\forestBase{}}}
\newcommand{\Basement}{\textbf{\basementBase{}}}

\newcommand{\doer}{\texttt{x-RIO}}
\newcommand{\asl}{\texttt{ASL RIO}}

\newcommand{\rc}[1]{{RC#1}}

\newcommand{\normal}[2]{\mathcal{N} \left( #1 , #2 \right)}
\newcommand{\uniform}[2]{\mathcal{U} \left( #1 , #2 \right)}
\newcommand{\noise}[1]{{n}_{#1}}
\newcommand{\Noise}[1]{\bm{n}_{\bm{#1}}}
\newcommand{\jacobian}[2]{\frac{\partial#1}{\partial#2}}

\newcommand{\E}[1]{\mathbb{E} \left[ #1 \right]}

\newcommand{\experiment}[1]{\texttt{#1}}
\newcommand{\config}[1]{\texttt{#1}}

\usepackage[nonumberlist,sort=use,automake,nogroupskip,style=alttree]{glossaries}

\makeglossaries
\glssetwidest{$\normal{\bm{x}}{\Sigma}$}
\newglossaryentry{iframe}{
    name={$\mathtt{I}$}, 
    description={Inertial frame}}
\newglossaryentry{bframe}{
    name={$\mathtt{B}$}, 
    description={\ac{imu}-aligned body frame}}
\newglossaryentry{rframe}{
    name={$\mathtt{R}$}, 
    description={Radar frame}}

\newglossaryentry{hat_notation}{
    name={$\hat{x}$}, 
    description={Estimate of $x$}}
\newglossaryentry{tilde_notation}{
    name={$\tilde{x}$}, 
    description={Measurement of $x$}}

\newglossaryentry{start_freq}{
    name={$f_c$}, 
    description={Chirp start frequency}}
\newglossaryentry{chirp_slope}{
    name={$S$}, 
    description={Chirp slope}}
\newglossaryentry{chirp_duration}{
    name={$T_c$}, 
    description={Chirp duration}}
\newglossaryentry{inst_freq}{
    name={$f(t)$},
    description={Instantaneous chirp frequecy}}
\newglossaryentry{num_chirps}{
    name={$N_c$}, 
    description={Number of chirps per frame}}
\newglossaryentry{beat_frequency}{
    name={$f_b$}, 
    description={Beat frequency from mixing transmitted and received chirps}}
\newglossaryentry{range}{
    name={$d$}, 
    description={Radar target range}}
\newglossaryentry{speed_light}{
    name={$c$}, 
    description={Speed of light}}
\newglossaryentry{range_resolution}{
    name={$\delta d$}, 
    description={Range resolution}}
\newglossaryentry{max_range}{
    name={$\max d$}, 
    description={Maximum range}}
\newglossaryentry{max_beat_freq}{
    name={$\max f_b$}, 
    description={Maximum beat frequency, limited by the \ac{adc} sampling rate}}
\newglossaryentry{doppler}{
    name={$v_r$}, 
    description={Radar target radial speed}}
\newglossaryentry{phase_shift}{
    name={$\Delta\phi$},
    description={Phase shift between \acp{if} from successive chirps}}
\newglossaryentry{wavelength}{
    name={$\lambda$}, 
    description={Mid-chirp wavelength}}
\newglossaryentry{doppler_resolution}{
    name={$\delta v_r$}, 
    description={Doppler resolution}}
\newglossaryentry{max_velocity}{
    name={$\max v_r$}, 
    description={Maximum Doppler}}
\newglossaryentry{phase_horizontal}{
    name={$w_y$}, 
    description={Horizontal phase shift from \ac{aoa} estimation}}
\newglossaryentry{phase_vertical}{
    name={$w_z$}, 
    description={Vertical phase shift from \ac{aoa} estimation}}
\newglossaryentry{azimuth}{
    name={$\theta$}, 
    description={Radar target azimuth angle}}
\newglossaryentry{elevation}{
    name={$\phi$}, 
    description={Radar target elevation angle}}

\newglossaryentry{uniform_dist}{
    name={$\uniform{a}{b}$}, 
    description={Uniform distribution over the interval $[a,b]$}}

\newglossaryentry{velocity_radar}{
    name={$\vec{v}{IR}{R}$}, 
    description={Linear velocity of the radar frame with respect to the inertial frame, expressed in the radar frame}}
\newglossaryentry{bearing_vector}{
    name={$\bm{\mu}$}, 
    description={Bearing unit vector to the radar target}}
\newglossaryentry{target_position}{
    name={$\bm{t}$}, 
    description={Radar target position}}
\newglossaryentry{range_measured}{
    name={$\tilde{d}$}, 
    description={Radar target range measurement}}
\newglossaryentry{range_noise}{
    name={$\noise{d}$}, 
    description={Range measurement noise}}
\newglossaryentry{range_bin_width}{
    name={$l_d$}, 
    description={Range \ac{fft} bin width}}

\newglossaryentry{expected_value}{
    name={$\mathbb{E}[\cdot]$}, 
    description={Expected value operator}}

\newglossaryentry{range_variance}{
    name={$\sigma_{\noise{d}}^2$}, 
    description={Range measurement noise variance}}

\newglossaryentry{doppler_measurement}{
    name={$\tilde{v}_r$}, 
    description={Radar target radial speed measurement}}
\newglossaryentry{doppler_noise}{
    name={$\noise{v_{r}}$}, 
    description={Radial speed measurement noise}}
\newglossaryentry{doppler_bin_width}{
    name={$l_{v_{r}}$}, 
    description={Radial speed \ac{fft} bin width}}
\newglossaryentry{doppler_variance}{
    name={$\sigma_{\noise{v_{r}}}^2$}, 
    description={Radial speed measurement noise variance}}
\newglossaryentry{phase_horiz_measurement}{
    name={$\tilde{w}_{y}$},
    description={Horizontal phase measurement}}
\newglossaryentry{phase_vert_measurement}{
    name={$\tilde{w}_{z}$},
    description={Vertical phase measurement}}
\newglossaryentry{phase_horiz_noise}{
    name={$\noise{w_y}$}, 
    description={Horizontal phase measurement noise}}
\newglossaryentry{noise_phase_vert}{
    name={$\noise{w_z}$}, 
    description={Vertical phase measurement noise}}
\newglossaryentry{phase_horiz_bin_width}{
    name={$l_{w_{y}}$}, 
    description={Horizontal phase \ac{fft} bin width}}
\newglossaryentry{phase_vert_bin_width}{
    name={$l_{w_{z}}$}, 
    description={Vertical phase \ac{fft} bin width}}
\newglossaryentry{phase_vector}{
    name={$\bm{w}$}, 
    description={\ac{aoa} phase vector $\begin{bmatrix} w_y & w_z \end{bmatrix}^\top$}}
\newglossaryentry{noise_phase_vector}{
    name={$\Noise{w}$}, 
    description={\ac{aoa} phase vector measurement noise}}
\newglossaryentry{phase_vector_variance}{
    name={$\Sigma_{\Noise{w}}$}, 
    description={\ac{aoa} phase vector measurement noise covariance}}
\newglossaryentry{bearing_measured}{
    name={$\bm{\tilde{\mu}}$}, 
    description={Bearing vector measurement}}
\newglossaryentry{noise_bearing}{
    name={$\Noise{\mu}$}, 
    description={Bearing vector measurement noise}}
\newglossaryentry{bearing_variance}{
    name={$\Sigma_{\Noise{\mu}}$}, 
    description={Bearing vector measurement noise covariance}}

\newglossaryentry{normal_dist}{
    name={$\normal{\bm{x}}{\Sigma}$}, 
    description={Gaussian distribution with mean $\bm{x}$ and covariance $\Sigma$}}

\newglossaryentry{rotation_body_inertial}{
    name={$\rot{R}{B}{I}$}, 
    description={Rotation matrix aligning the body frame to the inertial frame}}
\newglossaryentry{position_body}{
    name={$\vec{p}{IB}{I}$}, 
    description={Position of the body frame origin with respect to the inertial frame, expressed in the inertial frame}}
\newglossaryentry{velocity_body}{
    name={$\vec{v}{IB}{I}$}, 
    description={Linear velocity of the body frame with respect to the inertial frame, expressed in the inertial frame}}
\newglossaryentry{angrate_true}{
    name={$\vec{\omega}{IB}{B}$}, 
    description={Angular rate of the body frame with respect to the inertial frame, expressed in the body frame}}
\newglossaryentry{force_true}{
    name={$\vec{f}{IB}{B}$}, 
    description={Specific force of the body frame with respect to the inertial frame, expressed in the body frame}}
\newglossaryentry{angrate_meas}{
    name={$\vec{\tilde{\omega}}{IB}{B}$}, 
    description={Measured angular rate from the \ac{imu}}}
\newglossaryentry{force_meas}{
    name={$\vec{\tilde{f}}{IB}{B}$}, 
    description={Measured specific force from the \ac{imu}}}
\newglossaryentry{noise_gyro}{
    name={$\Noise{\omega}$}, 
    description={Gyroscope measurement noise}}
\newglossaryentry{noise_accel}{
    name={$\Noise{f}$}, 
    description={Accelerometer measurement noise}}
\newglossaryentry{covar_gyro}{
    name={$\Sigma_{\Noise{\omega}}$}, 
    description={Gyroscope measurement noise covariance}}
\newglossaryentry{covar_accel}{
    name={$\Sigma_{\Noise{f}}$}, 
    description={Accelerometer measurement noise covariance}}
\newglossaryentry{bias_gyro}{
    name={$\vec{b}{g}{}$}, 
    description={Gyroscope bias}}
\newglossaryentry{bias_accel}{
    name={$\vec{b}{a}{}$}, 
    description={Accelerometer bias}}
\newglossaryentry{bias_baro}{
    name={$\scalar{b}{b}{}$}, 
    description={Barometer bias}}
\newglossaryentry{rotation_radar_body}{
    name={$\rot{R}{R}{B}$}, 
    description={Rotation matrix aligning the radar frame to the body frame}}
\newglossaryentry{position_radar_body}{
    name={$\vec{l}{BR}{B}$}, 
    description={Lever arm from the body frame to the radar frame, expressed in the body frame}}
\newglossaryentry{state_total}{
    name={$\mathbf{x}$}, 
    description={Total state space}}
\newglossaryentry{state_tv}{
    name={$\mathbf{x}_\mathtt{TV}$}, 
    description={Time-varying states}}
\newglossaryentry{state_ti}{
    name={$\mathbf{x}_\mathtt{TI}$}, 
    description={Time-invariant states}}

\newglossaryentry{skew_symmetric}{
    name={$(\cdot)^\times$}, 
    description={Skew-symmetric matrix operator}}

\newglossaryentry{gravity}{
    name={$\vec{g}{}{I}$}, 
    description={Gravitational acceleration expressed in the inertial frame}}

\newglossaryentry{noise_accel_bias}{
    name={$\Noise{a}$}, 
    description={Accelerometer bias process noise}}
\newglossaryentry{noise_gyro_bias}{
    name={$\Noise{g}$}, 
    description={Gyroscope bias process noise}}
\newglossaryentry{noise_baro_bias}{
    name={$\noise{b}$}, 
    description={Barometer bias process noise}}
\newglossaryentry{covar_accel_bias}{
    name={$\Sigma_{\Noise{a}}$}, 
    description={Accelerometer bias process noise covariance}}
\newglossaryentry{covar_gyro_bias}{
    name={$\Sigma_{\Noise{g}}$}, 
    description={Gyroscope bias process noise covariance}}
\newglossaryentry{covar_baro_bias}{
    name={$\sigma_{\noise{b}}^2$}, 
    description={Barometer bias process noise covariance}}

\newglossaryentry{time}{
    name={$t_{k}$},
    description={Time at $k$}}
\newglossaryentry{smoother_lag}{
    name={$\ell+1$}, 
    description={Smoother window lag duration}}
\newglossaryentry{state_history}{
    name={$\mathcal{X}_{k-\ell:k}$}, 
    description={State history over smoother window from time $k-\ell$ to $k$}}
\newglossaryentry{measurement_imu_set}{
    name={$\mathcal{I}_{i,j}$}, 
    description={Set of \ac{imu} measurements between times $t_i$ and $t_j$}}
\newglossaryentry{measurement_radar}{
    name={$\mathcal{R}_i$}, 
    description={Radar point cloud measurement at time $t_i$}}
\newglossaryentry{measurement_barometer}{
    name={$\mathcal{B}_i$}, 
    description={Barometer measurement at time $t_i$}}
\newglossaryentry{residual_imu}{
    name={$\bm{e}_{\mathcal{I}_{i,j}}$}, 
    description={\ac{imu} preintegration factor residual}}
\newglossaryentry{covar_imu}{
    name={$\Sigma_{\mathcal{I}_{i,j}}$}, 
    description={\ac{imu} preintegration factor residual covariance}}
\newglossaryentry{num_radar_targets}{
    name={$N_{\mathcal{R}_i}$}, 
    description={Number of targets in the radar point cloud measurement at time $t_i$}}
\newglossaryentry{radar_point}{
    name={$\tau$},
    description={Point from the radar point cloud measurement}}
\newglossaryentry{residual_doppler}{
    name={$e_\mathcal{D_{\tau}}$}, 
    description={Doppler factor residual for radar point $\tau$}}
\newglossaryentry{covar_doppler}{
    name={$\sigma_{\mathcal{D}_{\tau}}^2$}, 
    description={Doppler factor residual variance for radar point $\tau$}}
\newglossaryentry{residual_geometry}{
    name={$\bm{e}_\mathcal{G_{\tau}}$}, 
    description={Geometry factor residual for radar point $\tau$}}
\newglossaryentry{covar_geometry}{
    name={$\Sigma_{\mathcal{G}_{\tau}}$}, 
    description={Geometry factor residual covariance for radar point $\tau$}}
\newglossaryentry{residual_barometry}{
    name={$e_{\mathcal{B}_{i}}$}, 
    description={Barometry factor residual}}
\newglossaryentry{covar_barometry}{
    name={$\sigma_{\mathcal{B}}^2$}, 
    description={Barometry factor residual variance}}
\newglossaryentry{state_history_optimal}{
    name={$\mathcal{X}_{k-\ell:k}^*$}, 
    description={Optimal state history estimate over smoother window}}
\newglossaryentry{margin_residual}{
    name={$\bm{e}_{0}$}, 
    description={Marginalization prior factor residual}}
\newglossaryentry{margin_covar}{
    name={$\Sigma_{0}$}, 
    description={Marginalization prior factor residual covariance}}
\newglossaryentry{influence_barometry}{
    name={$\rho_{\mathcal{B}}$}, 
    description={Huber M-estimator influence function applied to the barometry factor residual}}
\newglossaryentry{influence_doppler}{
    name={$\rho_{\mathcal{D}}$}, 
    description={Cauchy M-estimator influence function applied to the Doppler factor residual}}

\newglossaryentry{residual_imu_rotation}{
    name={$\bm{e}_{\mathcal{I}_{\mathbf{R}_{i,j}}}$}, 
    description={\ac{imu} preintegration factor rotation residual}}
\newglossaryentry{residual_imu_position}{
    name={$\bm{e}_{\mathcal{I}_{\bm{p}_{i,j}}}$}, 
    description={\ac{imu} preintegration factor position residual}}
\newglossaryentry{residual_imu_velocity}{
    name={$\bm{e}_{\mathcal{I}_{\bm{v}_{i,j}}}$}, 
    description={\ac{imu} preintegration factor velocity residual}}
\newglossaryentry{angrate_avg}{
    name={$\vec{\bar{\omega}}{IB}{B}$}, 
    description={Average angular rate during the chirping period}}
\newglossaryentry{angrate_avg_meas}{
    name={$\vec{\tilde{\bar{\omega}}}{IB}{B}$}, 
    description={Average angular rate calculated with measurements from the \ac{imu}}}
\newglossaryentry{num_angrate_samples}{
    name={$N_{\vec{\bar{\omega}}{IB}{B}}$}, 
    description={Number of noisy angular rate measurements used to compute the average}}
\newglossaryentry{noise_angrate_avg}{
    name={$\Noise{\bar{\omega}}$}, 
    description={Average angular rate measurement noise}}
\newglossaryentry{velocity_radar_est}{
    name={$\vec{\hat{v}}{IR}{R}$}, 
    description={Estimated linear velocity of the radar frame with respect to the inertial frame, expressed in the radar frame}}
\newglossaryentry{rotation_body_inertial_est}{
    name={$\rot{\hat{R}}{B}{I}$}, 
    description={Estimated rotation matrix aligning the body frame to the inertial frame}}
\newglossaryentry{velocity_body_est}{
    name={$\vec{\hat{v}}{IB}{I}$}, 
    description={Estimated linear velocity of the body frame with respect to the inertial frame, expressed in the inertial frame}}
\newglossaryentry{gyro_bias_est}{
    name={$\vec{\hat{b}}{g}{}$},
    description={Estimated gyroscope bias}}
\newglossaryentry{rotation_radar_body_est}{
    name={$\rot{\hat{R}}{R}{B}$}, 
    description={Estimated rotation matrix aligning the radar frame to the body frame}}
\newglossaryentry{position_radar_body_est}{
    name={$\vec{\hat{l}}{BR}{B}$}, 
    description={Estimated lever arm from the body frame to the radar frame, expressed in the body frame}}
\newglossaryentry{static_threshold}{
    name={$\kappa_\text{static}$}, 
    description={Threshold for static point classification, based on the whitened Doppler residual}}
\newglossaryentry{search_radius}{
    name={$\rho$}, 
    description={KD tree neighbor search radius for geometry factor}}
\newglossaryentry{centroid_mean}{
    name={$\vec{\tilde{q}}{}{}$}, 
    description={Mean of nearest neighbors from the KD tree}}
\newglossaryentry{centroid_covar}{
    name={$\Sigma_{\Noise{q}}$}, 
    description={Covariance of the nearest neighbors from the KD tree}}
\newglossaryentry{target_position_measured}{
    name={$\bm{\tilde{t}}$}, 
    description={Radar target position measurement}}
\newglossaryentry{position_body_est}{
    name={$\vec{\hat{p}}{IB}{I}$}, 
    description={Estimated position of the body frame origin with respect to the inertial frame, expressed in the inertial frame}}
\newglossaryentry{barometric_pressure}{
    name={$P$}, 
    description={Barometric pressure}}
\newglossaryentry{altitude_function}{
    name={$h(P)$}, 
    description={Function for calculating altitude from barometric pressure}}
\newglossaryentry{standard_temp}{
    name={$T_0$}, 
    description={Standard temperature}}
\newglossaryentry{temp_lapse_rate}{
    name={$L_0$}, 
    description={Temperature lapse rate}}
\newglossaryentry{standard_pressure}{
    name={$P_0$}, 
    description={Standard pressure}}
\newglossaryentry{gas_constant}{
    name={$R$}, 
    description={Gas constant}}
\newglossaryentry{gravity_mag}{
    name={$g$}, 
    description={Magnitude of acceleration due to gravity}}
\newglossaryentry{molar_mass_air}{
    name={$M$}, 
    description={Average molar mass of air at sea level}}
\newglossaryentry{barometric_pressure_meas}{
    name={$\tilde{P}$}, 
    description={Barometric pressure measurement}}
\newglossaryentry{height_est}{
    name={$\hat{z}$},
    description={Vertical position estimate}}
\newglossaryentry{bias_baro_est}{
    name={$\scalar{\hat{b}}{b}{}$}, 
    description={Estimated barometer bias}}

\newglossaryentry{yaw}{
    name={$\psi_{\text{sp}}$}, 
    description={Yaw angle setpoint}}
\newglossaryentry{yaw_rate}{
    name={$\dot{\psi}_{\text{sp}}$}, 
    description={Yaw-rate setpoint}}
\newglossaryentry{setpoint_x}{
    name={$x_{\text{sp}}$}, 
    description={Trajectory position setpoint in x direction}}
\newglossaryentry{setpoint_y}{
    name={$y_{\text{sp}}$}, 
    description={Trajectory position setpoint in y direction}}

\begin{document}

\title{On the Characterization and Limits of 4D Radar for Aided Inertial Navigation}

\receiveddate{XX Month, XXXX}
\reviseddate{XX Month, XXXX}
\accepteddate{XX Month, XXXX}
\publisheddate{XX Month, XXXX}
\currentdate{XX Month, XXXX}
\doiinfo{XXXX.2022.1234567}

\markboth{On the Characterization and Limits of 4D Radar for Aided Inertial Navigation}{Nissov {et al.}}
\author{Morten Nissov and Kostas Alexis}
\affil{Department of Engineering Cybernetics, Norwegian University of Science and Technology, Trondheim, Norway}
\corresp{Corresponding author: Morten Nissov (email: \texttt{morten.nissov@ntnu.no}).}
\authornote{This material was supported by the Research Council of Norway Award under project SENTIENT (NO-321435) and by the European Commission under Horizon Europe Grants SPEAR (EC 101119774) and SYNERGISE (EC 101121321).}

\begin{abstract}
\ac{fmcw} radar is a promising sensor for aided inertial navigation, due to its robustness in environments that challenge traditional alternatives, such as LiDAR and vision. However, its widespread adoption is hindered by complex, noisy measurements, which make reliable estimation difficult. This manuscript addresses these challenges by analyzing the fundamental measurement relations of \ac{fmcw} radar sensing and developing a reliable estimator. Noise models are derived by applying first principles to the underlying signal processing of a typical radar sensor. These models guide the design of a factor graph-based estimator, utilizing a first-order approximation for the measurement noise propagation. The approach is first examined through simulation, evaluating the significance of different noise sources, the validity of the first-order approximation, and the state-dependent nature of the covariance expressions. Extensive experiments demonstrate the superior robustness and accuracy of the proposed method across diverse field environments and flight profiles, including beyond the radar's standard operating range. Furthermore, the experiments confirm the insights from the simulation regarding the behavior and performance of different estimator configurations relative to their operating conditions. The evaluation data and estimator implementation are made available at \url{https://github.com/ntnu-arl/rig}.
\end{abstract}

\maketitle

\section{INTRODUCTION}
\IEEEPARstart{R}{eliable} estimation is necessary for robotic systems. Resulting from decades of research, modern state estimation methods perform well in a variety of conditions, including those with limited \ac{gnss} connectivity. However, particular environments can still expose the system's frailty, resulting in severe drift or even catastrophic failure \cite{cadena2016future,ebadi2024present}. Although LiDAR- and vision-based methods have become the de facto standard for aspiring autonomous platforms that require versatile state estimation, they demonstrate a lack of robustness in certain environments and share common failure cases~\cite{ebadi2024present}. For example, vision-based methods can struggle with changing lighting conditions (an effect ubiquitous in both natural and man-made environments), as well as motion blur. LiDAR-based methods, on the other hand, are susceptible to failure in environments with repeating geometries, such as tunnels. Meanwhile, both modalities face difficulties with airborne particulates (such as fog, dust, or rain). One can attempt to consider such scenarios as edge cases, but for platforms operating in anything but the most controlled environments, this is unrealistic.
Commonplace environments with challenging conditions include those with repeating geometries (such as tunnels \cite{nissov2024degradation}, silos \cite{hatleskog2024Probabilistic}, and fields \cite{nissov2024robust,tuna2025Informed}), adverse visual conditions like lack of visual texture or darkness (e.g., tunnels, mines, and other man-made environments \cite{khattak2020Thermal,tranzatto2022Cerberus}), or potential for obscurants (e.g., natural environments with dust, smoke, or fog \cite{tranzatto2022Cerberus,nissov2024degradation}).
Thus, robustness and the ability to operate anywhere and anytime are core necessities that modern robotic systems lack, prohibiting the more widespread adoption of the technology.

Motivated by these challenges, the research community has turned its attention to \si{\milli\meter}Wave radar systems~\cite{harlow2024new}. A key advantage of this sensing technology is the longer wavelength (generally \SIrange{1}{10}{\milli\meter}), which provides robustness against obscurants, one of the aforementioned challenges for LiDAR- and vision-based systems. Furthermore, \si{\milli\meter}Wave radars perceive directly relative velocity through the Doppler shift. This enables aided inertial navigation without feature matching or associations, eliminating a significant potential failure case as well as sensitivity to particular environmental geometry or a lack of texture. Furthermore, direct velocity sensing makes a compelling case for multi-modal fusion methods, where it has been shown that \si{\milli\meter}Wave radar can help anchor the state estimation,  even when other sensors fail~\cite{nissov2024degradation,nissov2024robust,noh2025garlio}. Despite these clear advantages for robust navigation, the technology has yet to achieve the same mass adoption as vision and LiDAR in the context of state estimation.

\IEEEpubidadjcol

A persistent hurdle for this adoption is the sparse and noisy nature of the point cloud measurements from \si{\milli\meter}Wave radars~\cite{harlow2024new}. Especially so for the most size-constrained sensors, as the angular resolution necessarily worsens with fewer antennas. Moreover, the measurement noise is often complex and inconsistent, with characteristics that vary as a function of the ego-motion. Existing works that consider more sophisticated noise modeling~\cite{thormann2023bearing,xu2025radarPointUncertainty,zhu2025Robustradar} do so without thorough evaluation of the impact, and without a clear connection to the actual nonlinear measurement relationships.

Motivated by the above, we propose noise models for the radar measurements of range, Doppler, and \ac{aoa}, develop an estimator based on these models, and analyze the behavior and performance of the models and estimator in simulation and with extensive flight experiments. The noise models are derived from first principles applied to the underlying on-chip signal processing, leading to principled parametrization for noise terms. Then, a factor graph-based aided \ac{ins} is proposed through the linearization of the nonlinear measurement relationships. 
The characteristics of the measurements and their relationship to estimator states (e.g., velocity) are analyzed in simulation. The behavior of the proposed estimator is further validated through robotic experiments across different field environments.
Thus, our contributions include:
\begin{itemize}
    \item Proposing a method which considers the bearing vector as a measurement with noise, corresponding to how the measurement is produced by a typical \ac{fmcw} radar sensor's beamforming.
    \item Investigating the nonlinearities present in the Doppler residual equation used for radar-inertial navigation and their impact with respect to the estimator.
    \item Analysis of the measurement behavior across different chirp configurations and ego-motion with respect to the noise properties and measurement aliasing.
    \item Extensive evaluation of the method across diverse field environments.
    \item Open-sourcing both the implementation and evaluation data, which can be found at \url{https://github.com/ntnu-arl/rig}.
\end{itemize}

The manuscript is organized as follows. \Cref{sec:method:radar} introduces the basics of \ac{fmcw} radar sensing, \cref{sec:method:noise} derives the relevant noise models based on standard \ac{fmcw} signal processing principles, and \cref{sec:method:navigation} formulates the aided \ac{ins} utilizing these models. \Cref{sec:evaluation:noise} evaluates the validity and significance of the modeling with simulation results, \cref{sec:evaluation:validation} evaluates the proposed noise models with empirical data, and \cref{sec:evaluation:flight} evaluates the proposed method with field experiments. Finally, \cref{sec:evaluation:summary} summarizes the evaluation and \cref{sec:conclusion} concludes the paper.
\section{RELATED WORK}\label{sec:related_work}
\ac{fmcw} radar sensing for estimation purposes can be broadly divided into two classes: spinning and \ac{soc} radars. 
The spinning radars used for estimation typically return 2D images and, historically, have not measured Doppler; however, recent updates are making this possible~\cite{lisus2025spinningDoppler}. As a result, approaches are generally utilizing feature extraction~\cite{cen2018Precise,cen2019Matching,lim2023orora} and/or keyframing~\cite{burnett2025RadarLidar,adolfsson2023radar}, following similar approaches popularized by visual-odometry. However, the size, weight, and \ac{fov} of such sensors can be prohibitive for aerial systems. For this reason, we focus on \ac{soc}-type (also referred to as 4D) radars, which have been more successfully miniaturized with respect to \ac{swap} constraints. 4D radars have a narrower azimuth \ac{fov} compared to their spinning counterparts, but compensate with elevation coverage and Doppler measuring capabilities, returning point cloud measurements instead of 2D images. This sensing technology has proven valuable in its role as an aiding sensor for inertial navigation, demonstrated by seminal works such as~\cite{kellner2013instantaneous,kellner2014dual}, where the authors motivate a least squares linear velocity solution from the radar point cloud for estimation in an automotive context.

Following on least-squares solutions for linear velocity from radar point clouds, many works have contributed to advancing the field of radar-aided inertial navigation with \ac{soc} radars. These can be roughly divided by the choice of estimation machinery, with most approaches utilizing \acp{ekf} or factor graphs.
The authors in~\cite{doer2020ekf} adopt a standard error-state Kalman filter framework for velocity-aided inertial navigation, using \ac{ransac} to increase the robustness of the 3D least squares solution for linear velocity. Furthermore, in~\cite{doer2021yaw} the authors propose methods for reducing yaw drift through radar point-derived yaw measurements, guided by Manhattan world assumptions. This work was extended to function with multiple radar sensors simultaneously in~\cite{doer2021xrio}.
In~\cite{zhuang2023iriom}, the authors move away from \ac{ransac} and instead adopt \ac{gnc} to estimate linear velocity. This linear velocity estimate is combined with distribution-to-distribution scan-to-submap matches for the \ac{iekf} measurement update. Furthermore, place revisits are detected using~\cite{kim2018scancontext} and integrated into a separate pose graph optimization.
The authors of~\cite{li2023preint} develop a radar-only \ac{slam} method, where point cloud refinement is accomplished by extracting the ground plane and filtering by comparing successive radar point cloud measurements. Furthermore, in the absence of an \ac{imu}, the authors develop a pre-integration factor derived from the \ac{ransac} least-squares ego-velocity estimate.
Augmenting with \ac{imu}, the authors of~\cite{herraez2025rai} improve upon radar-inertial \ac{slam} by optimizing both local and global graphs, supported by radar ego-velocity estimates, scan matching, and loop detection.
In~\cite{chen2023drio}, the authors propose a method for improving the radar point cloud velocity estimation by jointly detecting ground points and estimating velocity. The velocity estimate is fused with the \ac{imu} for odometry, and the ground point detections are used to refine the point cloud.
The authors of~\cite{wu2024efear} propose DBSCAN clustering on the 2D linear velocity estimates as an alternative to \ac{ransac} for removing dynamic outliers, which is used as a prior for their feature-based scan-to-submap registration.

Although a robust least-squares solution for linear velocity is still used for outlier rejection by many works (e.g., \cite{kramer2024suburban,michalczyk2023tight}), directly fusing radial speed in the estimator has become more common.
In~\cite{kramer2020visually}, the authors demonstrate the capability for velocity estimation using radar Doppler measurement residuals in a factor graph optimizer. The authors go on to propose a matching and mapping methodology for reducing yaw drift with less restrictive assumptions on environmental characteristics in~\cite{kramer2024suburban}, utilizing Doppler-derived velocity estimation in the graph and a separate position update sourced from the map.
The authors in~\cite{michalczyk2022tight} adopt a \ac{msckf}-based~\cite{mourikis2007MSCKF} approach, and include point-to-point associations, range measurements, and Doppler measurements in their error-state Kalman filter. This is expanded upon in~\cite{michalczyk2023tight}, where the online estimation of persistent landmarks is enabled in the estimator.
The authors in~\cite{girod2024radar} adopt a factor-graph based approach and include radial speed factors, barometry, as well as radar point-to-point associations with landmark estimation to facilitate zero-velocity updates. Furthermore, the authors analyze robust cost functions and their role in the context of Doppler aiding.
In~\cite{wang2025dopplerslam}, the authors propose a unified approach, for both \ac{fmcw} radar and LiDAR sensors. This method uses an \ac{iekf} to fuse point-to-plane and Doppler residuals with \ac{imu} for odometry, alongside loop closure detection, which drives the map and extrinsic optimization.
However, these works require point-to-point associations~\cite{michalczyk2022tight,michalczyk2023tight,girod2024radar} or normal estimates~\cite{wang2025dopplerslam}, which can be difficult given the noisy nature of the measurements~\cite{harlow2024new,kubelka2024need}.

Common across all the aforementioned works, studying the role of \ac{fmcw} radar in the context of inertial navigation, is the assumption of an overly simplistic noise model, which considers purely additive noise contributions or noise of equal magnitude for all points in the cloud. 
Addressing the latter, in~\cite{xiang2025vgc} Doppler residuals are weighted by their spatial distribution, in order to equalize the Doppler residual penalty across the point cloud. Furthermore, a histogram-based point-descriptor is proposed and used for adding point-to-point matching residuals from successive radar scans to the radar-inertial windowed smoother.
Alternatively, the authors in~\cite{thormann2023bearing} do consider a potential noise model for the 1D bearing (azimuth only), in an \ac{ukf}-based aided \ac{ins}. The authors utilize the model proposed in~\cite{bordonaro2012unbiased} for debiasing and modeling the statistical impact of the conversion from spherical to Cartesian coordinates.
In~\cite{wang2024riv}, the authors propose a method for radar \ac{slam} which extracts and removes ground from the point cloud measurements, and includes scan-to-scan and scan-to-submap registration factors, with anisotropic noise terms, for odometry and mapping, respectively.
In~\cite{xu2025radarPointUncertainty}, the authors consider Gaussian distributions for the range and the \ac{aoa} (modeled as an element of the 2-sphere $\mathbb{S}^2$) noise, disregarding uncertainty in the Doppler velocity measurement. With this, the authors propose a factor graph-based \ac{slam} estimator that incorporates landmark estimation and probabilistically motivated point associations, with noise propagation facilitated by linearization.
In~\cite{zhu2025Robustradar}, the authors consider the same model for range and \ac{aoa} uncertainty, but also include Doppler measurement noise in the Doppler residuals. This is used for point-to-distribution scan matching against either an online incrementally aggregated map or a prior map, if available.
Contributing to the challenge is the inherent complexity of radar sensors, even in the face of environmental conditions they are thought to be impervious to. This is examined in~\cite{chan2025noise} by studying measurement consistency for automotive radars subject to rain and snow. The authors also create an overview of how different phenomena (with respect to assembly, usage, and the environment) can impact the output quality.

While the aforementioned methods do consider the possibility of more complicated noise models, they do not show a complete picture. The modeling in these works is lacking in either dimensionality~\cite{thormann2023bearing} for modern \ac{soc} radars or not related to how measurements are calculated by the radar sensors (e.g., the specific beamforming algorithms)~\cite{xu2025radarPointUncertainty,zhu2025Robustradar}.
Furthermore, by introducing separate noise models for Doppler and \ac{aoa}, as in \cite{xu2025radarPointUncertainty,zhu2025Robustradar} without relating them to the underlying signal processing, the added noise parameters become tuning knobs with limited intuition. Doing so ignores the potential insight that could be gained by investigating the sensor characteristics and signal processing.

To address these limitations, we derive noise models based on the signal processing of an \ac{fmcw} radar sensor and construct an estimator on this basis, leveraging this knowledge of the measurement relations and their resultant properties.

\section{METHOD}\label{sec:method}
This section is organized as follows. First, notation is introduced in \cref{sec:method:notation}. Then, \cref{sec:method:radar} introduces the basics of \ac{fmcw} radar sensing. \Cref{sec:method:noise} develops the noise models for the radar measurements of range, Doppler, and bearing. Finally, \cref{sec:method:navigation} describes the resulting factor graph-based estimator.

\subsection{Notation}\label{sec:method:notation}
The following sections require the use of different coordinate systems and notation to frame the estimation problem. The relevant coordinate frames for this work are the inertial frame \coord{I}, the robot-fixed body frame \coord{B} (assumed to be aligned with the \ac{imu}), and the radar frame \coord{R}. With respect to notation, let the position of \coord{R} with respect to \coord{B}, expressed in \coord{I} be $\vec{p}{BR}{I}\in\mathbb{R}^3$ and similarly the velocity $\vec{v}{BR}{I}\in\mathbb{R}^3$. Let the rotation matrix aligning \coord{B} to \coord{I} be $\rot{R}{B}{I}\in\textit{SO}(3)$. Furthermore, let estimated quantities be denoted with the hat as $\hat{x}$ and measured quantities be denoted with a tilde as $\tilde{x}$.

\subsection{\acs{fmcw} Radar Sensing}\label{sec:method:radar}
This section presents an overview of \ac{fmcw} radar sensing, following \cite{richards2013Fundamentals,iovescu2020fundamentals} as it pertains to noise modeling and the application of aided inertial navigation.

\begin{figure}[h]
    \centering
    \includegraphics[width=\linewidth]{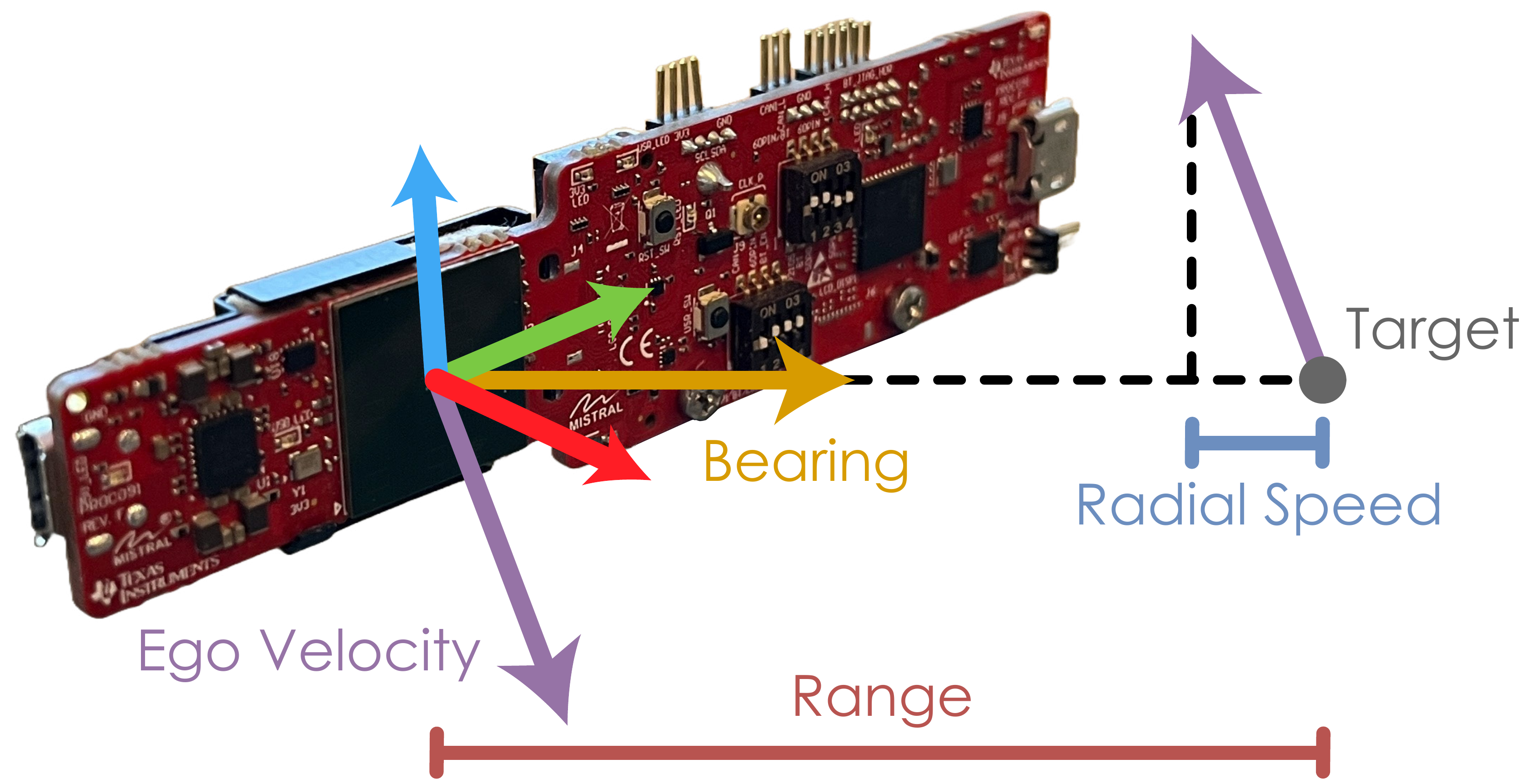}
    \spacingFigureCaption
    \caption{A radar sensor with non-zero ego-velocity measuring the range, Doppler, and bearing (derived from the \ac{aoa}) of a single target in the environment with relative motion arising from the sensor ego-velocity.}
    \label{fig:method:radar}
\end{figure}

\begin{figure}[h]
    \centering
    \includegraphics[width=\linewidth]{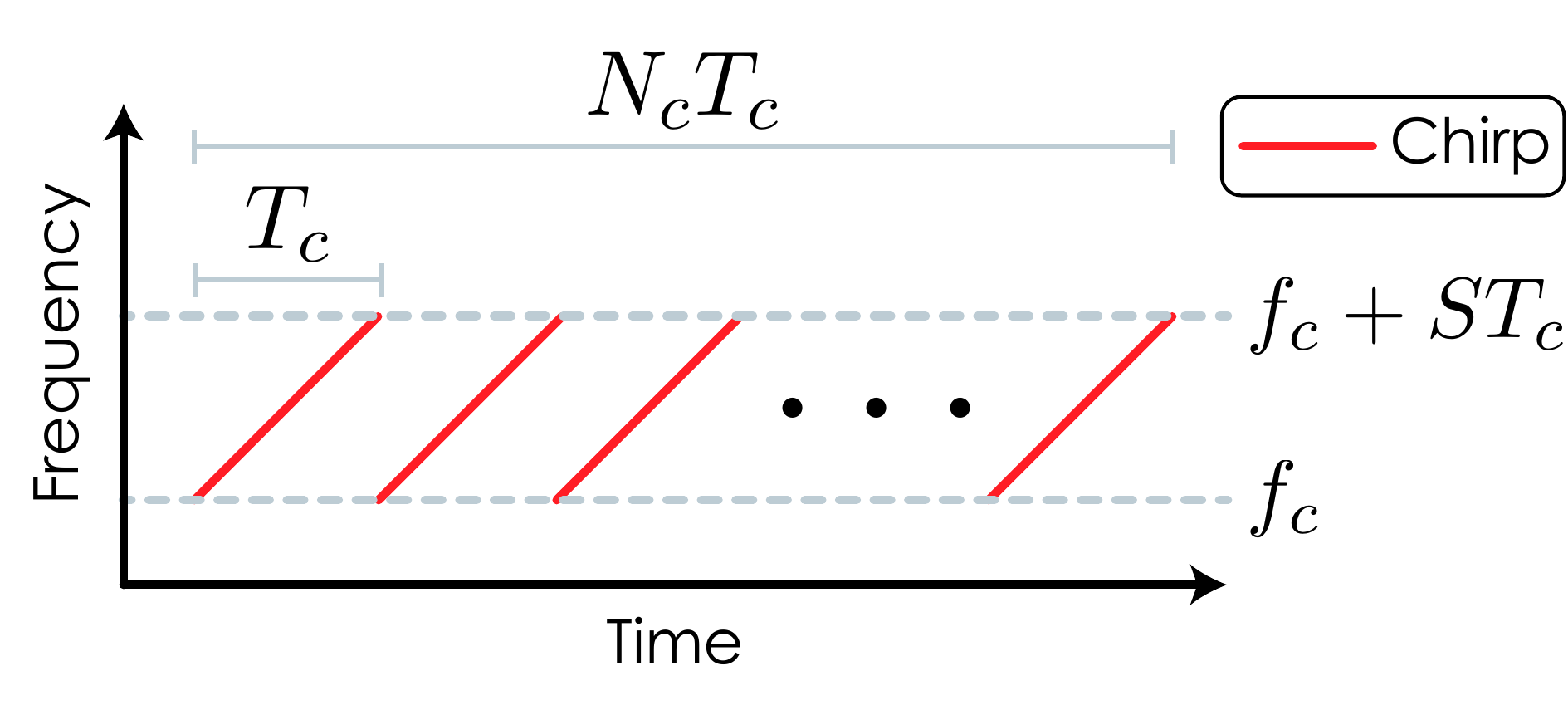}
    \spacingFigureCaption
    \caption{The temporal evolution of an exemplary transmitted chirp frame with respect to its defining characteristics.}
    \label{fig:method:radar:chirp}
\end{figure}

A typical \ac{fmcw} radar sensor returns measurements of range, radial speed (also known as Doppler), and the \ac{aoa} (azimuth and elevation in the 3D case), as seen in the visualization in \cref{fig:method:radar}. 
The aforementioned measurements are taken by transmitting groups of chirps, called frames, and evaluating the resultant echoes reflected by targets in the environment. Different modulation strategies for chirps exist, but a simple and commonly used modulation is that of linearly increasing frequency with respect to time~\cite{richards2013Fundamentals}. Such a chirp is parametrized by the starting frequency $f_c$, chirp slope $S$, and the chirp duration $T_c$ such that the instantaneous chirp frequency $f(t)$ is given by

\begin{equation}
    f(t + \Delta t) = f_c + S \Delta t,\quad \Delta t\in \left[ 0, T_c \right].
\end{equation}
After having transmitted a frame (containing $N_c$ chirps), the received signals are sampled by an \ac{adc} and processed to return measurements. An example of such a chirp frame can be seen in \cref{fig:method:radar:chirp}.

Consider the reflected chirps from a single obstacle placed sufficiently far away from the radar sensor to generate only a single return, i.e., in the radar's far-field. The transmitted and received chirps are mixed, resulting in an \ac{if}. The mixing operation takes as input two sinusoids and outputs a new sinusoid whose instantaneous frequency and phase shift are each equal to the difference of the input signals' instantaneous frequencies and phase shifts~\cite{iovescu2020fundamentals}. The received signal from the single target comes with a time delay with respect to the transmitted signal, and as a result, the \ac{if} signal has a constant beat frequency $f_b$ which is proportional to the time delay and chirp slope, and therefore the distance to the target. Thus, the target range $d$ is

\begin{equation}\label{eq:method:range}
    d = \frac{f_b c}{2 S},
\end{equation}
where $c$ is the speed of light. Multiple targets in the environment will result in multiple beat frequencies in the \ac{if}, which can be resolved using the \ac{fft}.

A key property associated with radar sensors is their ability to differentiate targets with similar properties (range, Doppler, \ac{aoa}). This is referred to as the resolution of the sensor with respect to that property.
For example, the range resolution $\delta d$ is a function of the chirp parameters, such that

\begin{equation}
    \delta d = \frac{c}{2 S T_c}.
\end{equation}
Another key property of radar sensing is that these measurements have maximum quantities, also determined by the chirp configuration or antenna layout. For example, the maximum possible range $\max d$ is limited by the chirp configuration such that

\begin{equation}
    \max d = \frac{\max f_b c}{2S},
\end{equation}
where $\max f_b$ is the maximum beat frequency (dictated by the \ac{adc} sampling rate).
Assume now that the aforementioned target is moving toward the radar sensor with constant radial speed $v_r$. This is correlated with the phase shift $\Delta\phi$ between \acp{if} resulting from successive chirps transmitted and reflected by the same target, following

\begin{equation}\label{eq:method:doppler}
    v_r = \frac{\lambda \Delta\phi}{4 \pi T_c},
\end{equation}
where $\lambda$ is the mid-chirp wavelength.

For environments with multiple targets, the radial speed measurements are calculated by taking an \ac{fft} over the peaks isolated by the range processing in order to separate the phase shifts resulting from targets moving at different speeds. Similarly to the range, there are limitations in the resolution $\delta v_r$ and the maximum possible value $\max v_{r}$, related to the chirp parameters by

\begin{align}
    \delta v_r &= \frac{\lambda}{2 N_{c} T_c},\\
    \max v_r &= \frac{\lambda}{4 T_c}.
\end{align}
The \ac{aoa} can be found in a similar way by instead considering the phase difference across multiple chirps transmitted and received by multiple antennas. This is because the wavefront of the return signal from a single chirp will be received with delay, by spatially distributed antennas, as a function of the \ac{aoa}. 
The horizontal and vertical phase shifts $w_y$, $w_z$ are related to the azimuth and elevation angles $\theta$, $\phi$ following

\begin{align}\label{eq:method:wy}
    w_y &= \pi \sin(\theta)\cos(\phi),\\
    w_z &= \pi \sin(\phi). \label{eq:method:wz}
\end{align}
This relationship, distinct from range and Doppler, shows the desired quantities (azimuth and elevation) as a nonlinear function of the measured quantities (horizontal and vertical phase). This will have a significant impact when modeling the noise properties.

The process of determining the \ac{aoa} is known as beamforming, and in general has more complexity than calculating range or Doppler due to the nonlinearity of \cref{eq:method:wy,eq:method:wz} and the correlation between azimuth and elevation. Although multiple options for beamforming exist \cite{veen1988beamforming}, a relatively inexpensive method \cite{capon1969Adaptive} is to calculate the \ac{fft} across the signals received by separate transmit-receive antenna pairs to resolve the aforementioned phase shifts. This is the algorithm used by the compute-constrained, consumer-grade \si{\milli\meter}Wave radars considered in this work \cite{rao2017tiMIMO,iovescu2020fundamentals}. As with range and Doppler, the angle measurements have resolution and maxima, which here are a function of the antenna spacing rather than the chirp parameters. As such, they are largely uncontrollable save for a complete sensor re-design.

\subsection{Noise Modeling}\label{sec:method:noise}
Motivated by the fundamentals of \ac{fmcw} radar sensing, this section develops noise models rooted in the typical signal processing algorithms used to create radar measurements of range, Doppler, and \ac{aoa}.

\begin{figure}[h]
    \centering
    \includegraphics[width=\linewidth]{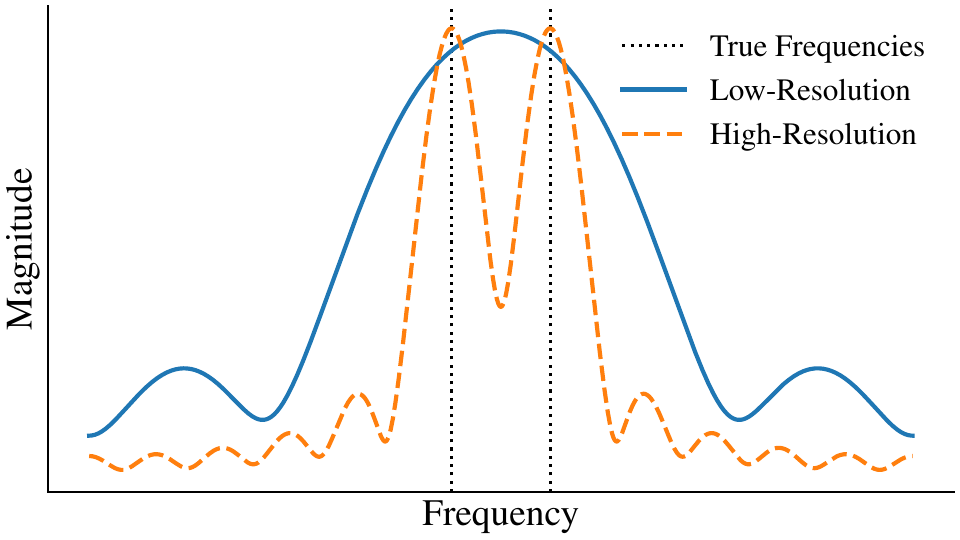}
    \spacingFigureCaption
    \caption{The performance of low- and high-resolution \acp{fft} on a signal with closely spaced frequencies. Note, the low-resolution \ac{fft} is unable to resolve these frequencies.}
    \label{fig:method:fft:resolution}
\end{figure}

All of the measurements introduced in \cref{sec:method:radar} can be calculated by isolating frequencies from a given signal using the Fourier transform. Since these are calculated by digital computers using discrete signal measurements, the pitfalls associated with Fourier transforms as well as the \ac{fft} algorithm apply. 
Such limitations include the resolution and maxima constraints described in \cref{sec:method:radar}, but also extend to the effects of quantization, which will be the basis of the modeling. The effects of limits in resolution are visualized for a low- and high-resolution \ac{fft} of a signal with two closely spaced frequencies in \cref{fig:method:fft:resolution}.
Since the \ac{fft} produces results which are quantized, the accuracy of a given measurement can be related to the degree of quantization. For example, given bins of width $l_\text{bin}$, the error can be modeled as a centered, uniform distribution with bounds $\pm \tfrac{l_\text{bin}}{2}$, denoted as $\uniform{-\tfrac{l_\text{bin}}{2}}{\tfrac{l_\text{bin}}{2}}$. 

\begin{figure}[h]
    \centering
    \includegraphics[width=\linewidth]{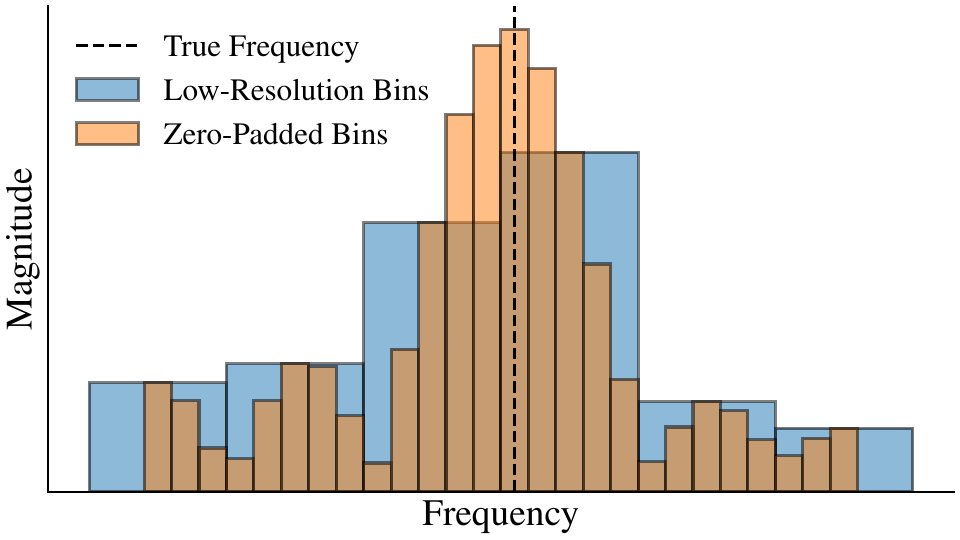}
    \spacingFigureCaption
    \caption{A comparison of how \ac{fft} quantization can be reduced at the cost of additional compute by zero-padding the signal.}
    \label{fig:method:fft:accuracy}
\end{figure}

This approach will be applied in the following sections when considering the noise modeling of the radar measurements. 
Note that whereas the \ac{fft} quantization can be changed by padding the signal with zeros, at the cost of additional computation, the resolution requires changes to the chirp or antenna layout, following \cref{sec:method:radar}. The effects of increased zero-padding in an \ac{fft} can be seen in \cref{fig:method:fft:accuracy}, where the resulting reduced quantization is clearly seen.
Changes to the chirp will have ramifications on the resolution/maxima of other measurements, and changes to the antenna layout may not even be possible. 
For this reason, the \ac{fft} dimension is typically either matching or with higher accuracy than what results from the resolution, especially for measurements with characteristics defined by the antenna layout.
Therefore, for noise modeling, we concern ourselves with the uncertainty derived from \ac{fft} discretization, assuming that resolution, particularly angular resolution, can be coarser and may misrepresent the sensor's actual accuracy. 
Measurements which result from ambiguity due to limited resolution will have different noise characteristics. The degree to which this differs depends on the discrepancy between the \ac{fft} quantization and resolution.

Of the measurements mentioned previously, two classes emerge: those that result from linear transformations of the \ac{fft} result, and those from non-linear transformations. As a measurement derived from an \ac{fft} can be considered a random variable, this distinction is of importance. Linear transformations of Gaussian or uniformly distributed random variables only scale and offset the distribution, whereas nonlinear transformations potentially alter the distribution entirely. As such, for measurements belonging to the linear class, it is equally convenient to work with the actual measurement (e.g., phase shift) as it is to work with the derived quantity (e.g., Doppler). However, for measurements which arise from nonlinear transformations of the \ac{fft} result, this is not the case.

With regards to applicability for aided inertial navigation, there are two mutually compatible pathways: utilizing the geometrical information of an environment (measured through range and \ac{aoa}) or the environment-relative velocity (measured through Doppler and \ac{aoa}). Radial speed is a powerful measurement for aided inertial navigation, as a given point's radial speed is related to the instantaneous radar ego-velocity $\vec{v}{IR}{R}$ as a function of the point's bearing vector $\bm{\mu}$ by

\begin{equation}\label{eq:method:doppler_from_ego}
    v_r = - \bm{\mu}^\top \vec{v}{IR}{R},
\end{equation}
thus enabling velocity-aiding with instantaneous measurements. Despite the noisy and problematic characteristics of the position measurements (resulting from effects such as multi-path reflections), they can also be used for aiding, where a target's position $\bm{t}$ is simply

\begin{equation}
    \bm{t} = \bm{\mu} d.
\end{equation}
Both of the aforementioned strategies utilize the bearing vector, which is related to the angles of azimuth and elevation of a target by

\begin{equation}\label{eq:method:radar:bearing}
    \bm{\mu} = \begin{bmatrix}\cos\theta\cos\phi\\ \sin\theta\cos\phi\\ \sin\phi \end{bmatrix}.
\end{equation}
As a result, neither Doppler nor position residuals can be expressed solely with additive measurement noise, as the term $\bm{\mu}$ is also a measurement of the radar and thus includes its own uncertainty. This means that the usage of either the radial speed or position measurements for aided inertial navigation includes both additive and multiplicative measurement noise terms. The following sections will consider each measurement type and derive noise models based on the underlying sensing principles. The underlying assumption here is that measurements are independent from one another. Furthermore, in the following sections, noise sources will be modeled as uniform, as this relates to the quantization of the \acs{fft} output. However, this is not practical to use in the estimator framework, which only considers the noise mean and standard deviation. As a result, an approximate Gaussian distribution can be created to convey their mean and standard deviation. This is justified by the estimator averaging many residuals during optimization; as a result, the overall distribution is expected to approach a Gaussian distribution as the number of samples increases. However, for significantly skewed distributions, this may not be sufficient. In such cases, the estimator will weigh measurement residuals incorrectly, and thus outlier rejection will be less effective, degrading the performance. Such effects are expected to be more significant for \acs{aoa} measurements near the theoretical \SI{\pm 90}{\degree} limits, where the nonlinearities are most exaggerated.

\subsubsection{Range}\label{sec:method:noise:range}
Range is a linear transformation of the output of an \ac{fft}, following~\cref{eq:method:range}. Assuming that the noise resulting from the \ac{fft} quantization is dominating, the statistics are simple. Consider the uncertainty of the range measurement $\tilde{d}$, which can be modeled as

\begin{equation}
    \tilde{d} = d + \noise{d},\quad \noise{d} \sim \uniform{-\frac{l_{d}}{2}}{\frac{l_{d}}{2}},
\end{equation}
for the range measurement noise $\noise{d}$ and \ac{fft} bin width $l_d$, after being converted to appropriate units following \cref{eq:method:range}. Since the transformation from \ac{fft} output to range measurement is linear, the measurement noise retains the uniform distribution, and as a result, the mean and variance of the noise are

\begin{align}
    \E{\noise{d}} &= 0,\\
    \sigma_{\noise{d}}^2 &= \frac{l_{d}^2}{12},
    \label{eq:method:range:variance}
\end{align}
where $\E{\cdot}$ is the expected value and $\sigma_{\noise{d}}^2$ is the range measurement noise variance.

\subsubsection{Radial Speed}\label{sec:method:noise:radial_speed}
Similarly to the range, the radial speed measurement is also a linear transformation of the \ac{fft} output, following \cref{eq:method:doppler}, and thus, it will be modeled in the same manner. Thus the Doppler velocity measurement $\tilde{v}_r$ can be written as

\begin{equation}
    \tilde{v}_r = v_r + \noise{v_r},\quad \noise{v_r} \sim \uniform{-\frac{l_{v_r}}{2}}{\frac{l_{v_r}}{2}},
\end{equation}
for the Doppler measurement noise $\noise{v_r}$ and \ac{fft} bin width $l_{v_r}$, after conversion to appropriate units. Thus the mean and variance $\sigma_{\noise{v_r}}^2$ of $\noise{v_r}$ are

\begin{align}
    \E{\noise{v_r}} &= 0,\\
    \sigma_{\noise{v_r}}^2 &= \frac{l_{v_r}^2}{12}. 
    \label{eq:method:doppler:variance}
\end{align}

\subsubsection{Bearing}\label{sec:method:noise:bearing}
The bearing vector can be calculated as a function of the azimuth and elevation following \cref{eq:method:radar:bearing}.
However, this does not provide a clear mapping to the measured quantities. In this case, the mapping from phase angles to \ac{aoa} is a nonlinear transformation (see \cref{eq:method:wy,eq:method:wz}) and thus cannot be modeled as simply as was the case for range and Doppler in \cref{sec:method:noise:range,sec:method:noise:radial_speed}.
By inspection of \cref{eq:method:wy,eq:method:wz}, one can find an equivalent expression for bearing directly as a function of the measured phases:

\begin{equation}\label{eq:method:bearing_from_phase}
    \bm{\mu} = \begin{bmatrix}\sqrt{1 - \frac{1}{\pi^2} (w_y^2 + w_z^2)}\\ \frac{1}{\pi}w_y\\ \frac{1}{\pi}w_z \end{bmatrix}.
\end{equation}
As before, assume the uncertainty in the phase measurements $\tilde{w}_y$, $\tilde{w}_z$ is dominated by the quantization such that

\begin{align}
    \tilde{w}_y &= w_y + \noise{w_y},\quad \noise{w_y} \sim \uniform{-\frac{l_{w_y}}{2}}{\frac{l_{w_y}}{2}},\\
    \tilde{w}_z &= w_z + \noise{w_z},\quad \noise{w_z} \sim \uniform{-\frac{l_{w_z}}{2}}{\frac{l_{w_z}}{2}},
\end{align}
for the measurement noises $\noise{w_y}$, $\noise{w_z}$ and bin widths $l_{w_y}$, $l_{w_z}$. The phase shift measurement statistics can thus be modeled as

\begin{align}
    \E{\Noise{w}} &= 0,\\
    \Sigma_{\Noise{w}} &=  
    \begin{bmatrix} \frac{l_{w_y}^2}{12} &0\\ 0 &\frac{l_{w_z}^2}{12} \end{bmatrix},
\end{align}
for the \ac{aoa} phase vector $\bm{w} = \begin{bmatrix} w_y &w_z\end{bmatrix}^\top$, noise vector $\Noise{w} = \begin{bmatrix} \noise{w_y} &\noise{w_z}\end{bmatrix}^\top$, and covariance matrix $\Sigma_{\Noise{w}}$. Note that the horizontal and vertical phase noises are assumed to be uncorrelated, as the corresponding measurements are calculated from separate \acp{fft} and thus have independent quantization noise. Residual errors arising from systemic phenomena, e.g., calibration errors, may give rise to correlations in practice, but these are assumed to be negligible.

Since the analytical expectation is intractable, the resulting bearing covariance can be approximated to first order by linearization. The gradient of the noisy bearing measurement $\bm{\tilde{\mu}}$ with respect to the noise source $\Noise{w}$ is

\begin{equation}
    \jacobian{\bm{\tilde{\mu}}}{\Noise{w}} = \begin{bmatrix} \frac{-\tilde{w}_y}{\pi^2 \sqrt{1 - \frac{1}{\pi^2} (\tilde{w}_y^2 + \tilde{w}_z^2)}} &\frac{-\tilde{w}_z}{\pi^2 \sqrt{1 - \frac{1}{\pi^2} (\tilde{w}_y^2 + \tilde{w}_z^2)}}\\ \frac{1}{\pi}  &0\\ 0  &\frac{1}{\pi} \end{bmatrix}.
\end{equation}
As a result, the approximate noise model for $\bm{\tilde{\mu}}$ is

\begin{align}
    \E{\Noise{\mu}} &\approx \bm{0},\\
    \Sigma_{\Noise{\mu}} 
    &\approx \jacobian{\bm{\tilde{\mu}}}{\Noise{w}} \Sigma_{\Noise{w}} \left( \jacobian{\bm{\tilde{\mu}}}{\Noise{w}} \right)^\top,\label{eq:method:noise:bearing}
\end{align}
for the bearing measurement noise $\Noise{\mu}$ and covariance matrix $\Sigma_{\Noise{\mu}}$. Although not strictly correct, it is convenient when designing the estimator to approximate this noise as normally distributed, denoted in this case as $\normal{\bm{0}}{\Sigma_{\Noise{\mu}}}$.

This covariance depends on unknown values for which we only have noisy measurements and assumes zero-mean offsets, which may not necessarily be true. Both conditions are unfortunate realities that are typically encountered with nonlinear transformations of random variables and their linear approximations.

\subsection{Aided Inertial Navigation}\label{sec:method:navigation}
\begin{figure*}[t!]
    \centering
    \includegraphics[width=\linewidth]{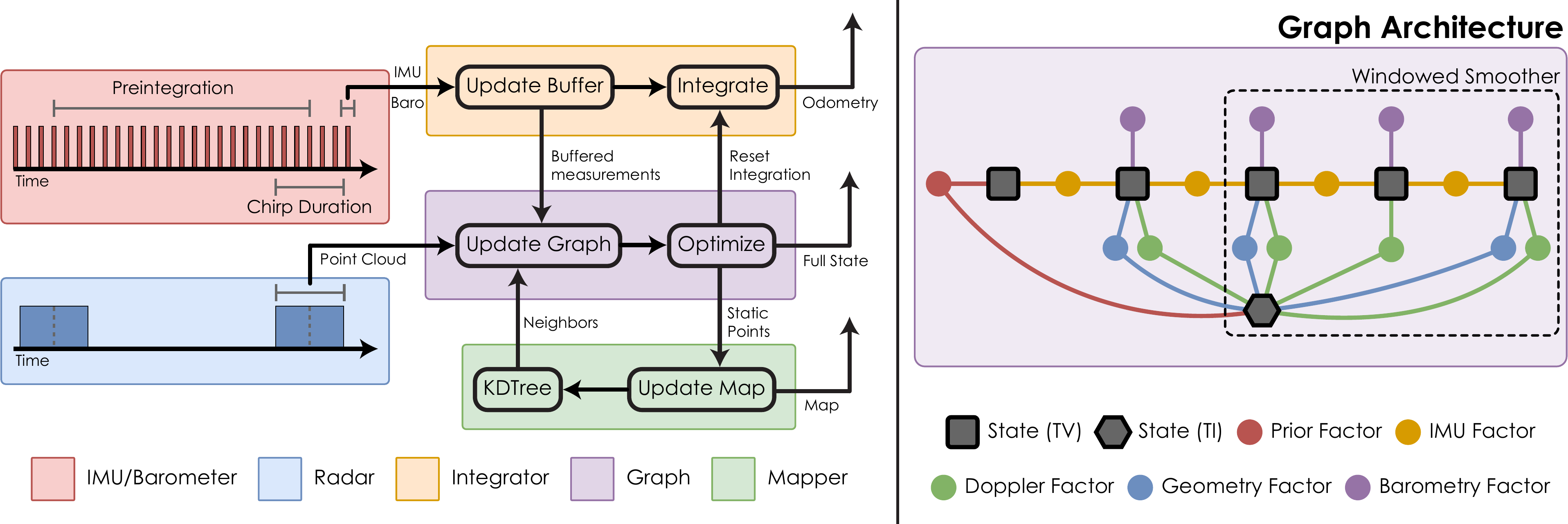}
    \spacingFigureCaption
    \caption{Visualization of the information flow from the sensors through the method and a depiction of the architecture of the proposed factor graph estimator.}
    \label{fig:method:architecture}
\end{figure*}
From the principles of \ac{fmcw} radar sensing and noise modeling presented in the previous section, this section develops a factor graph-based estimator. The motivation is twofold: firstly, to have an estimator that relates meaningfully to how the measurements are calculated, and secondly, to use such modeling to motivate estimator noise parameters.

Radar measurements, in particular radial speed, can be sources of high-quality information in the context of an aided \ac{ins}. The goal of the \ac{ins} is to create real-time estimates of the platform attitude $\rot{R}{B}{I}$, position $\vec{p}{IB}{I}$, and linear velocity $\vec{v}{IB}{I}$, all with respect to an inertial coordinate frame. Typically, due to imperfections in the \ac{imu} sensor, additional states are added to further increase performance. In this architecture, we assume inputs of angular velocity $\vec{\omega}{IB}{B}$ and specific force $\vec{f}{IB}{B}$ from the \ac{imu}, with noisy measurements $\vec{\tilde{\omega}}{IB}{B}$ and $\vec{\tilde{f}}{IB}{B}$ modeled as

\begin{align}
    \vec{\tilde{\omega}}{IB}{B} &= \vec{\omega}{IB}{B} + \vec{b}{g}{} + \Noise{\omega},
    &\Noise{\omega} \sim \normal{\bm{0}}{\Sigma_{\Noise{\omega}}},\\
    \vec{\tilde{f}}{IB}{B} &= \vec{f}{IB}{B} + \vec{b}{a}{} + \Noise{f},
    &\Noise{f} \sim \normal{\bm{0}}{\Sigma_{\Noise{f}}},
\end{align}
where $\Noise{\omega}$ and $\Noise{f}$ are the Gaussian measurement noises (with diagonal covariance matrices $\Sigma_{\Noise{\omega}}$ and $\Sigma_{\Noise{f}}$) for angular rate and specific force, 
and $\vec{b}{g}{}$ and $\vec{b}{a}{}$ are the gyroscope and accelerometer biases. Consequently, \ac{imu} measurement bias states for the accelerometer and gyroscope are included in the state space to be estimated online. Optionally, the proposed method can utilize a barometer for relative height information. This is not always appropriate, as some environments (namely, indoors) can contain excessive disturbances, which can complicate the use of such a sensor \cite{parviainen2008barometry}. To aid in handling unmodeled aerodynamic disturbances, a barometer bias term $\scalar{b}{b}{}$ is appended to the state vector. Furthermore, it can often be of interest to consider online estimation of the aiding sensor extrinsic transform, including the rotation $\rot{R}{R}{B}$ and translation $\vec{l}{BR}{B}$ of the radar frame with respect to the body frame. The extrinsic transform is assumed to have zero dynamics; thus, the total estimator state space $\mathbf{x}$ can be partitioned into time-varying $\mathbf{x}_\mathtt{TV}$ and time-invariant states $\mathbf{x}_\mathtt{TI}$ such that

\begin{align}
    \mathbf{x} &= \left( \mathbf{x}_\mathtt{TV},\ \mathbf{x}_\mathtt{TI} \right),\\
    \mathbf{x}_\mathtt{TV} &= \left( \rot{R}{B}{I},\ \vec{p}{IB}{I},\ \vec{v}{IB}{I},\ \vec{b}{a}{},\ \vec{b}{g}{},\ \scalar{b}{b}{} \right),\\
    \mathbf{x}_\mathtt{TI} &= \left( \rot{R}{R}{B},\ \vec{l}{BR}{B} \right).
\end{align}
With respect to dynamics, we consider a local-tangent plane inertial frame and neglect the effects of earth-rate and transport-rate, such that the dynamics of the state are

\begin{align}
    \rot{\dot{R}}{B}{I} &= \rot{R}{B}{I} (\vec{\omega}{IB}{B})^\times,\\
    \vec{\dot{p}}{IB}{I} &= \vec{v}{IB}{I},\\
    \vec{\dot{v}}{IB}{I} &= \rot{R}{B}{I} \vec{f}{IB}{B} + \vec{g}{}{I},\\
    \vec{\dot{b}}{a}{} &= \Noise{a}, &&\Noise{a} \sim \normal{\bm{0}}{\Sigma_{\Noise{a}}},\\
    \vec{\dot{b}}{g}{} &= \Noise{g}, &&\Noise{g} \sim \normal{\bm{0}}{\Sigma_{\Noise{g}}},\\
    \scalar{\dot{b}}{b}{} &= \noise{b}, &&\noise{b} \sim \normal{0}{\sigma_{\noise{b}}^2},\\
    \rot{\dot{R}}{R}{B} &= \bm{0},\\
    \vec{\dot{l}}{BR}{B} &= \bm{0},
\end{align}
where $(\cdot)^\times$ denotes the skew-symmetric matrix of a vector in $\mathbb{R}^3$, $\vec{g}{}{I}$ is the acceleration due to gravity, and $\Noise{a}$, $\Noise{g}$, $\noise{b}$ are white, Gaussian noise processes (with diagonal covariance matrices $\Sigma_{\Noise{a}}$, $\Sigma_{\Noise{g}}$ and variance $\sigma_{\noise{b}}^{2}$) driving the accelerometer, gyroscope, and barometer bias dynamics, respectively.

The proposed method produces both low-rate and high-rate output estimates. Upon receiving an \ac{imu} measurement, the state vector is propagated following the dynamics model, and the updated odometry estimate is output. This high-rate output is critical for onboard tasks such as control. Upon receiving a radar measurement, the graph is updated and a new, optimal state estimate over the smoother window is calculated, resulting in an update to the full state vector. Following the optimization, the high-rate odometry integrator is re-integrated from the new state estimate up to the timestamp of the newest \ac{imu} measurement. This ensures the high-rate odometry remains consistent with respect to the low-rate state estimates.

The optimal state estimate is calculated with a factor graph-based estimator, taking advantage of the GTSAM library~\cite{gtsam} and the iSAM2 optimizer~\cite{kaess2011isam2}. The estimator fuses \ac{imu} preintegration factors, radar Doppler factors, and optionally radar geometric registration and/or differential barometry factors. To ensure real-time performance, the factor graph will solve the optimization problem in a windowed smoother, bounded in length by time. Once a factor ages out of the window, it is removed from the optimization~\cite{kaess2011isam2} and replaced by a newly added marginal factor connected to the affected states, following~\cite{gtsam}. Thus, the optimal state estimation problem considers the states from the current state at time $t_{k}$ to the state at $t_{k-\ell}$, given the smoother lag duration $\ell+1$, such that the state history is $\mathcal{X}_{k-\ell:k} = \{ \mathbf{x}_{\mathtt{TV},k},\ \mathbf{x}_{\mathtt{TV},k-1},\ \ldots,\ \mathbf{x}_{\mathtt{TV},k-\ell},\ \mathbf{x}_\mathtt{TI} \}$.

Measurement factors are incorporated into the factor graph by a residual-covariance matrix pair, which relates the measured quantity to the states of interest and weights the residuals according to their uncertainties. Notation for the measurements themselves are as follows, the set of \ac{imu} measurements timestamped between times $t_i$ and $t_j$ are $\mathcal{I}_{i,j}$ the radar measurement with mid-chirp time at $t_i$ is $\mathcal{R}_i$, and the barometer measurement at time $t_i$ is $\mathcal{B}_i$. Due to imperfect time alignment, high-rate measurements such as \ac{imu} and barometer can be interpolated to match the timestamps of a lower-rate aiding sensor.
The \ac{imu} measurements for a given interval are used to create a preintegration factor with residual $\bm{e}_{\mathcal{I}_{i,j}}$ and covariance $\Sigma_{\mathcal{I}_{i,j}}$ following~\cite{forster2017manifold,gtsam}, which is briefly detailed in \cref{sec:method:imu}. The radar measurement with mid-chirp timestamp $i$ and $N_{\mathcal{R}_i}$ targets results in as many measurements for range, Doppler, and \ac{aoa}. The point cloud is first filtered for validity by removing points that exceed either the minimum or maximum limits of range, azimuth, or elevation. For each point in the filtered point cloud $\tau \in \mathcal{R}_i$ we add to the graph Doppler factors (with per-point residual $\bm{e}_{\mathcal{D}_{\tau}}$ and variance $\sigma_{\mathcal{D}_{\tau}}^2$) following \cref{sec:method:doppler} and optional geometry registration factors (with per-point residual $\bm{e}_{\mathcal{G}_{\tau}}$ and covariance $\Sigma_{\mathcal{G}_{\tau}}$) following \cref{sec:method:dist_to_dist}. Upon adding the radar-derived factors, the barometer measurements are interpolated to the mid-chirp timestamp to calculate the relative height residual $e_{\mathcal{B}_i}$ with variance $\sigma_{\mathcal{B}}^{2}$.
The optimal state history estimate over the smoother window $\mathcal{X}_{k-\ell:k}^{*}$ is thus

\begin{equation}
\begin{aligned}
    \mathcal{X}_{k-\ell:k}^{*} &= \underset{\mathcal{X}_{k-\ell:k}}{\arg\min} \Bigg[ 
    \lVert \bm{e}_0 \rVert_{\Sigma_0}^{2} 
    + \sum_{i=k-\ell+1}^{k} \bigg( \lVert \bm{e}_{\mathcal{I}_{i-1,i}} \rVert_{\Sigma_{\mathcal{I}_{i,j}}}^{2} \bigg)\\
    &+ \sum_{i=k-\ell}^{k} \bigg( \rho_{\mathcal{B}} \left( \lVert e_{\mathcal{B}_i} \rVert_{\sigma_{\mathcal{B}}^{2}}^{2} \right)\\
    &+ \sum_{\tau\in\mathcal{R}_i} \bigg( \rho_{\mathcal{D}} \left( \lVert {e}_{\mathcal{D}_{\tau}} \rVert_{\sigma_{\mathcal{D}_{\tau}}^{2}}^{2} \right) 
    + \lVert \bm{e_}{\mathcal{G}_{\tau}} \rVert_{\Sigma_{\mathcal{G}_{\tau}}}^{2} \bigg) \bigg)
    \Bigg],
\end{aligned}
\end{equation}
where $\bm{e}_{0}$, $\Sigma_{0}$ are the marginalization prior residual and covariance matrix, and $\rho_{\mathcal{B}}$, $\rho_{\mathcal{D}}$ denote the Huber and Cauchy M-estimator influence functions applied to the barometry and Doppler factor residuals, respectively. The influence functions enhance the robustness of the solution against outliers by introducing residual weighting, following \cite{huber2004robust}. Different functions can provide different results, and more details for each factor will be covered in their respective sections. The optimization is iterated to refine the estimate, which also enables updating of the noise model linearization, further decreasing linearization errors. Furthermore, the method utilizes an initialization routine to set the prior factor on startup. Assuming the platform starts at rest, estimates for the gyro bias and angles of roll and pitch can be made by averaging the \ac{imu} for a short duration. Roll and pitch estimates will be worsened by the uncompensated bias in the accelerometer; however, this effect is small and considered by inflating the initial uncertainty. The architecture for the overall method, as well as the graph architecture, is shown in \cref{fig:method:architecture}.

Note that a naive implementation of point-based factors (e.g., the Doppler factor) would create a separate factor for each point in a point cloud measurement. This can be computationally expensive to handle, as expensive operations such as marginalization would need to be considered per factor. Thus, such factors are included as Hessian factors, i.e., a single factor containing all residuals derived from $N_{\mathcal{R}_i}$ points from a given point cloud measurement $\mathcal{R}_i$ (e.g., all the Doppler residuals for a point cloud). This is well-defined for any size point cloud (assuming at least 1 point) and, as the points are considered independent, lends itself well to multi-threading.

\subsubsection{IMU Factor}\label{sec:method:imu}
To connect successive nodes in the graph, an \ac{imu} preintegration factor is added following the derivation of~\cite{forster2017manifold}. This is a computational-cost saving measure that allows one to avoid creating nodes at the rate of the \ac{imu}, which would otherwise become prohibitively expensive. Instead, this preintegration factor encodes the effects of all \ac{imu} measurements from time $t_i$ to $t_j$ on orientation, position, and velocity into a single factor, with residuals $\bm{e}_{\mathcal{I}_{\mathbf{R}_{i,j}}}$, $\bm{e}_{\mathcal{I}_{\bm{p}_{i,j}}}$, and $\bm{e}_{\mathcal{I}_{\bm{v}_{i,j}}}$ such that
\begin{equation}
    \bm{e}_{\mathcal{I}_{i,j}} = \begin{bmatrix} \bm{e}_{\mathcal{I}_{\mathbf{R}_{i,j}}} &\bm{e}_{\mathcal{I}_{\bm{p}_{i,j}}}  &\bm{e}_{\mathcal{I}_{\bm{v}_{i,j}}} \end{bmatrix}.
\end{equation}

\subsubsection{Doppler Factor}\label{sec:method:doppler}
Note that in this section, the per-point notation is dropped for simplicity.
To make use of the relationship between radar ego-motion and the radial speed of a static point in the environment, the relationship from \cref{eq:method:doppler_from_ego} must be written as a function of the estimator state. By considering the effect of the lever arm between the \ac{imu} and radar sensors, the radar-frame linear velocity is

\begin{equation}
    \vec{v}{IR}{R} = (\rot{R}{R}{B})^\top \left( (\rot{R}{B}{I})^\top \vec{v}{IB}{I} + \left( \vec{\bar{\omega}}{IB}{B} \right)^{\times} \vec{l}{BR}{B} \right),
\end{equation}
given the average angular rate during the chirping period $\vec{\bar{\omega}}{IB}{B}$. The measured average angular rate $\vec{\tilde{\bar{\omega}}}{IB}{B}$, taken from $N_{\vec{\bar{\omega}}{IB}{B}}$ noisy measurements, can be approximated as

\begin{equation}
\begin{aligned}\label{eq:method:residual:average_angrate}
    \vec{\tilde{\bar{\omega}}}{IB}{B} \approx \vec{\bar{\omega}}{IB}{B} + \vec{b}{g}{} + \Noise{\bar{\omega}}, \quad \Noise{\bar{\omega}} \sim \normal{\bm{0}}{\frac{1}{N_{\vec{\bar{\omega}}{IB}{B}}}\Sigma_{\Noise{\omega}}},
\end{aligned}
\end{equation}
for the averaged angular rate noise $\Noise{\bar{\omega}}$, under the assumption that the radar chirp duration is sufficiently short such that the effects of bias evolution and changing body-frame attitude do not have meaningful effects.
Thus, the typical Doppler residual used for radar-inertial estimation is

\begin{equation}
    e_\mathcal{D} = -\bm{\mu}^\top \vec{\hat{v}}{IR}{R} - \tilde{v}_r,
\end{equation}
given the radar-frame velocity estimate $\vec{\hat{v}}{IR}{R}$. The assumption here is that measurement noise only enters in the Doppler term, following the modeling of \cref{sec:method:noise:radial_speed}. This can be problematic as the bearing angle and the angular rate are also measurements with, potentially, significant measurement noise. A more accurate residual should consider radial speed, bearing, and angular rate as measurements, using the results of \cref{sec:method:noise:radial_speed,sec:method:noise:bearing}, such that the residual is of the form

\begin{equation}\label{eq:method:residual:doppler:mean}
\begin{aligned}
    e_\mathcal{D} &= -\bm{\tilde{\mu}}^\top \vec{\hat{v}}{IR}{R} - \tilde{v}_r,\\
    &= -\bm{\tilde{\mu}}^\top (\rot{\hat{R}}{R}{B})^\top \left( (\rot{\hat{R}}{B}{I})^\top \vec{\hat{v}}{IB}{I} + (\vec{\tilde{\bar{\omega}}}{IB}{B} - \vec{\hat{b}}{g}{})^{\times} \vec{\hat{l}}{BR}{B} \right)\\ & \phantom{{}=} - \tilde{v}_r.
\end{aligned}
\end{equation}
using the estimates for attitude $\rot{\hat{R}}{B}{I}$, velocity $\vec{\hat{v}}{IB}{I}$, gyroscope bias $\vec{\hat{b}}{g}{}$, and radar-\ac{imu} extrinsic rotation $\rot{\hat{R}}{R}{B}$ and translation $\vec{\hat{l}}{BR}{B}$.
The covariance for the Doppler residual can be approximated to first order as

\begin{equation}\label{eq:method:residual:doppler:covariance}
\begin{aligned}
    \sigma_{\mathcal{D}}^{2} &\approx \sigma_{\noise{v_r}}^2 + (\vec{\hat{v}}{IR}{R})^\top \Sigma_{\Noise{\mu}} \vec{\hat{v}}{IR}{R}\\ &\phantom{{} \approx \sigma_{\noise{v_r}}^2} + \jacobian{e_\mathcal{D}}{\Noise{\bar{\omega}}} \Sigma_{\Noise{\bar{\omega}}} \left( \jacobian{e_\mathcal{D}}{\Noise{\bar{\omega}}} \right)^\top,\\
    \jacobian{e_\mathcal{D}}{\Noise{\bar{\omega}}} &= \bm{\tilde{\mu}}^\top (\rot{\hat{R}}{R}{B})^\top (\vec{\hat{l}}{BR}{B})^\times,
\end{aligned}
\end{equation}
by applying the results of \cref{eq:method:doppler:variance,eq:method:noise:bearing}. As this depends on estimated variables, the covariance will need to be updated per iteration as the state estimate evolves in order to reduce bias in the final estimate.

The gyroscope-derived noise term here is shown for completeness, but will actually be omitted from the implementation. The effect from this term is already small due to the relatively low noise in the gyroscope, and is further reduced in magnitude by the averaging and the short \ac{imu}-radar lever arm acting through the gradient $\jacobian{e_\mathcal{D}}{\Noise{\bar{\omega}}}$. Vehicles with larger \ac{imu}-radar lever arms or relatively less precise gyroscopes may want to consider this effect.

The properties of the residual covariance approximation will be further analyzed in \cref{sec:evaluation:noise}.

\paragraph{Outlier Rejection} Many other methods~\cite{kramer2024suburban,michalczyk2023tight} solve a \ac{ransac} least squares problem to separate outliers; however, this is only feasible when point cloud measurements are sufficiently dense and cover sufficient \ac{fov}. Following the motivation for fusing radial speed directly, we find that the Cauchy M-estimator suffices as a replacement, which does not require similar assumptions, and performs better for this task than more strict M-estimators, such as the Welsch considered in \cite{girod2024radar}. It is worth noting that outlier rejection of some type is important, particularly so for flight regimes which could potentially exceed the Doppler limit, as this will result in a region of the sensor \ac{fov} returning aliased measurements. This effect will be examined with greater detail in \cref{sec:evaluation:flight}.

\subsubsection{Distribution-to-Distribution Factor}\label{sec:method:dist_to_dist}
Note that in this section, the per-point notation is dropped for simplicity.
Although sparse, the radar returns can also be used for geometry-based fusion. However, due to the noisiness of the point cloud, relying on point-to-point association is impractical. Instead, an approach motivated by the Normal Distributions Transform~\cite{biber2003NDT} can be taken, similarly to~\cite{zhang20234DRadarSlam,zhu2025Robustradar}, but utilizing the proposed noise models. For \ac{swap}-limited radar sensors, the viability of such a method depends on the overlap between the chirp configuration and the environment's characteristics. For example, in a too sparse environment, it may not be practically feasible to gain performance from such a factor.

The justification is as follows: due to range and \ac{aoa} ambiguity, multi-path effects, and other phenomena, the point cloud remains too noisy to consider point-to-point residuals. However, the aggregated point clouds still represent geometry in a recognizable way. As a result, it is logical to consider point-to-distribution or, in reality, distribution-to-distribution residuals to include the proposed noise models.

To facilitate this, static points from the radar point clouds are stored along with the estimated pose in a monolithic map, which is updated and voxel downsampled after completing graph optimization. Static points are selected as those with Doppler residuals which are less than a threshold $\kappa_\text{static}$ after whitening, e.g., $\tfrac{e_{\mathcal{D}}}{\sigma_{\mathcal{D}}} < \kappa_\text{static}$. Upon receiving a new point cloud measurement, the down-sampled map is used to create a KD tree~\cite{blanco2014nanoflann}. For each point $\bm{t}$ in the measurement point cloud, a search is performed for all neighbors within a radius $\rho$. If the number of neighbors is greater than a threshold, then the centroid mean $\vec{\tilde{q}}{}{}$ and covariance $\Sigma_{\Noise{q}}$ of the neighbors are calculated. This results in the following residual defined per point position measurement $\bm{\tilde{t}}$ in the current point cloud

\begin{equation}
\begin{aligned}\label{eq:method:residual:registration}
    \bm{e}_\mathcal{G} &= \rot{\hat{R}}{B}{I} (\rot{\hat{R}}{R}{B} \bm{\tilde{t}} + \vec{\hat{l}}{BR}{B}) + \vec{\hat{p}}{IB}{I} - \vec{\tilde{q}}{}{},\\
    &= \rot{\hat{R}}{B}{I} (\rot{\hat{R}}{R}{B} (\tilde{d}\bm{\tilde{\mu}}) + \vec{\hat{l}}{BR}{B}) + \vec{\hat{p}}{IB}{I} - \bm{\tilde{q}},
\end{aligned}
\end{equation}
for the position estimate $\vec{\hat{p}}{IB}{I}$.
The residual covariance can then be calculated, using the first-order approximation, as

\begin{equation}\label{eq:method:residual:registration:covariance}
    \Sigma_{\mathcal{G}} \approx \rot{\hat{R}}{B}{I}\rot{\hat{R}}{R}{B} ( \sigma_d^2 \bm{\tilde{\mu}} \bm{\tilde{\mu}}^\top + {\tilde{d}}^2 \Sigma_{\Noise{\mu}} ) ( \rot{\hat{R}}{B}{I}\rot{\hat{R}}{R}{B} )^\top + \Sigma_{\Noise{q}},
\end{equation}
by applying the range measurement variance from \cref{eq:method:range:variance} and assuming uncorrelated noise sources. Note that points with resolution ambiguity problems are not easily separated out and, therefore, will be added to the map. As a result, the neighbor covariance will reflect this fact and typically dominate over the point noise in this factor.

\subsubsection{Barometry Factor}\label{sec:method:diff_baro}
Note that in this section, the time index is omitted for simplicity. Vertical drift is a typical problem encountered with radar-inertial odometry methods \cite{michalczyk2022tight,nissov2024roamer}. Barometric pressure measurements have been used to substantially mitigate this effect \cite{doer2020ekf,girod2024radar}. The altitude can be calculated as a function of the pressure $h(P)$, following \cite{nasa1976atmosphere}, such that

\begin{equation}
    h(P) = \frac{T_0}{L_0} \left( 1 - \left( \frac{P}{P_0} \right)^\frac{R L_{0}}{g M} \right),
\end{equation}
for physical constants of standard temperature $T_0$,  temperature lapse rate $L_0$,  standard pressure $P_0$, gas constant $R$, standard acceleration due to gravity $g$, and average molar mass of air at sea level $M$. However, this absolute altitude will be erroneous without accurate information about environmental conditions. Instead, the pressure can be used as a relative measurement \cite{parviainen2008barometry,doer2021xrio}, conveying information about the change in height since the mission start. Thus, the barometer bias is initialized by the estimator to the initial height, such that the barometry factor residual is

\begin{equation}
    e_\mathcal{B} = \hat{z} - (h(\tilde{P}) - \scalar{\hat{b}}{b}{}),
\end{equation}
for the measured pressure $\tilde{P}$ and estimates of vertical position $\hat{z}$ and barometer bias $\scalar{\hat{b}}{b}{}$.

Given the high likelihood of disturbances to the sensor, the residual noise $\sigma_{\mathcal{B}}$ will likely need to be overinflated with respect to the sensor's actual precision. Furthermore, the online bias estimation will enable resilience to changes in atmospheric conditions. For outlier rejection against more sudden disturbances, a Huber M-estimator is used. Note that, for particular environments, especially indoors, the potential for aerodynamic disturbances can result in difficulties when using barometer measurements~\cite{parviainen2008barometry}; as such, this factor is not applied universally here.

\section{EVALUATION}\label{sec:evaluation}
\begin{figure*}
    \centering
    \includegraphics[angle=0,width=\linewidth]{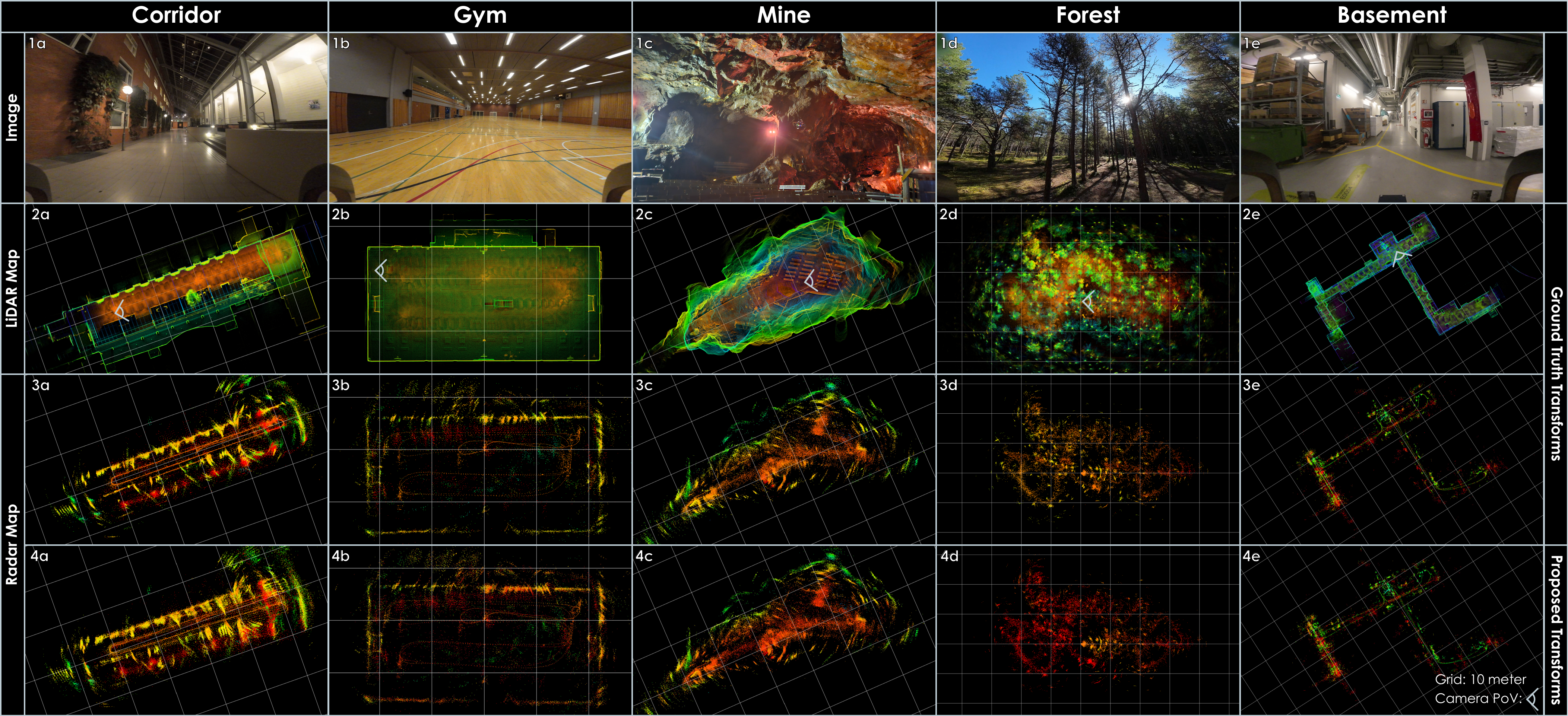}
    \spacingFigureCaption
    \caption{Images from the experiment environments (row 1) as well as maps resulting from the LiDAR (row 2) and radar point cloud data (rows 3 and 4) are shown for each of the \experiment{rec\_4mps\_align} (column a), \experiment{manual\_1} (column b), \experiment{manual\_3} (column c), \experiment{manual\_4} (column d), and \experiment{manual\_11} (column e) experiments. The maps from rows 2 and 3 are made with the ground truth transforms. The maps from row 4 are made with transforms from the proposed method (using the \config{noise} configuration). Note the clear differences in density and accuracy between the LiDAR- and radar-derived point clouds, resulting from increased noise in the radar measurements. Note also the minor but noticeable differences between rows 3 and 4, resulting from the drift of the proposed method, with respect to the ground truth.}
    \label{fig:evaluation:environments}
\end{figure*}
The evaluation is divided into two parts: first, in \cref{sec:evaluation:noise}, an evaluation of the Doppler factor residual is conducted using simulations; second, in \cref{sec:evaluation:validation} the theoretical noise models are validated empirically; third, in \cref{sec:evaluation:flight}, the proposed method is evaluated through flight experiments.

Four different chirp configurations, with theoretical properties as shown in \cref{tab:evaluation:chirp}, are used to highlight the impact of the proposed changes across different radar sensors and operating regimes. This is relevant, as the angular noise can be overwhelmed if the additive Doppler noise is too high. These configurations will be referred to as \rc{\#}, and are designed as follows.
Of these, \rc{1} prioritizes point cloud density (through range resolution) at the cost of Doppler accuracy, while \rc{2} is designed to minimize the Doppler standard deviation. Both of these configurations are designed for the Texas Instruments IWR6843AOPEVM radar sensor.
The remaining two configurations target general-purpose performance with an Anteral uRAD Automotive radar (referred to as \rc{3}) and long-range performance with a Texas Instruments AWR1843BOOST radar (referred to as \rc{4}).

\begin{table}[ht!]
    \centering
    \caption{Radar Chirp Configurations}
    \label{tab:evaluation:chirp}
    \sisetup{
  round-mode=places,
  round-precision=3,
  round-pad=false,
  table-format=3.3,
  scientific-notation=false,
}%
\setlength{\tabcolsep}{2.25pt}%
\newcommand{\rotation}{90}%
\newcommand{\vmove}{-0.675ex}%
\begin{tabular}{l l S S S S l}
    \toprule
      \multicolumn{2}{l}{\multirow{2}{*}[\vmove]{Parameter}}   &\multicolumn{2}{c}{IWR6843AOP} &{uRAD Auto.}  &{AWR1843BOOST} &\multirow{2}{*}[\vmove]{Unit}\\ \cmidrule(lr){3-4} \cmidrule(lr){5-5} \cmidrule(lr){6-6}
      & &\rc{1} &\rc{2} &\rc{3}  &\rc{4} &\\
    \midrule
      &Frequency          &60         &60       &77     &77   &\si{\giga\hertz}\\
      % &Bandwidth          &1911.273   &700.066  &768    &     &\si{\mega\hertz}\\
    \midrule
    \multirow{2}{*}{\rotatebox{\rotation}{Max.}}
      &Range              &20.013 &13.713 &25.000 &62.495 &\si{\meter}\\
      &Doppler            &3.995  &3.148  &3.879  &2.021  &\si{\meter\per\second}\\
    \midrule
    \multirow{4}{*}{\rotatebox{\rotation}{Resolution}}
      &Range              &0.078  &0.214  &0.195  &0.244  &\si{\meter}\\
      &Doppler            &0.133  &0.049  &0.065  &0.126  &\si{\meter\per\second}\\
      &Azimuth            &29     &29     &29     &14     &\si{\degree}\\
      &Elevation          &29     &29     &38     &57     &\si{\degree}\\
    \midrule
    \multirow{3}{*}{\rotatebox{\rotation}{\ac{fft} Size}}
      &Range              &256    &64     &128    &256    &\si{\bins}\\
      &Doppler            &64     &128    &128    &32     &\si{\bins}\\
      &\ac{aoa} Phase     &64     &64     &64     &64     &\si{\bins}\\
    \bottomrule
\end{tabular}

\end{table}

The experiments were conducted using a custom quadrotor and handheld platforms. The quadrotor platform includes a Khadas VIM4 single-board-computer and VectorNav VN-100 \ac{imu} (\SI{200}{\hertz}), Texas Instruments IWR6843AOPEVM radar (\SI{10}{\hertz}), Anteral uRAD Automotive (\SI{10}{\hertz}), and Ouster OS0-128 LiDAR (\SI{10}{\hertz}) sensors. Synchronization is handled by an onboard microcontroller~\cite{nissov2025Sync}: the \ac{imu} and radar by hardware triggering, the LiDAR by \ac{nmea} messages, and the computer by \ac{ptp}.

The handheld platform allows for larger, longer-range radar sensors. It includes an NVIDIA Orin AGX computer and VectorNav VN-100 \ac{imu} (\SI{200}{\hertz}), Texas Instruments AWR1843BOOST radar (\SI{10}{\hertz}), and Ouster OS0-128 LiDAR (\SI{10}{\hertz}) sensors. The radar on this platform is synchronized by the \ac{imu} trigger pulse, and the LiDAR by \ac{ptp}. 

For the aerial platform, the LiDAR sensor is used in combination with the \ac{imu} to provide highly-accurate online odometry, adopting the LiDAR-only version of the estimator from~\cite{nissov2024degradation}. This is used by the position controller of the PX4 autopilot~\cite{meier2015px4,px4} to enable trajectory tracking-based missions with high repeatability in \cref{sec:evaluation:flight:doppler_accuracy,sec:evaluation:flight:doppler_limit}.

To evaluate the proposed method, different ablations are considered; however, all configurations include the Doppler factor as described in \cref{sec:method:doppler}. The proposed method, assuming zero angle noise (a common simplifying assumption), is denoted \config{base}. Note, that this simplification means that $\sigma_{\mathcal{D}}=\sigma_{\noise{v_r}}$. The proposed method with non-zero angle noise, resulting in the Doppler factor variance from \cref{eq:method:residual:doppler:covariance}, is denoted as \config{noise}. The ablation resulting from adding the distribution-distribution factors (from \cref{sec:method:dist_to_dist}) to the \config{noise} configuration is denoted as \config{geometry}, with the variance as given by \cref{eq:method:residual:registration:covariance}. Finally, the ablation resulting from adding the barometry factors to the \config{noise} configuration is denoted as \config{noise + baro}.

The experiments were carried out across five environments: inside a university corridor (referred to as \corridor{}), inside a handball court (referred to as \gym{}), inside an abandoned mine (referred to as \mine{}), throughout a boreal forest (referred to as \forest{}), and through narrow basement corridors (referred to as \basement{}). Note that not all radar sensors were used in each environment, the usage is summarized in \cref{tab:evaluation:radar_environment_matrix}. Representative images and point cloud maps are shown for each of the environments in \cref{fig:evaluation:environments}. This figure shows maps aggregated with the LiDAR-inertial odometry transforms, removing the possibility of blurring caused by radar-inertial inaccuracies, alongside maps created using the \config{noise} configuration of the proposed method. Note the clear differences in density and accuracy between the LiDAR- and radar-derived point clouds. This underlines the challenges of using geometry-based measurements and simultaneously the importance of Doppler measurements for estimation with such radars. While the underlying geometry is apparent, establishing clear associations between individual points in the LiDAR and radar maps is not feasible. However, groups of points seem to aggregate around distinct objects from the LiDAR map. Furthermore, since the radars on the aerial platform are mounted facing downwards, the radar measurements trace a line of points on the ground, reflecting the path flown by the system. This effect is most noticeable in the \corridor{} and \gym{} environments. Note also that the \mine{} environment is similar in size to the \gym{}, however, with a less planar and more textured interior, different from the narrow tunnel one might expect. This region was chosen for its length and free space to safely facilitate high-speed experiments.

\begin{table}[h]
    \centering
    \caption{Radar Sensor Usage by Environment}
    \label{tab:evaluation:radar_environment_matrix}
    \setlength{\tabcolsep}{5.0pt}%
\newcommand{\vmove}{-0.675ex}%
\begin{tabular}{lccccc}
    \toprule
    \multirow{2}{*}[\vmove]{Sensor} &\multicolumn{5}{c}{Environments}\\\cmidrule(l){2-6}
        &\Corridor{}    &\Gym{} &\Mine{}    &\Forest{}  &\Basement{}\\
    \midrule
    IWR6843AOP      &\checkmark  &\checkmark   &\checkmark   &\checkmark   &-\\
    uRAD Auto.      &-  &-   &-   &-   &\checkmark\\
    AWR1843BOOST    &\checkmark  &-   &-   &-   &-\\
    \bottomrule
\end{tabular}
\end{table}

In \cref{sec:evaluation:flight}, ground truth pose estimates are necessary for creating numerical results; for this purpose, we use the aforementioned LiDAR-inertial odometry solution. The LiDAR-inertial estimator is a reasonable choice for ground truth since the experiments are conducted in geometrically well-conditioned environments. As such, one should expect high accuracy from this method, as evidenced by the map quality (shown for a subset of experiments in \cref{fig:evaluation:environments}). 

Noise parameters for the \ac{imu} are calculated following the Allan variance method using~\cite{buchanan2021Allan}. Noise parameters for the different radar configurations are calculated according to \cref{sec:method:noise}, using the values from \cref{tab:evaluation:chirp}. The same parameters were used across different experiments and environments, and are all taken from either empirical models or from sensor specifications. Noise parameters for the radar are shown in \cref{tab:evaluation:noise_params:radar} and for the \ac{imu} are shown in \cref{tab:evaluation:noise_params:imu}.

\begin{table}[h]
    \centering
    \caption{Radar Noise Parameters}
    \label{tab:evaluation:noise_params:radar}
    \sisetup{
  round-mode=places,
  round-precision=3,
  table-format=1.3,
}
\newcommand{\vmove}{-0.675ex}%
\begin{tabular}{l S S S S l}
  \toprule
  \multirow{2}{*}[\vmove]{Parameter}  &\multicolumn{4}{c}{Values} &\multirow{2}{*}[\vmove]{Unit}\\ \cmidrule(lr){2-5}
    &\rc{1} &\rc{2} &\rc{3} &\rc{4} &\\
  \midrule
  $\sigma_{\noise{d}}$          &\num{0.0225} &\num{0.0618}  &\num{0.0563}  &\num{0.0704}                         &\si{\meter}\\
  $\sigma_{\noise{v_{r}}}$      &\num{0.03603928633457117}  &\num{0.01419920818288236}  &\num{0.017496419485832488} &\num{0.0364632779385073}                         &\si{\meter\per\second}\\
  $\sqrt{\Sigma_{\Noise{w}}}$   &\num{1.6237976320958225}   &\num{1.6237976320958225}   &\num{1.6237976320958225}   &\num{1.6237976320958225} &\si{\degree}\\
  \bottomrule
\end{tabular}
\end{table}
\begin{table}[h]
    \centering
    \caption{\ac{imu} Noise Parameters}
    \label{tab:evaluation:noise_params:imu}
    \sisetup{
  round-mode=places,
  round-precision=2,
  table-format=1.2,
}
\begin{tabular}{lSl}
    \toprule
    Parameter &{Value} &Unit\\
    \midrule
    $\sqrt{\Sigma_{\Noise{\omega}}}$      &\num{5.4380545102010436e-05}   &\si{\radian\per\second\per\sqrthertz}\\
    $\sqrt{\Sigma_{\Noise{f}}}$           &\num{1.3886655606357616e-3}    &\si{\meter\per\square\second\per\sqrthertz}\\
    $\sqrt{\Sigma_{\Noise{g}}}$           &\num{1.6587925152480572e-06}   &\si{\radian\per\second\sqrthertz}\\
    $\sqrt{\Sigma_{\Noise{a}}}$           &\num{8.538212723310593e-05}    &\si{\meter\per\square\second\sqrthertz}\\
    \bottomrule
\end{tabular}

\end{table}

\begin{figure*}
    \centering
    \subfloat[\label{fig:evaluation:noise:simulation:low_accuracy}]{\includegraphics[width=0.49\linewidth]{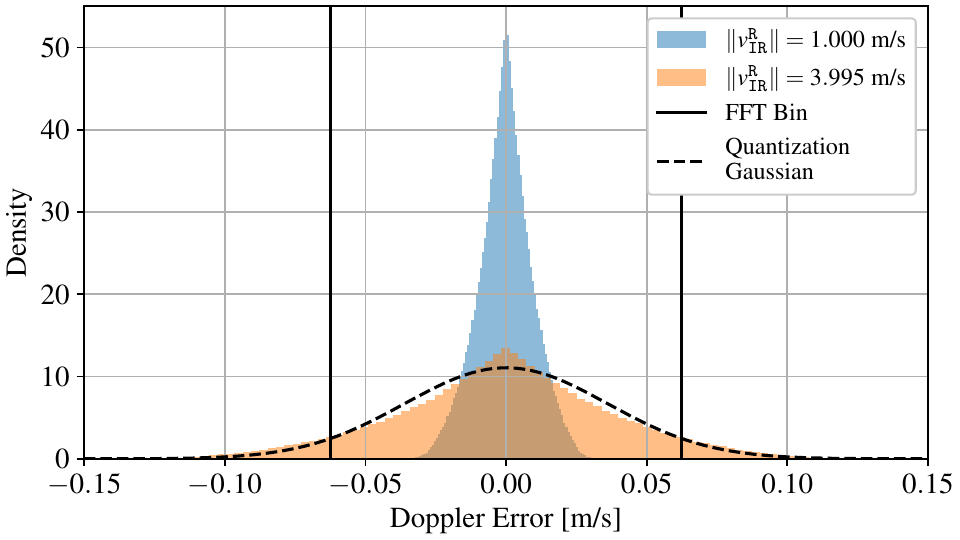}}
    \hfill
    \subfloat[\label{fig:evaluation:noise:simulation:high_accuracy}]{\includegraphics[width=0.49\linewidth]{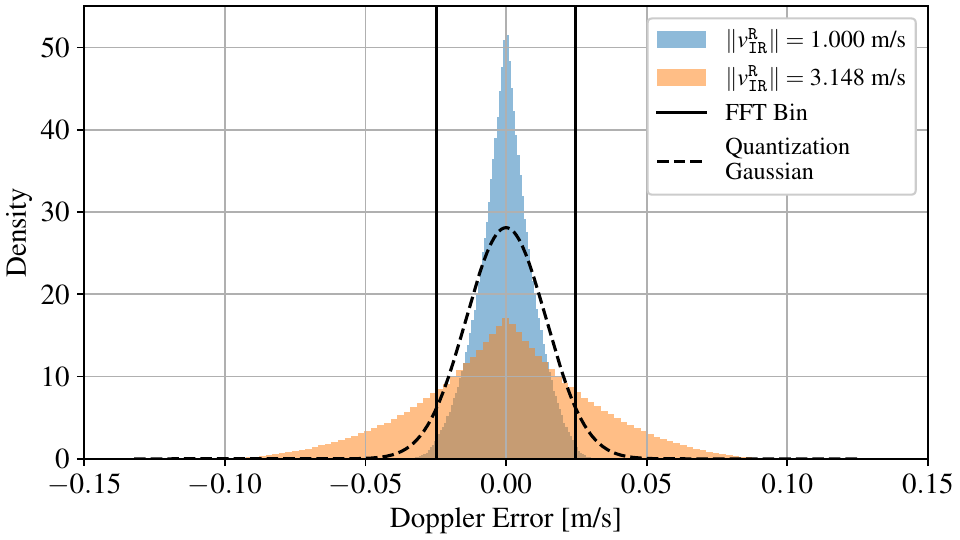}}
    \caption{Simulations of the Doppler factor residual error for (a) \rc{1} and (b) \rc{2} configurations for their respective maximum radar ego-velocities with proposed uniform distribution-based Doppler and \ac{aoa} measurement noise models. Note, the smaller Doppler bin width of \rc{2} compared to \rc{1} exaggerates the effects of the \ac{aoa} uncertainty.}
    \label{fig:evaluation:noise:simulation}
\end{figure*}
\subsection{Noise Modeling Simulation}\label{sec:evaluation:noise}
Regarding the noise modeling, the following pertinent questions arise. First, it concerns the necessity of the additional effort in modeling and compensating for these noise sources, and second, it concerns the validity of the first-order approximation in the absence of an analytical solution. Both will be analyzed in the context of the Doppler residual from \cref{eq:method:residual:doppler:mean} and its associated covariance. The necessity will be analyzed by comparing the contributions from Doppler and \ac{aoa} measurements to the factor covariance from \cref{eq:method:residual:doppler:covariance}. The prevailing assumption for typical radar-inertial methods is that the Doppler residual noise is constant and equal per point \cite{doer2021xrio,michalczyk2022tight,kramer2024suburban}. While this may be true for the Doppler measurement, it is not true for the bearing vector, which is a necessary component of the Doppler factor residual. Thus, a similar order of magnitude between $\sigma_{\noise{v_r}}^2$ and $(\vec{\hat{v}}{IR}{R})^\top \Sigma_{\Noise{\mu}} \vec{\hat{v}}{IR}{R}$, see \cref{eq:method:residual:doppler:covariance}, signifies that only considering Doppler uncertainty results in an optimistic measurement noise. The validity will be analyzed by comparing the approximated noise models with sample statistics in a simulated scenario.
These concepts will be analyzed with respect to the parameters of both of the chirp configurations illustrated in \cref{tab:evaluation:chirp}, in order to study the impact that the chirp signal has on the radial speed measuring accuracy.

\subsubsection{Significance of Doppler Accuracy}\label{sec:evaluation:noise:doppler_accuracy}
The simulation draws uniform samples for the azimuth and elevation angles with limits of \SI{\pm 60}{\degree}, matching the reported beam width of the radar sensor in question. Larger angles could be considered, as the half-wavelength antenna spacing results in a larger theoretical \ac{fov} (\SI{\pm 90}{\degree}). However, this will only exaggerate the effects of the \ac{aoa} uncertainty.

These angles are then converted to phases using \cref{eq:method:wy,eq:method:wz} and quantized according to the sensor configurations (\rc{1} and \rc{2}) from \cref{tab:evaluation:chirp}. This results in the noisy phase values $\tilde{w}_y$ and $\tilde{w}_z$, limited to \SI{\pm 180}{\degree} at \SI{5.625}{\degree} intervals. These phases are used to calculate the noisy bearing vector $\bm{\tilde{\mu}}$ following \cref{eq:method:bearing_from_phase}.

The Doppler error can thus be calculated as the difference between the true and noisy calculations of \cref{eq:method:doppler_from_ego} for different known values for the radar ego-velocity. The contribution of this to the overall Doppler residual will vary with respect to the velocity, following \cref{eq:method:residual:doppler:mean}. As a result, both slow (i.e., those with a magnitude of \SI{1}{\meter\per\second}) and fast (i.e., those with a magnitude equal to the Doppler limit) velocities are considered. Both scenarios are visualized for both radar configurations (from \cref{tab:evaluation:chirp}) in \cref{fig:evaluation:noise:simulation:low_accuracy,fig:evaluation:noise:simulation:high_accuracy}. There, the Doppler residual is visualized with noise-free ego-velocity and Doppler values, thus highlighting the impact of the \ac{aoa} noise, alongside the \ac{fft} bin size and the resulting Gaussian distribution.

It can be seen that as the Doppler accuracy increases (i.e., as the \ac{fft} bin width decreases), the contribution from angle measurement noise becomes more significant. For example, in the \rc{1} configuration, the contribution from the \ac{aoa} noise is a similar order of magnitude as the uncertainty derived from the \ac{fft} bin size. However, for the \rc{2} configuration, the \ac{aoa} noise contribution is more significant, showing heavy tails outside the \ac{fft} bin. This consideration applies to uniform distributions of \ac{aoa}. For skewed distributions, the significance can be greater due to the dependence on phase in the covariance calculation from \cref{eq:method:residual:doppler:covariance}.

\subsubsection{Validity of Approximation}\label{sec:evaluation:noise:validity}
The previous section demonstrated that the inaccuracy resulting from quantization can lead to an increase in the Doppler residual noise level. However, an analytical solution for this behavior remains intractable, motivating the first-order approximation from \cref{eq:method:residual:doppler:covariance}. This section will investigate the validity of the first-order approximation in simulation.

\begin{figure}[h]
    \centering
    \includegraphics[width=\linewidth]{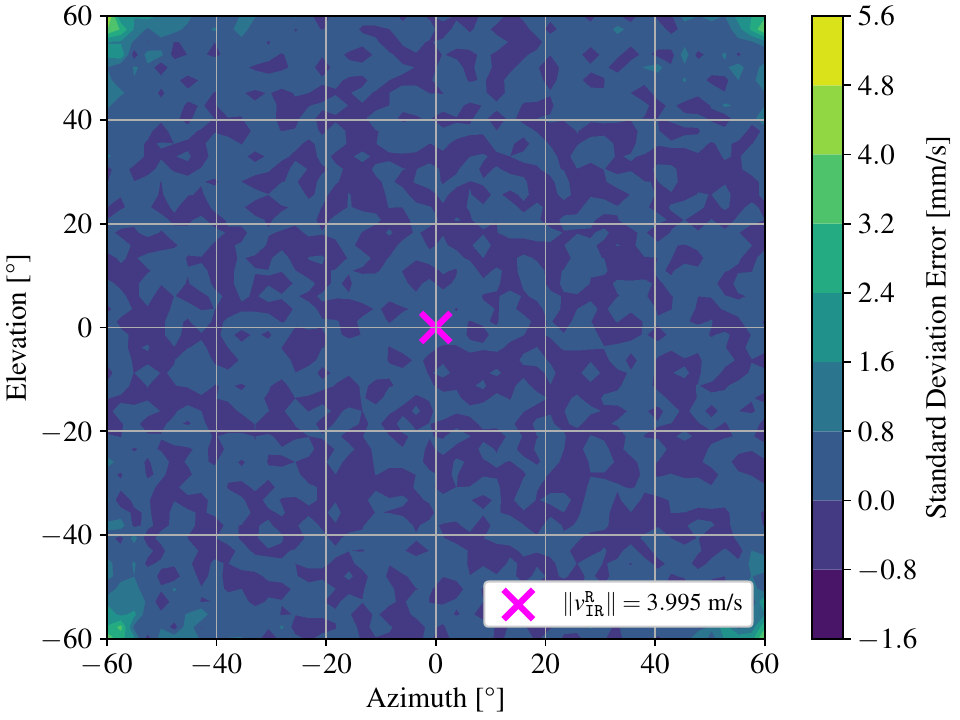}
    \spacingFigureCaption
    \caption{The error committed by applying a first-order approximation for the covariance propagation with the \rc{1} configuration assuming speed equal to the Doppler limit.}
    \label{fig:evaluation:noise:first_order}
\end{figure}
\begin{figure*}[t]
    \centering
    \subfloat[\label{fig:evaluation:noise:contour_lowSpeed}]{\includegraphics[width=0.49\linewidth]{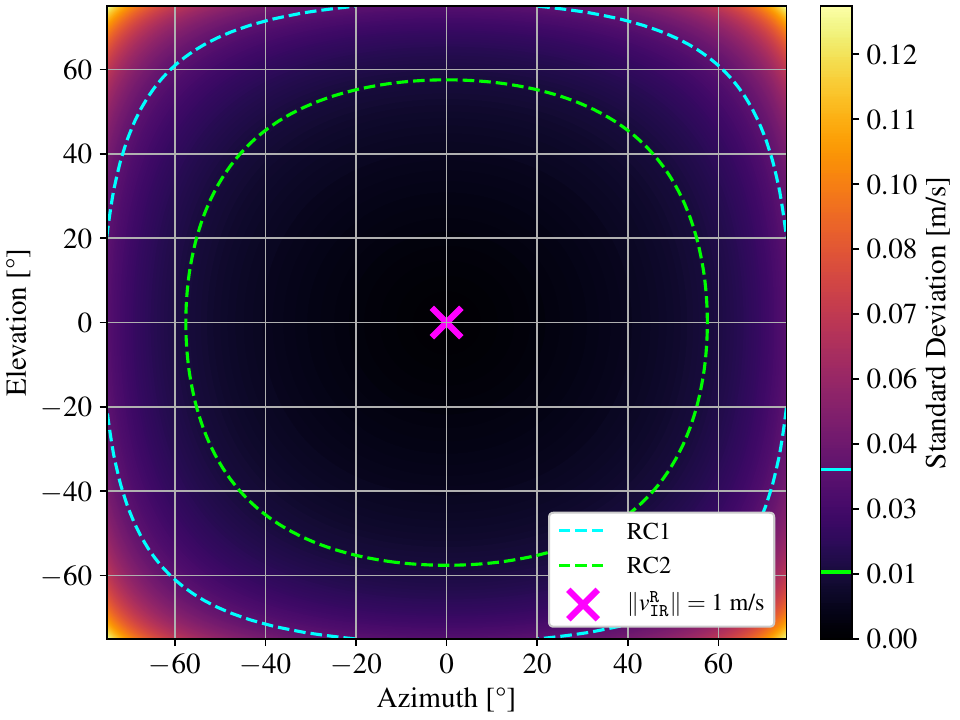}}
    \hfill
    \subfloat[\label{fig:evaluation:noise:contour_highSpeed:lar}]{\includegraphics[width=0.49\linewidth]{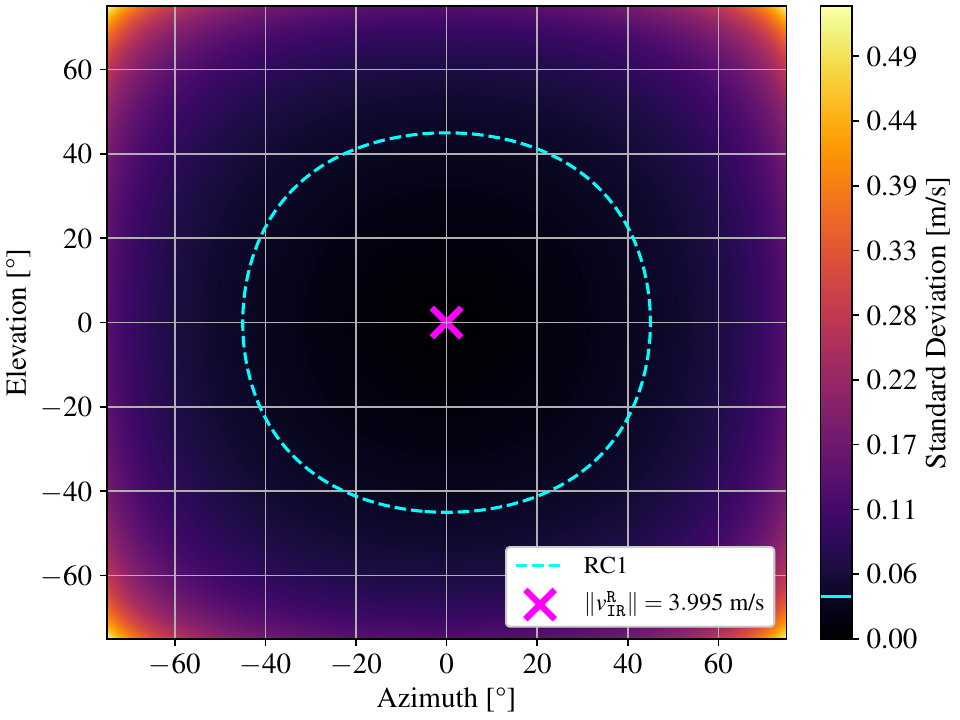}}
    \hfill
    \subfloat[\label{fig:evaluation:noise:contour_highSpeed:har}]{\includegraphics[width=0.49\linewidth]{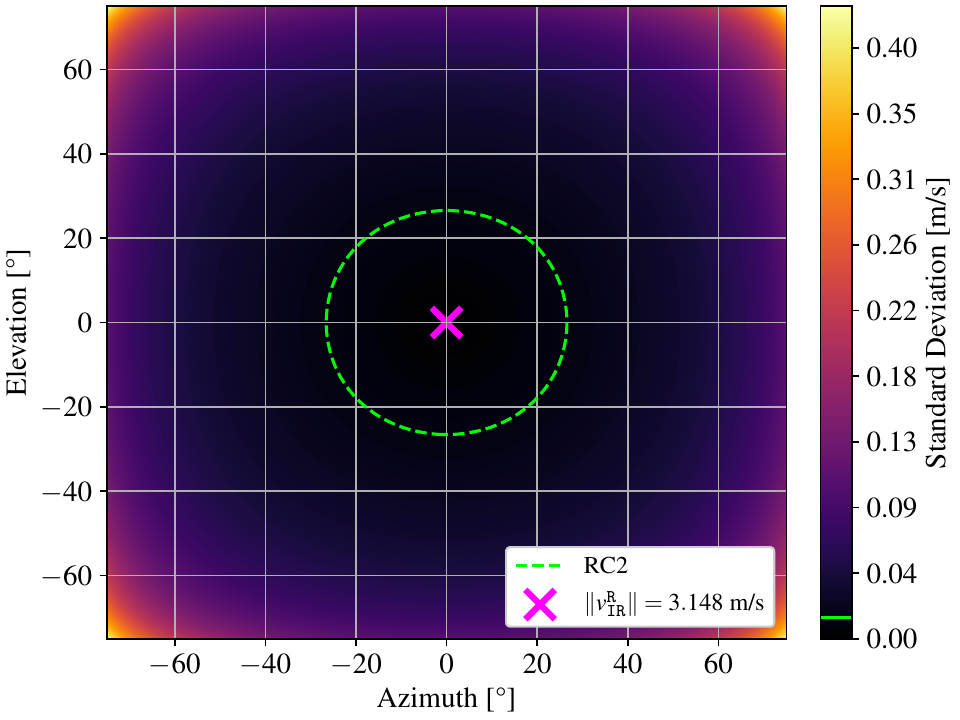}}
    \hfill
    \subfloat[\label{fig:evaluation:noise:contour_weirdSpeed}]{\includegraphics[width=0.49\linewidth]{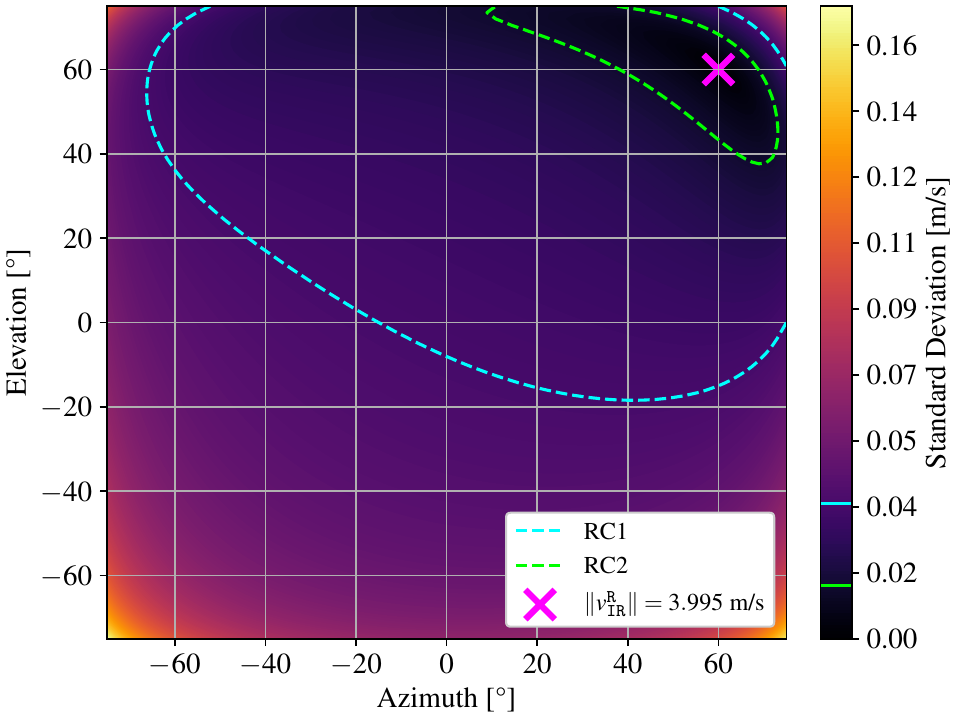}}
    \caption{Doppler residual standard deviation resulting from the \ac{aoa} uncertainty model as distributed across the sensor \ac{fov} given velocities of different magnitudes and directions, visualized for both chirp configurations from \cref{tab:evaluation:chirp}. The \rc{1} and \rc{2} contours designate where the noise contributed from Doppler and \ac{aoa} are equivalent. Note, the correlation between higher speeds and the greater importance of \ac{aoa} uncertainty compared to Doppler. Note also the warped shape of the equivalent noise level set as a function of the ego-velocity direction.}
    \label{fig:evaluation:noise:contour}
\end{figure*}

This is done by uniformly sampling phase angles across the beam width of the radar \ac{fov}. These values are corrupted by uniformly distributed noise matching the quantization magnitude. Taking the \rc{1} configuration as reference, noisy Doppler values are calculated, assuming forward velocity at the Doppler limit, and compared with the equivalent approximation. The results of this simulation are depicted in \cref{fig:evaluation:noise:first_order}, where it can be seen that the absolute difference is small (\SI{<6}{\milli\meter\per\second}). For most of the \ac{fov}, the difference is even smaller (\SI{<1}{\milli\meter\per\second}), increasing only at the corners where the effects of the nonlinearities are more pronounced. In short, the linearization does not seem to result in meaningful errors.

\subsubsection{Correlation with Speed and \acl{aoa}}\label{sec:evaluation:noise:correlation}
The relationship between phase angles and the resulting \ac{aoa} and bearing vector is clearly non-linear (see \cref{eq:method:wy,eq:method:wz,eq:method:bearing_from_phase}). As a result, \ac{aoa}-dependent behavior can be observed. Such effects, and their relationship to the Doppler residual accuracy, are studied here in simulation.

The noise contributed to the Doppler residual from the phase angle uncertainty is simulated by uniformly sampling azimuth and elevation angles. The first-order approximation of the resulting covariance is thus calculated according to \cref{eq:method:noise:bearing}. Following \cref{eq:method:residual:doppler:covariance}, the noise contributions from Doppler and bearing measurements contribute independently. Since the Doppler measurement noise is not \ac{aoa} dependent, its contribution can be separated. Thus, we can highlight the conditions under which the \ac{aoa} noise becomes significant in comparison to the Doppler noise, according to the first-order approximation.

In \cref{fig:evaluation:noise:contour_lowSpeed}, the bearing measurement noise contribution is visualized over the radar's \ac{fov}, with a marked contour to represent the Doppler measurement contribution. In such a visualization, measurements within the Doppler contour have greater noise contributions from the Doppler measurements than the \ac{aoa} measurements. For measurements outside the Doppler contour, this is reversed.  Included in the visualization are both \rc{1} and \rc{2} configurations, assuming forward-flight at a speed of \SI{1}{\meter\per\second}. Note, the \ac{fov} covered here is greater than \SI{\pm 60}{\degree} for clarity of the Doppler noise contribution boundary. Following the analysis of \cref{sec:evaluation:noise}, the smaller Doppler bin width of \rc{2} results in higher significance of the angle noise. However, \cref{eq:method:residual:doppler:covariance} shows correlation with linear velocity as well as \ac{aoa}. In \cref{fig:evaluation:noise:contour_highSpeed:lar,fig:evaluation:noise:contour_highSpeed:har}, the same simulation is repeated instead for the max Doppler speed of each configuration. Unsurprisingly, the noise-equal-level-set is found at a lower \ac{aoa} than before, due to the velocity magnitude acting as a scaling factor for the uncertainty. Velocity direction also plays a role, as seen in \cref{fig:evaluation:noise:contour_weirdSpeed} for \SI{\sim11}{\meter\per\second}. The region where Doppler noise dominates is warped when the radar frame velocity is not directed straight ahead. All in all, the measurement quality is highly dependent on the \ac{aoa} as well as the radar-frame velocity, both in terms of direction and magnitude. Furthermore, it can be seen, especially in \cref{fig:evaluation:noise:contour_highSpeed:har}, that the bearing noise contribution can dominate over the Doppler noise as a function of the configuration accuracy, speed, and velocity direction.

\subsection{Model Validation}\label{sec:evaluation:validation}
In this section, analysis is done to confirm that the theoretical models for the different radar measurements correspond well to the behavior observed from empirical data. These experiments are conducted with the uRAD Automotive radar (using configuration \rc{3}) and a custom-made \SI{75}{\milli\meter} side-length aluminum corner reflector, both situated in a Qualisys motion capture arena. This particular radar is used for its simplicity with respect to alignment, and the corner reflector serves as a consistent, high \ac{snr} radar target. Ground truth for these experiments is generated from the motion-capture tracking of the radar and corner reflector poses. Note that these experiments are not performed in any type of anechoic chamber; as a result, significant noise pollution of the results is to be expected, originating from the non-corner-reflector reflections and interference. Furthermore, because the radar's transmitted signal penetrates surfaces, this noise is not confined to the testing room. Furthermore, due to fabrication inconsistencies in such radars, there can be errors in range and \ac{aoa} measuring. We have performed calibrations to address such errors; however, because of the lack of such a controlled environment, noise can also contaminate the calibrations. Accordingly, these results serve mostly to qualitatively reason about the sensor's performance.

The experiments themselves consist of two varieties. The first wherein the corner reflector and radar are situated on a linear rail to maintain their relative attitudes. The radar is mounted statically on the rail, and the corner reflector is mounted on a movable cart, controlled by a stepper motor and pulley drivetrain. This drivetrain is controlled by an Arduino Uno microcontroller. The second setup consists of two tripods: one with a pan-and-tilt head and the other static. The radar is mounted on the actuatable tripod, so the relative azimuth and elevation of the corner reflector can be adjusted and maintained.

In order to create measurements for range and Doppler, the linear rail setup is used. The stepper motor drives the corner reflector, forwards and backwards, along the rail at fixed speeds from approximately \qtyrange{0.17}{0.67}{\meter\per\second}. Care was taken to capture data only during the fixed speed portions, as the phases of acceleration and deceleration were not always smooth due to friction on the rail and slack in the drivetrain. The corner reflector can then be isolated from the radar point cloud using motion-capture tracking and used to analyze Doppler and range measurement statistics. The results from this are shown in \cref{tab:evaluation:validation}, where the theoretical and empirical values match closely.

In order to analyze the \ac{aoa} phase behavior, the tripod setup was used. Azimuth and elevation experiments were conducted separately, with the radar manually positioned at a specific azimuth and elevation angle relative to the corner reflector. These were done in discrete steps from approximately \SI{-3}{\degree} to \SI{3}{\degree}. Once the radar orientation was locked in place, data was collected from both the radar and motion capture systems. As before, the motion capture tracking of the radar and corner reflector poses was used to select the corner reflector from the radar point cloud. From this, the measurement statistics can be calculated, and as seen in \cref{tab:evaluation:validation}, there is again close matching between the empirical and theoretical values. Here, the mean error for the vertical phase is slightly higher than in the other experiments. As these experiments could not be conducted in a radar-reflection-free environment, interference or latent calibration errors will contaminate the results. This contamination would then cause deviations from the theoretical models, as seen in the vertical phase mean result. Thus, these results are of mostly qualitative value.

\begin{table}[h]
    \centering
    \caption{Comparing Theoretical and Empirical Measurement Statistics}
    \label{tab:evaluation:validation}
    \sisetup{
    round-mode=places,
    round-precision=3,
    round-pad=true,
    table-format=1.3,
    scientific-notation=false,
}%
\newcommand{\vmove}{-0.675ex}%
\begin{tabular}{lSSSSl}
    \toprule
    \multirow{2}{*}[\vmove]{Measurement}    &\multicolumn{2}{c}{Empirical}   &\multicolumn{2}{c}{Theoretical}   &\multirow{2}{*}[\vmove]{Unit}\\\cmidrule(lr){2-3}\cmidrule(lr){4-5}
        &{Mean}   &{Std. Dev.}  &{Mean}   &{Std. Dev.}\\
    \midrule
    Doppler             &\num{0.00015088341701308667}   &\num{0.017375344104114857}   &\num{0}   &\num{0.017496419485832488}    &\si{\meter\per\second}\\
    Range               &\num{0.003989032866451474} &\num{0.059404706672109855}   &\num{0}   &\num{0.05634285687738268} &\si{\meter}\\
    Horizontal Phase    &\num{-0.023187055295273096}   &\num{1.6262833993771102}   &\num{0}   &\num{1.6237976320958225} &\si{\degree}\\
    Vertical Phase      &\num{0.7904172914514711}   &\num{1.6170916801050983}   &\num{0}   &\num{1.6237976320958225} &\si{\degree}\\
    \bottomrule
\end{tabular}
\end{table}

\subsection{Experiments}\label{sec:evaluation:flight}
\begin{figure*}[t!]
    \newcommand{\width}{0.33\linewidth}
    \centering
    \subfloat[\label{fig:evaluation:flight:trajectories:helix}]{\includegraphics[width=\width]{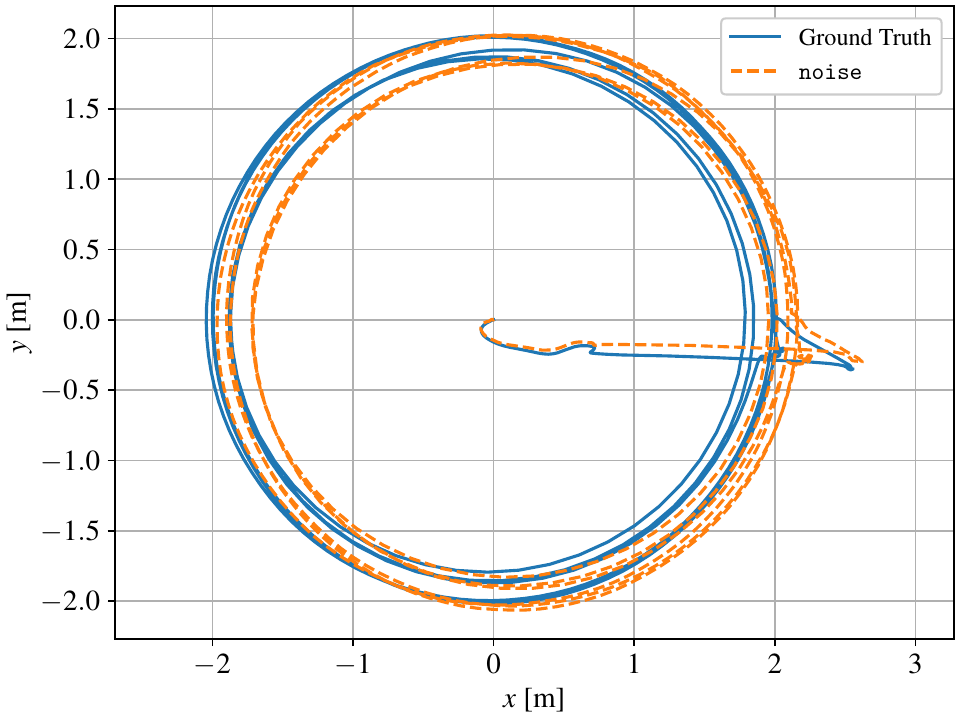}}
    \hfill
    \subfloat[\label{fig:evaluation:flight:trajectories:rectangle}]{\includegraphics[width=\width]{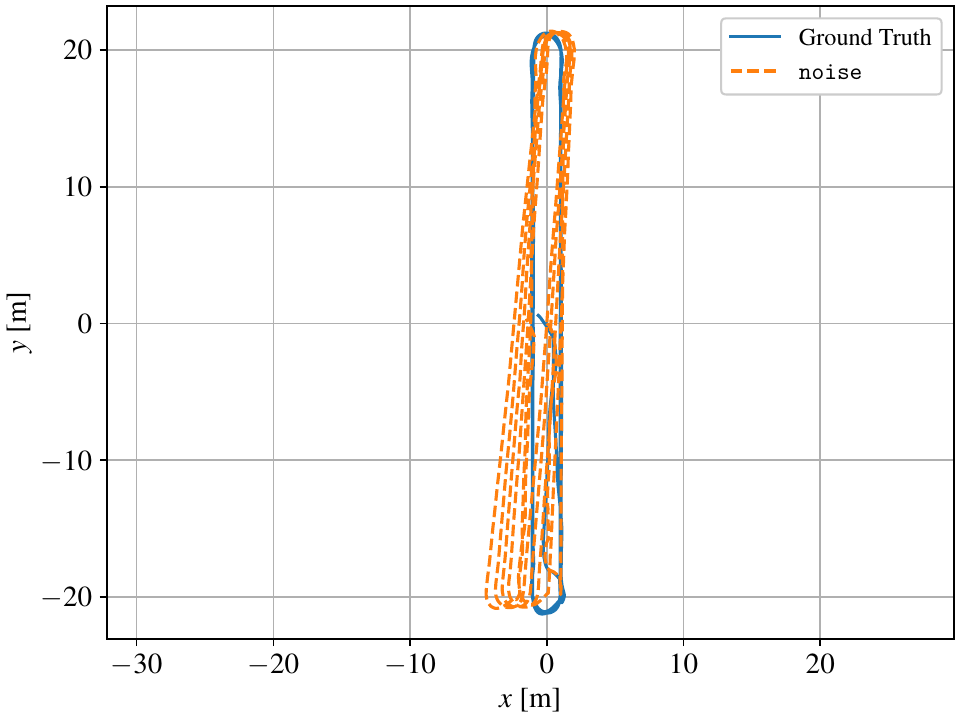}}
    \hfill
    \subfloat[\label{fig:evaluation:flight:trajectories:handheld2}]{\includegraphics[width=\width]{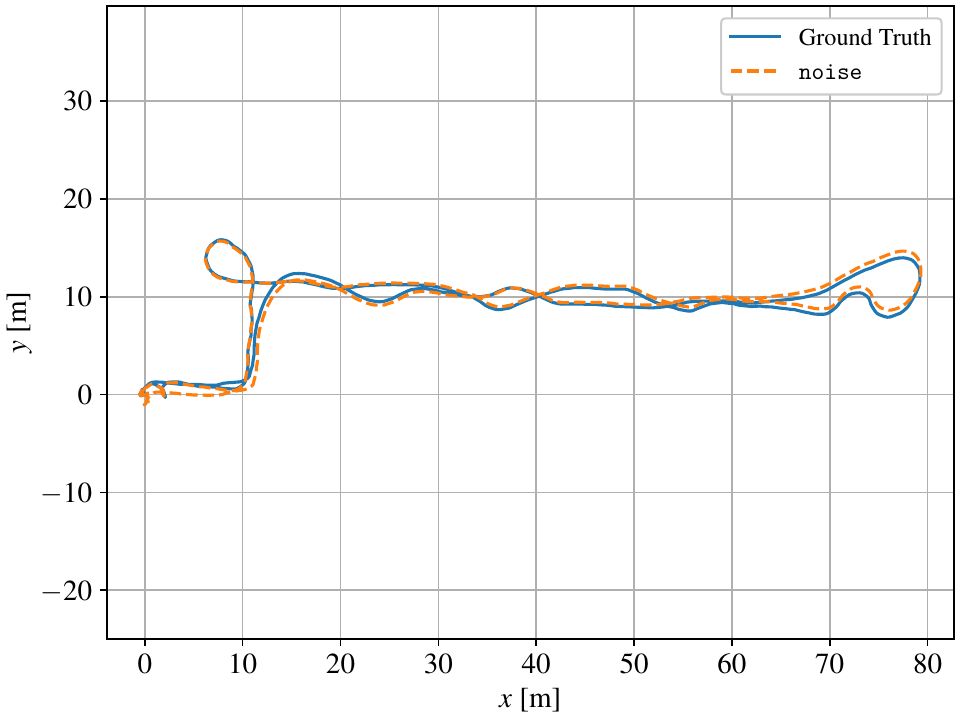}}
    \hfill
    \subfloat[\label{fig:evaluation:flight:trajectories:lemniscate}]{\includegraphics[width=\width]{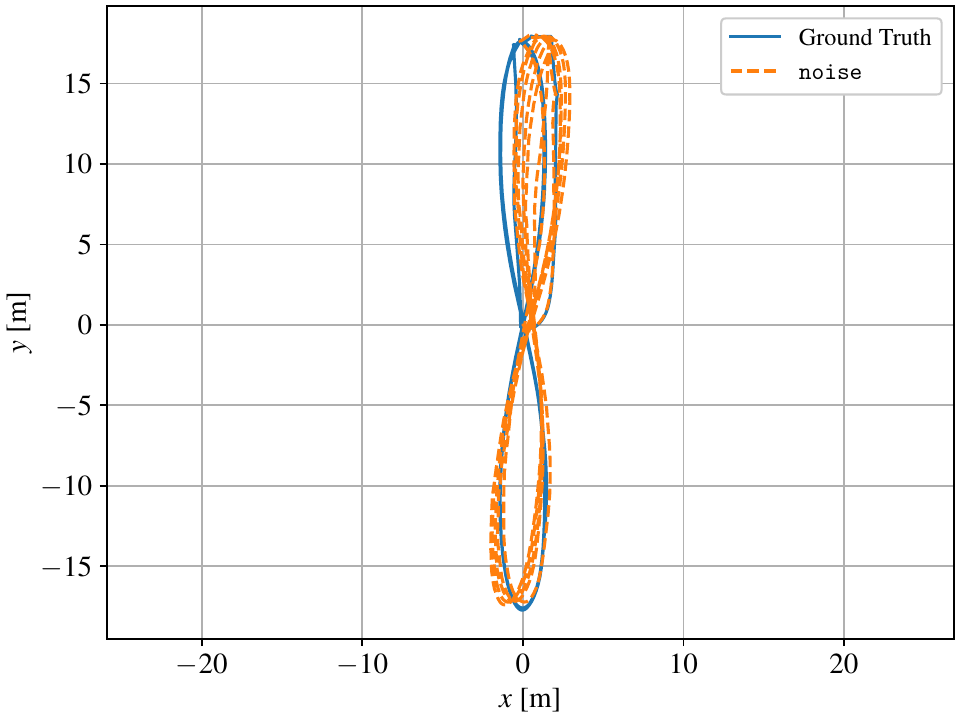}}
    \hfill
    \subfloat[\label{fig:evaluation:flight:trajectories:line}]{\includegraphics[width=\width]{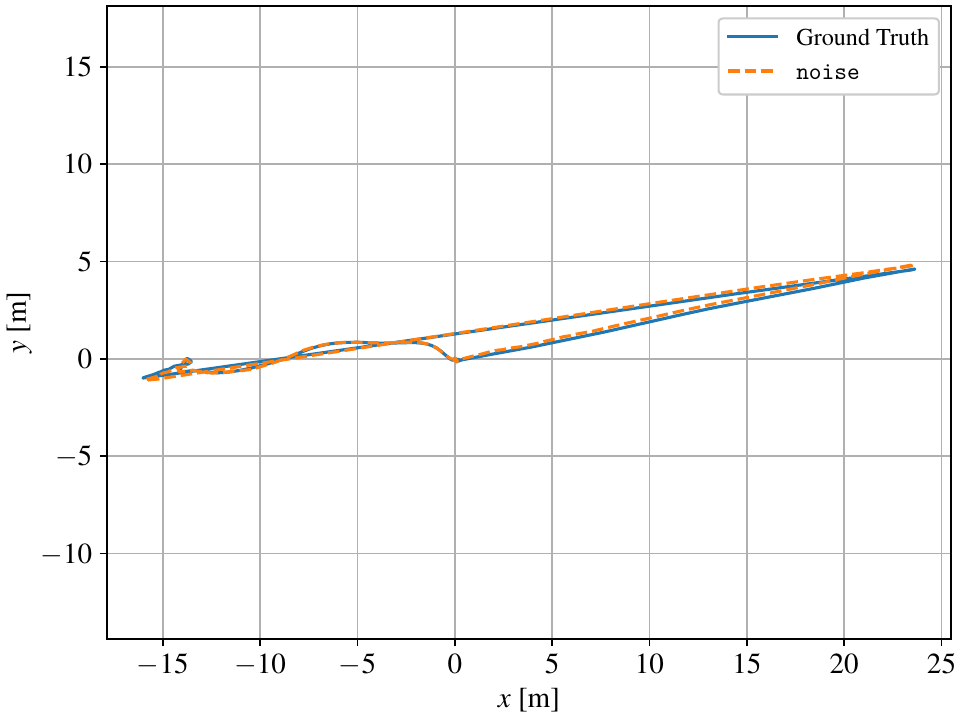}}
    \hfill
    \subfloat[\label{fig:evaluation:flight:trajectories:manual1}]{\includegraphics[width=\width]{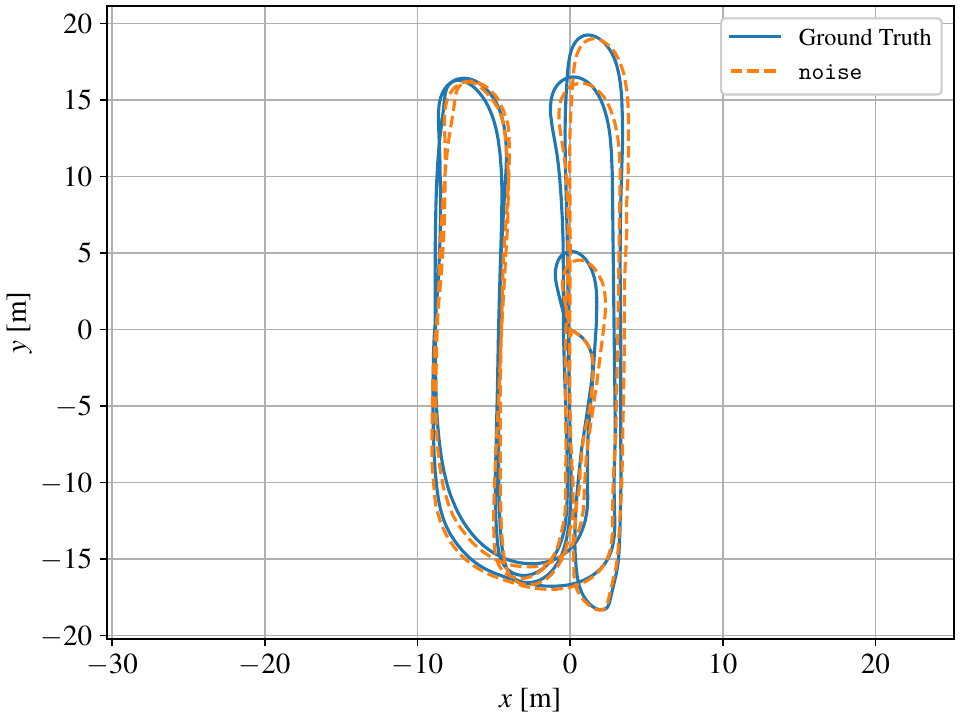}}
    \hfill
    \subfloat[\label{fig:evaluation:flight:trajectories:manual2}]{\includegraphics[width=\width]{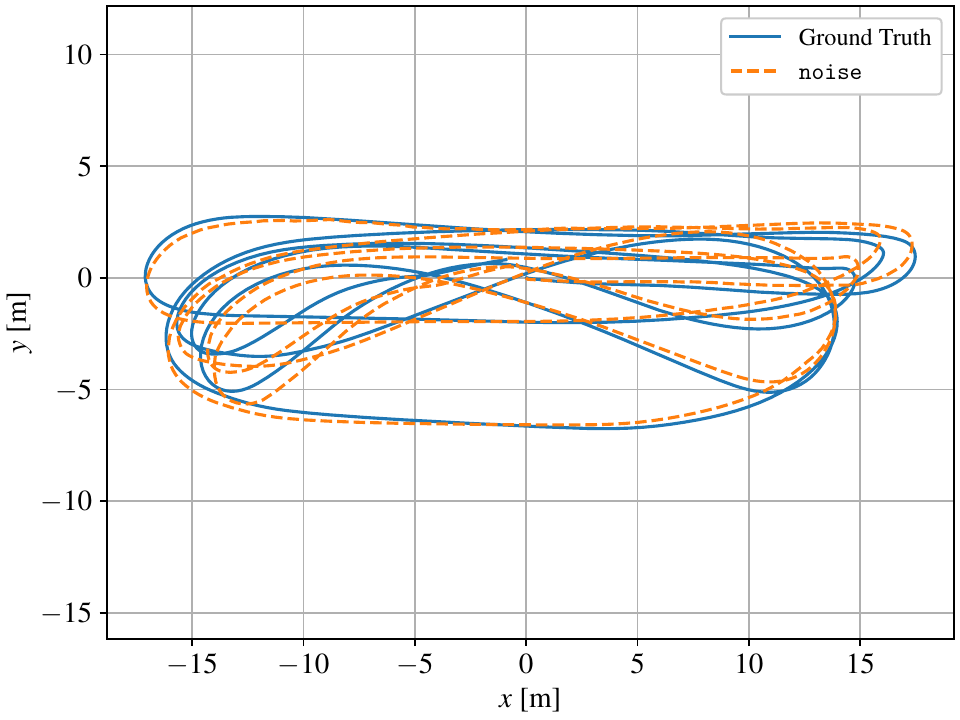}}
    \hfill
    \subfloat[\label{fig:evaluation:flight:trajectories:manual3}]{\includegraphics[width=\width]{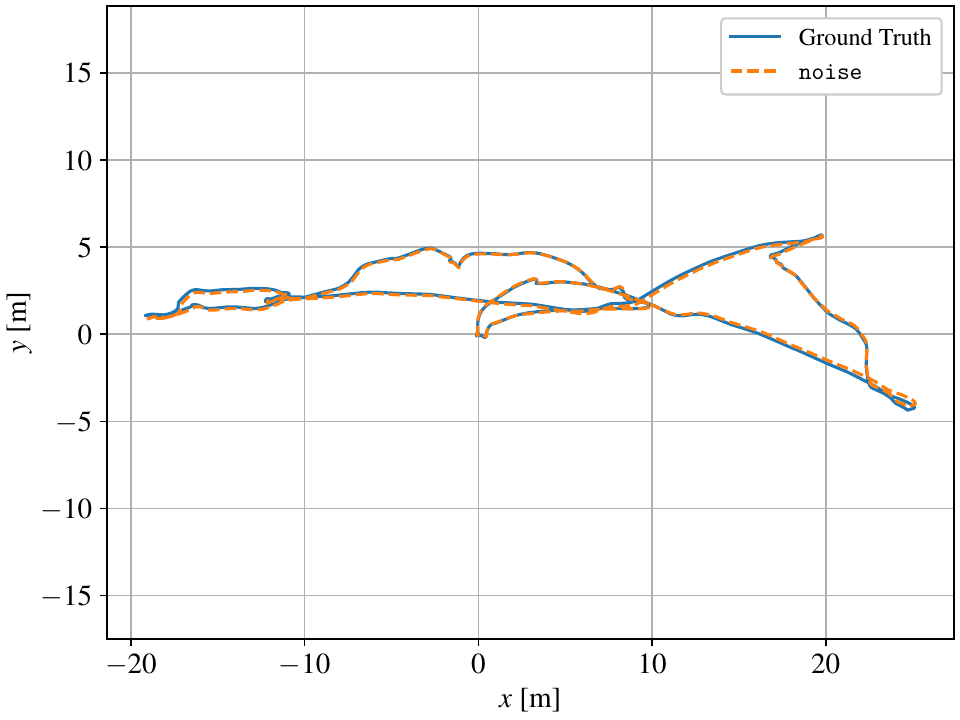}}
    \hfill
    \subfloat[\label{fig:evaluation:flight:trajectories:manual4}]{\includegraphics[width=\width]{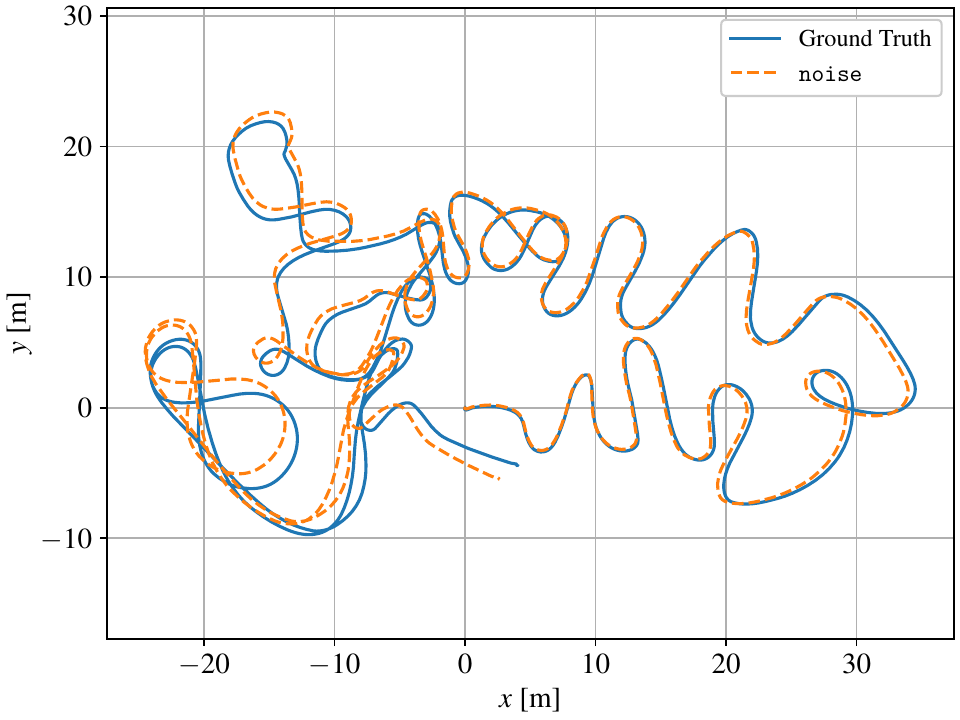}}
    \vfill
    \hfill
    \subfloat[\label{fig:evaluation:flight:trajectories:manual8}]{\includegraphics[width=\width]{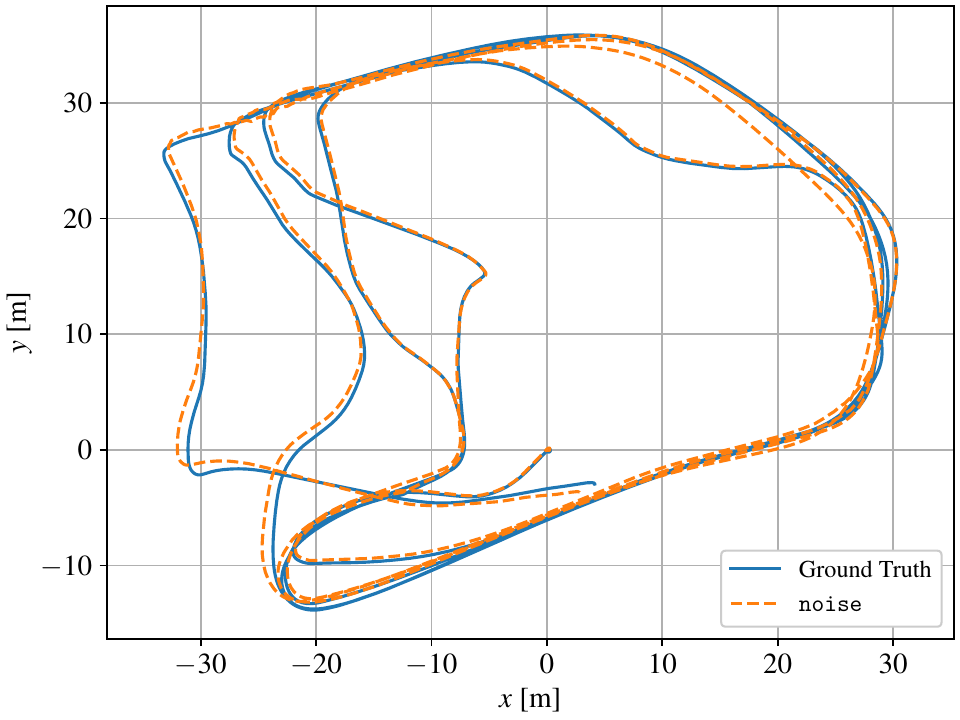}}
    \hfill
    \subfloat[\label{fig:evaluation:flight:trajectories:manual11}]{\includegraphics[width=\width]{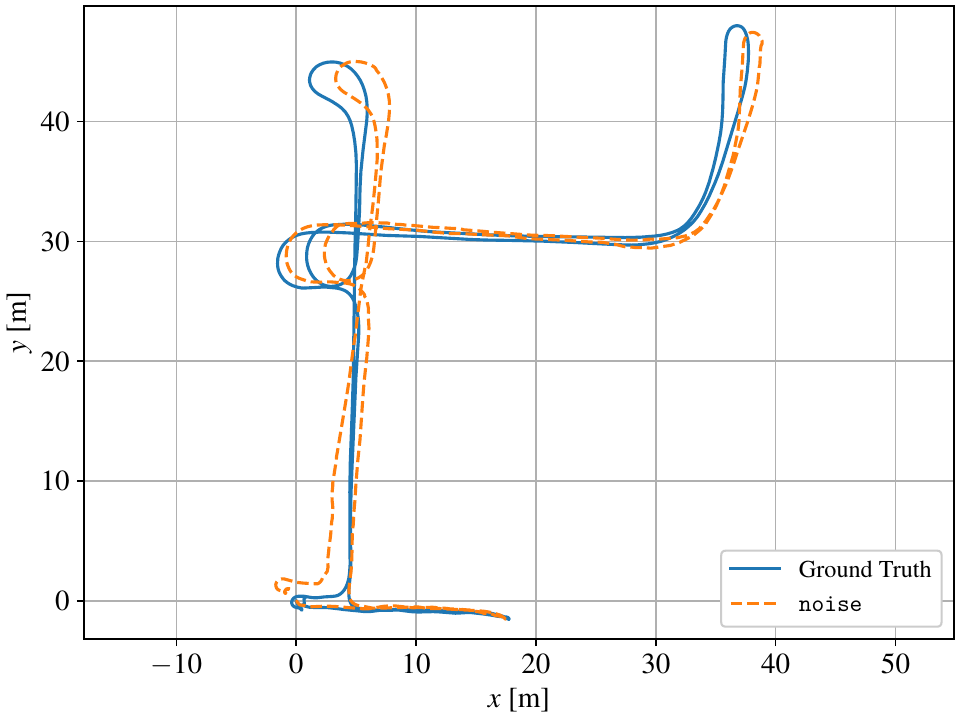}}
    \hfill
    \caption{Position estimates of the proposed method (with \config{noise} configuration) against ground truth for a representative set of trajectories: (a) \experiment{hel\_4mps\_align}, (b) \experiment{rec\_4mps\_align}, (c) \experiment{handheld\_2}, (d) \experiment{lem\_4mps\_align}, (e) \experiment{flight\_11mps}, (f) \experiment{manual\_1}, (g) \experiment{manual\_2}, (h) \experiment{manual\_3}, (i) \experiment{manual\_4}, (j) \experiment{manual\_8}, and (k) \experiment{manual\_11}.}
    \label{fig:evaluation:flight:trajectories}
\end{figure*}

The robotic experiment datasets will address the following questions: the impact of Doppler accuracy on estimator performance across different radar configurations, and the operational limits of the estimator with respect to different environments and speeds. A representative set of the trajectories studied in this section is shown in \cref{fig:evaluation:flight:trajectories}, and an overview of the experiment's characteristics is presented in \cref{tab:evaluation:flight:overview}.

\begin{table}[ht!]
    \centering
    \caption{Overview of the Datasets Used for Evaluation}
    \label{tab:evaluation:flight:overview}
    \begin{tabular}{
	ll
	S[table-format=3.0]
	S[table-format=2.2]
	S[table-format=3.0]
}
\toprule
	&	&	&\multicolumn{2}{c}{Maximum Speed}\\\cmidrule{4-5}
\makecell[l]{Sequence}	&\makecell[l]{Radar\\Chirp}	&{\makecell{Length\\{[\si{\meter}]}}}	&{\makecell{Linear\\{[\si{\meter\per\second}]}}}	&{\makecell{Angular\\{[\si{\degree\per\second}]}}}\\
\midrule
\Corridor{}	&	&	&	&\\
\texttt{hel\_3mps\_align}	&\rc{2}	&88	&2.34	&105\\
\texttt{hel\_4mps\_align}	&\rc{1}	&90	&2.34	&138\\
\texttt{rec\_3mps\_align}	&\rc{2}	&457	&3.08	&132\\
\texttt{rec\_3mps\_const}	&\rc{2}	&457	&3.07	&103\\
\texttt{rec\_4mps\_align}	&\rc{1}	&479	&4.00	&207\\
\texttt{rec\_4mps\_const}	&\rc{1}	&457	&3.96	&229\\
\texttt{handheld\_1}	&\rc{4}	&201	&1.46	&117\\
\texttt{handheld\_2}	&\rc{4}	&211	&1.46	&155\\
\midrule
\Gym{}	&	&	&	&\\
\texttt{lem\_3mps\_align}	&\rc{2}	&403	&3.10	&258\\
\texttt{lem\_3mps\_const}	&\rc{2}	&400	&3.07	&198\\
\texttt{lem\_4mps\_align}	&\rc{1}	&404	&4.00	&166\\
\texttt{lem\_4mps\_const}	&\rc{1}	&400	&3.95	&226\\
\texttt{manual\_1}	&\rc{1}	&343	&5.80	&79\\
\texttt{manual\_2}	&\rc{1}	&339	&6.75	&114\\
\midrule
\Mine{}	&	&	&	&\\
\texttt{flight\_4mps}	&\rc{1}	&92	&4.49	&219\\
\texttt{flight\_5mps}	&\rc{1}	&89	&5.11	&165\\
\texttt{flight\_6mps}	&\rc{1}	&86	&6.14	&109\\
\texttt{flight\_7mps}	&\rc{1}	&90	&7.10	&220\\
\texttt{flight\_8mps}	&\rc{1}	&84	&7.99	&152\\
\texttt{flight\_9mps}	&\rc{1}	&85	&8.90	&94\\
\texttt{flight\_10mps}	&\rc{1}	&87	&9.97	&172\\
\texttt{flight\_11mps}	&\rc{1}	&86	&10.80	&153\\
\texttt{manual\_3}	&\rc{1}	&139	&1.95	&86\\
\midrule
\Forest{}	&	&	&	&\\
\texttt{manual\_4}	&\rc{1}	&453	&4.05	&142\\
\texttt{manual\_5}	&\rc{1}	&551	&4.40	&153\\
\texttt{manual\_6}	&\rc{1}	&482	&3.58	&134\\
\texttt{manual\_7}	&\rc{1}	&521	&4.53	&150\\
\texttt{manual\_8}	&\rc{1}	&779	&8.73	&148\\
\midrule
\Basement{}	&	&	&	&\\
\texttt{manual\_9}	&\rc{3}	&174	&2.37	&116\\
\texttt{manual\_10}	&\rc{3}	&205	&3.47	&108\\
\texttt{manual\_11}	&\rc{3}	&252	&3.74	&69\\
\bottomrule
\end{tabular}
% total length: 8981.95857348472

\end{table}

To increase consistency across experiments investigating the impact of different configurations, the aerial platform will follow fixed trajectories, using LiDAR-inertial odometry for online pose estimation. The trajectories in question are created by supplying waypoints to a non-minimum jerk optimization problem with penalties on duration and for exceeding velocity/acceleration limits, following~\cite{wang2021Trajectories}. This results in a smooth trajectory defined for position, velocity, and acceleration, which are queried at a fixed rate and sent to the quadrotor platform as setpoints. Setpoints for yaw $\psi_{\text{sp}}$ and yaw-rate $\dot{\psi}_{\text{sp}}$ can be calculated as

\begin{align}
        \psi_{\text{sp}} &= \tan^{-1}\left( \frac{y_{\text{sp}}}{x_{\text{sp}}} \right),\\
        \dot{\psi}_{\text{sp}} &= \frac{\dot{x}_{\text{sp}} \ddot{y}_{\text{sp}} - \ddot{x}_{\text{sp}} \dot{y}_{\text{sp}}}{\dot{x}_{\text{sp}}^2 + \dot{y}_{\text{sp}}^2},
\end{align}
given setpoints for the 2D positions $x_{\text{sp}}$ and $y_{\text{sp}}$, and their derivatives up to acceleration, resulting in trajectory-aligned yaw and corresponding yaw-rate setpoints.

This section will include comparisons of the position and attitude \ac{rmse} with \ac{ape} and \ac{rpe}, using \SI{10}{\meter} segment length, metrics calculated according to~\cite{sturm2012benchmark,grupp2017evo}. Comparisons include different configurations of the proposed method and relevant state-of-the-art methods, such as~\cite{doer2021xrio} (denoted \doer{}) and \cite{girod2024radar} (denoted \asl{}). Note that \doer{} is used without yaw aiding, and both \doer{} and \asl{} are used without the barometer, to highlight the radar-fusion aspects of the methods and make for a fair comparison. Since the proposed method and \asl{} both use GTSAM~\cite{gtsam}, the comparison is made with identical noise and iSAM2~\cite{kaess2011isam2} parameters. The \doer{} method does not perform well with the same noise parameters, due to differences in the back-end implementation; therefore, custom parameters are used. Additionally, \doer{} is modified for compatibility with our trigger timestamps. \asl{} is modified as well; the method expects initialization from a separate attitude estimator. Instead, we use the proposed method's initialization routine. Additionally, basic filtering for radar points, matching that of the proposed method, is added. Without this, the \asl{} method creates false-positive point associations on the body of the robotic platforms, leading to estimation failure in many of the experiments.

\subsubsection{Significance of Doppler Accuracy}\label{sec:evaluation:flight:doppler_accuracy}
This section will analyze the impact of the noise model similarly to the studies conducted in \cref{sec:evaluation:noise:doppler_accuracy}, and how this impact changes with varying levels of Doppler accuracy from the chirp configuration. As such, this study will examine flight experiments conducted with the aerial platform flying in position control against pre-computed trajectories in the \corridor{} and \gym{} environments. 

The experiments are designed with the goal of flying as close to the Doppler limit as possible for as long as possible in a controlled and repeatable manner. For this purpose, three trajectories are created: the Rounded Rectangle (abbreviated as \experiment{rec}), Lemniscate (abbreviated as \experiment{lem}), and Helix (abbreviated as \experiment{hel}) with representative examples shown in \cref{fig:evaluation:flight:trajectories:rectangle,fig:evaluation:flight:trajectories:lemniscate,fig:evaluation:flight:trajectories:helix}. From \cref{sec:evaluation:noise:correlation}, it can be seen that the direction of the velocity has a significant role in the impact of the noise model. In order to address this, permutations of the Rounded Rectangle and Lemniscate trajectories are created, one with trajectory-aligned yaw (denoted \experiment{align}) and another with constant yaw setpoints (denoted \experiment{const}). In addition, all three trajectories will be repeated with both radar chirp configurations from \cref{tab:evaluation:chirp}, in order to facilitate a comparison of the benefits for varying levels of Doppler accuracy.

\begin{table}[h!]
    \centering
    \caption{Ablation of Method Configurations for Position \ac{rmse}}
    \label{tab:evaluation:flight:doppler_accuracy1}
    \begin{tabular}{@{}lcc@{}}
\toprule
\multirow{2}{*}{Sequence}	&\multicolumn{2}{c}{\texttt{base} -- \texttt{noise} -- \texttt{geometry}}\\
	&{APE [\si{\meter}]}	&{RPE ($\Delta:\SI{10}{\meter}$) [\si{\meter}]}\\
\midrule
\Corridor{}	&	&\\
\texttt{hel\_3mps\_align}	&0.544 -- 0.529 -- \textbf{0.515}	&0.221 -- 0.201 -- \textbf{0.178}\\
\texttt{hel\_4mps\_align}	&0.416 -- 0.389 -- \textbf{0.084}	&0.153 -- 0.152 -- \textbf{0.068}\\
\texttt{rec\_3mps\_align}	&3.031 -- \textbf{2.112} -- 2.122	&0.292 -- 0.257 -- \textbf{0.253}\\
\texttt{rec\_3mps\_const}	&2.446 -- \textbf{2.342} -- 3.342	&0.275 -- \textbf{0.266} -- 0.288\\
\texttt{rec\_4mps\_align}	&1.239 -- 1.147 -- \textbf{0.504}	&0.181 -- 0.180 -- \textbf{0.175}\\
\texttt{rec\_4mps\_const}	&1.334 -- 1.238 -- \textbf{0.447}	&0.218 -- \textbf{0.215} -- 0.264\\
\midrule
\Gym{}	&	&\\
\texttt{lem\_3mps\_align}	&1.671 -- \textbf{0.885} -- 1.752	&0.397 -- \textbf{0.339} -- 0.355\\
\texttt{lem\_3mps\_const}	&0.654 -- 0.621 -- \textbf{0.349}	&0.184 -- \textbf{0.178} -- 0.190\\
\texttt{lem\_4mps\_align}	&\textbf{0.915} -- 0.955 -- 1.246	&0.264 -- \textbf{0.262} -- 0.278\\
\texttt{lem\_4mps\_const}	&0.958 -- 0.952 -- \textbf{0.607}	&0.309 -- \textbf{0.308} -- 0.309\\
\bottomrule
\end{tabular}

\end{table}

\begin{table}[h!]
    \centering
    \caption{Ablation of Method Configurations for Attitude \ac{rmse}}
    \label{tab:evaluation:flight:doppler_accuracy3}
    \begin{tabular}{@{}lcc@{}}
\toprule
\multirow{2}{*}{Sequence}	&\multicolumn{2}{c}{\texttt{base} -- \texttt{noise} -- \texttt{geometry}}\\
	&{APE [\si{\degree}]}	&{RPE ($\Delta:\SI{10}{\meter}$) [\si{\degree}]}\\
\midrule
\Corridor{}	&	&\\
\texttt{hel\_3mps\_align}	&1.075 -- 1.075 -- \textbf{0.965}	&1.532 -- 1.533 -- \textbf{1.470}\\
\texttt{hel\_4mps\_align}	&0.258 -- \textbf{0.257} -- 0.502	&0.379 -- 0.378 -- \textbf{0.330}\\
\texttt{rec\_3mps\_align}	&2.133 -- 2.207 -- \textbf{1.072}	&\textbf{0.953} -- 0.955 -- 0.959\\
\texttt{rec\_3mps\_const}	&2.046 -- 2.134 -- \textbf{0.663}	&0.501 -- 0.503 -- \textbf{0.493}\\
\texttt{rec\_4mps\_align}	&2.611 -- 2.665 -- \textbf{1.073}	&0.435 -- 0.436 -- \textbf{0.401}\\
\texttt{rec\_4mps\_const}	&1.221 -- 1.229 -- \textbf{0.663}	&0.648 -- \textbf{0.644} -- 0.712\\
\midrule
\Gym{}	&	&\\
\texttt{lem\_3mps\_align}	&3.697 -- 3.733 -- \textbf{3.469}	&1.029 -- \textbf{1.026} -- 1.056\\
\texttt{lem\_3mps\_const}	&1.842 -- 1.821 -- \textbf{0.660}	&0.505 -- 0.502 -- \textbf{0.483}\\
\texttt{lem\_4mps\_align}	&\textbf{3.281} -- 3.302 -- 5.691	&\textbf{1.306} -- 1.307 -- 1.433\\
\texttt{lem\_4mps\_const}	&3.181 -- 3.212 -- \textbf{0.879}	&0.701 -- 0.701 -- \textbf{0.646}\\
\bottomrule
\end{tabular}

\end{table}

Relevant results comparing the method's performance under \config{base} and \config{noise} configurations are presented in \cref{tab:evaluation:flight:doppler_accuracy1}. The same table also features the proposed method with the \config{geometry} configuration. From the table, it can be seen that adding the noise model generally improves the estimation performance, and that the improvement is greater for the \rc{2} experiments. Furthermore, adding the geometry-based factors serves to significantly reduce the \ac{ape} and slightly reduce \ac{rpe}. However, it can also be seen that the \rc{2} is not sufficiently dense for using geometry-based factors in \corridor{}. Furthermore, neither configuration is suitable for geometry-based factors in \gym{}, due to the environment being almost completely empty. Occasionally, significant improvements from the \config{geometry} configuration can also be seen in the attitude \ac{rmse} in \cref{tab:evaluation:flight:doppler_accuracy3}. As before, this can be inconsistent, due to the difficulties posed by the noisy radar point cloud. Such improvements are expected from registration-type factors; however, the number of outliers in the radar point cloud makes such an endeavor challenging in certain environments. Otherwise, insignificant differences can be seen between \config{base} and \config{noise}. This is generally unsurprising given that the proposed noise model has a necessarily limited effect on improving the missing yaw observability.

\begin{table}[h!]
    \centering
    \caption{Position \ac{rmse} for the Experiments from \cref{sec:evaluation:flight:doppler_accuracy}}
    \label{tab:evaluation:flight:doppler_accuracy2}
    \begin{tabular}{@{}lccc@{}}
\toprule
\multirow{2}{*}{Sequence}	&\multicolumn{3}{c}{APE $|$ RPE ($\Delta:\SI{10}{\meter}$) [\si{\meter}]}\\
	&{\texttt{noise}}	&{\texttt{x-RIO}~\cite{doer2021xrio}}	&{\texttt{ASL RIO}~\cite{girod2024radar}}\\
\midrule
\Corridor{}	&	&	&\\
\texttt{hel\_3mps\_align}	&\textbf{0.529} $|$ \textbf{0.201}	&1.111 $|$ 0.314	&1.171 $|$ 0.267\\
\texttt{hel\_4mps\_align}	&0.389 $|$ 0.152	&\textbf{0.179} $|$ 0.147	&0.246 $|$ \textbf{0.123}\\
\texttt{rec\_3mps\_align}	&\textbf{2.112} $|$ \textbf{0.257}	&6.613 $|$ 0.565	&6.351 $|$ 1.995\\
\texttt{rec\_3mps\_const}	&2.342 $|$ \textbf{0.266}	&\textbf{2.338} $|$ 0.450	&6.690 $|$ 2.135\\
\texttt{rec\_4mps\_align}	&1.147 $|$ \textbf{0.180}	&1.024 $|$ 0.229	&\textbf{0.831} $|$ 0.264\\
\texttt{rec\_4mps\_const}	&1.238 $|$ \textbf{0.215}	&\textbf{0.649} $|$ 0.264	&1.818 $|$ 0.315\\
\midrule
\Gym{}	&	&	&\\
\texttt{lem\_3mps\_align}	&\textbf{0.885} $|$ \textbf{0.339}	&22.918 $|$ 3.104	&30.365 $|$ 5.408\\
\texttt{lem\_3mps\_const}	&\textbf{0.621} $|$ \textbf{0.178}	&1.058 $|$ 0.464	&1.884 $|$ 0.203\\
\texttt{lem\_4mps\_align}	&\textbf{0.955} $|$ \textbf{0.262}	&4.281 $|$ 0.452	&2.569 $|$ 0.315\\
\texttt{lem\_4mps\_const}	&\textbf{0.952} $|$ \textbf{0.308}	&1.520 $|$ 0.371	&1.466 $|$ 0.324\\
\bottomrule
\end{tabular}

\end{table}

Comparisons with state-of-the-art methods in the same experiments are presented in \cref{tab:evaluation:flight:doppler_accuracy2}. In feature-rich environments (e.g., \corridor{}), both \cite{doer2021xrio,girod2024radar} can be seen performing well (except for some experiments where the point-matching for zero-velocity updates from~\cite{girod2024radar} causes divergence). On the other hand, the \gym{} environment features an open space with few discernible objects present, posing a different challenge to radar-based odometry, which has a limited maximum range and range resolution. With lesser geometric diversity, the point cloud size reduces, and as a result, the least squares velocity calculation of~\cite{doer2021xrio} can be under-determined or challenged with respect to observability. The point sparsity also proves challenging for \cite{girod2024radar}, resulting in estimation failure.

\begin{figure*}[t]
    \centering
    \subfloat[\label{fig:evaluation:flight:doppler_limit:fov:forward}]{\includegraphics[width=0.49\linewidth]{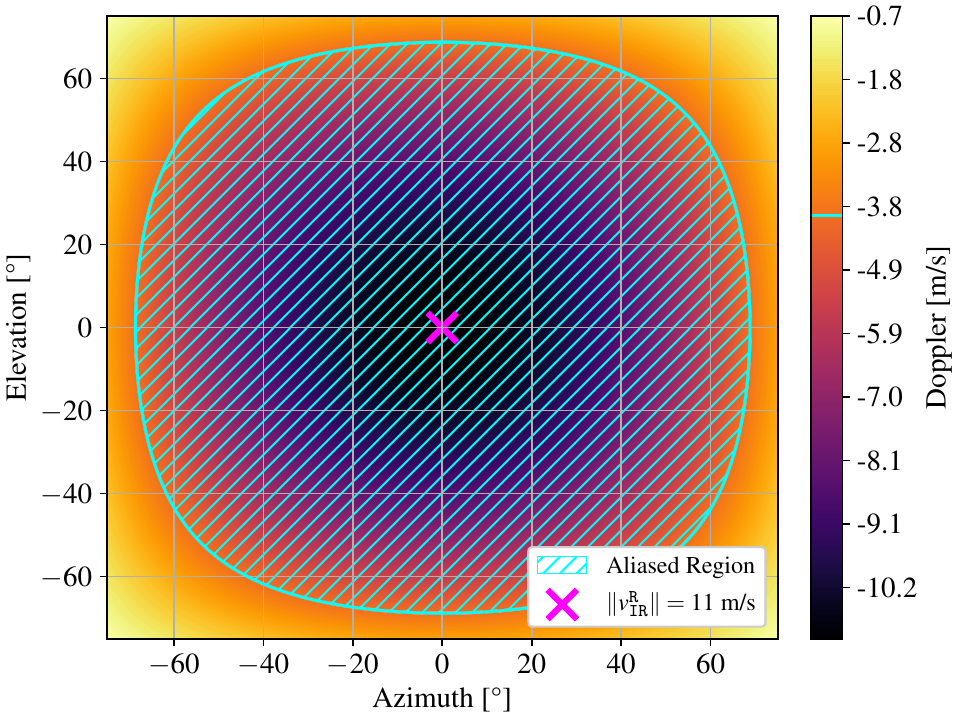}}
    \hfill
    \subfloat[\label{fig:evaluation:flight:doppler_limit:fov:upRight}]{\includegraphics[width=0.49\linewidth]{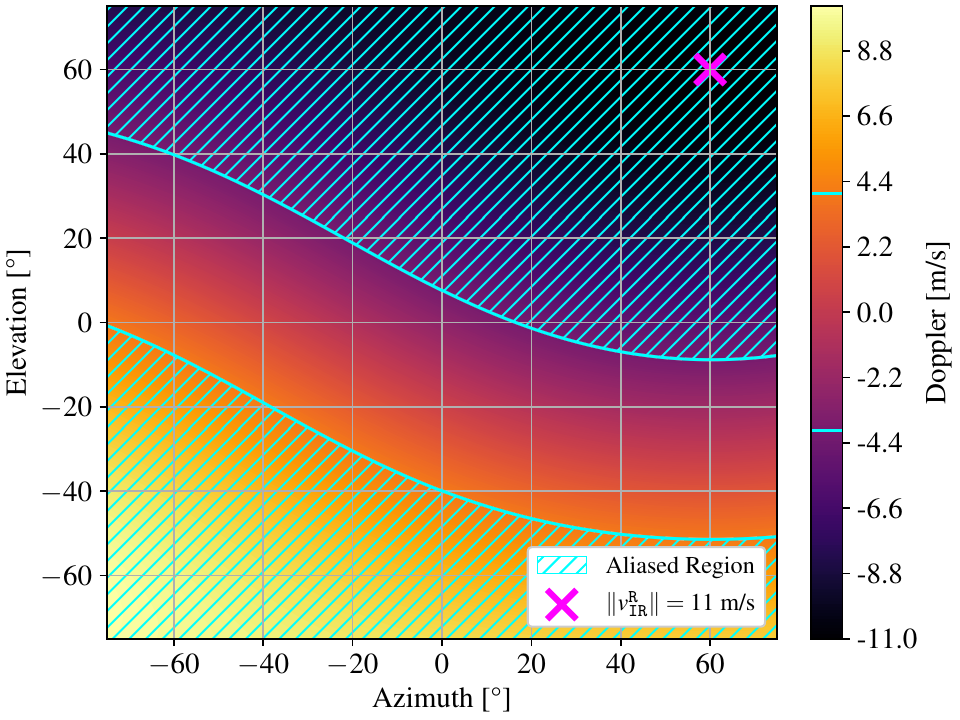}}
    \caption{Simulations of the radial speed at different \ac{aoa} across the \ac{fov} for \SI{11}{\meter\per\second} radar-frame velocities along different directions. Note the level-set of the \rc{1} Doppler limit marking the region boundary where aliasing would be occurring, and how the shape of this region can vary with respect to the direction of the radar-frame linear velocity.}
    \label{fig:evaluation:flight:doppler_limit:fov}
\end{figure*}
\begin{figure*}[t]
    \centering
    \subfloat[\label{fig:evaluation:flight:doppler_limit:flight_6mps}]{\includegraphics[width=0.49\linewidth]{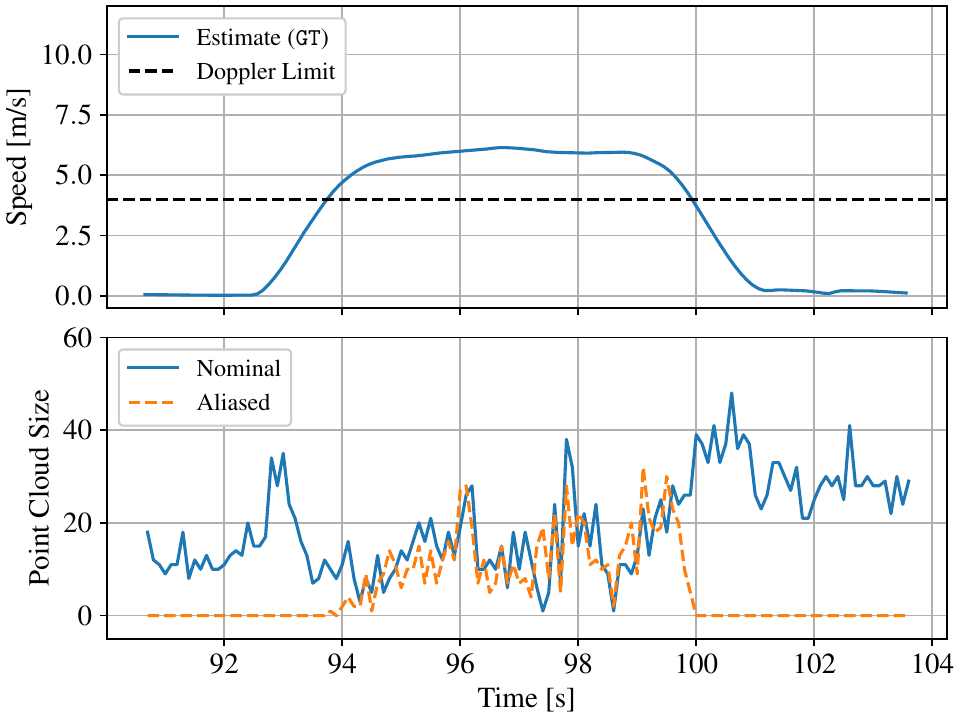}}
    \hfill
    \subfloat[\label{fig:evaluation:flight:doppler_limit:flight_11mps}]{\includegraphics[width=0.49\linewidth]{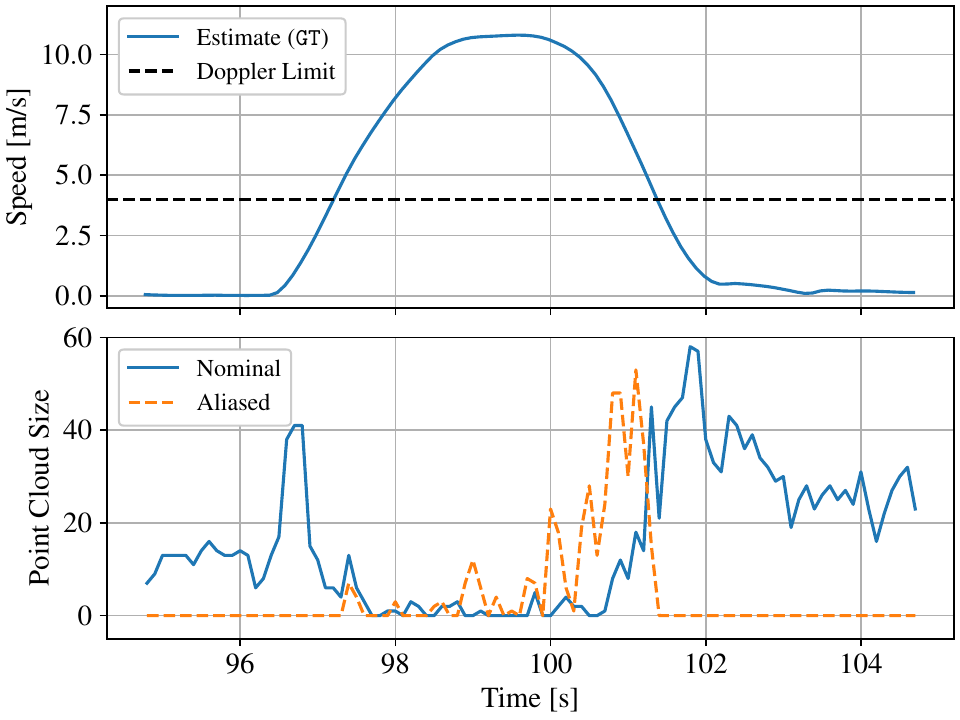}}
    \caption{Comparison between the count of nominal and aliased points compared to the platform speed in the (a) \experiment{flight\_6mps} and (b) \experiment{flight\_11mps} missions for the \config{noise} configuration. Note that despite exceeding the Doppler limit, the number of usable measurements does not decrease to zero. Note also the increased aliasing at higher vehicle speeds, where reducing the number of non-aliased points to zero for any long duration likely results in performance degradation.}
    \label{fig:evaluation:flight:doppler_limit:cloud_size}
\end{figure*}
\subsubsection{Exceeding the Doppler Limit}\label{sec:evaluation:flight:doppler_limit}
The question of how this performance correlates with exceeding the Doppler limit of the radar configuration remains largely unanswered. Some works (e.g.,~\cite{girod2024radar}) showcase experiments that exceed this limit, but without an organized study. As such, in the \mine{} environment, a set of experiments was conducted with the intent of studying the impact that exceeding the Doppler limit has on the resulting odometry performance. These experiments are referred to as \experiment{flight\_\#mps} where \# is replaced with the target speed. One of such trajectories can be seen in \cref{fig:evaluation:flight:trajectories:line}.

Increasing the Doppler limit on the radar side is feasible, as seen in \cref{sec:method:radar}, the limit is inversely proportional to the chirp duration. Thus, reducing the chirp duration results in a greater maximum Doppler. The challenge, however, is that by doing so, one worsens range and Doppler resolution, which can in turn be improved with higher chirp slopes and additional chirps per frame, each with their own associated costs. This limit and these tradeoffs are relevant because aerial robots are agile platforms with tight constraints on size, weight, and power, and many of the aforementioned changes would perturb the weight and power balance of such a system.

The actual impact of the Doppler limitation is that a region of the \ac{fov} will only return aliased measurements, resulting from violating the Nyquist theorem limit of the Doppler \ac{fft}. The aliased measurements will be treated as outliers by the M-estimator of the Doppler factor, as the estimator is not aware of the aliasing phenomenon. This means a region of the \ac{fov} will only return outliers for this regime of flight. Furthermore, similarly to the noise properties covered in \cref{sec:evaluation:noise:doppler_accuracy}, the shape and location of this region are a function of the radar-frame velocity, and thereby time-varying. This is visualized in \cref{fig:evaluation:flight:doppler_limit:fov} for radar-frame velocities along the indicated directions with \SI{11}{\meter\per\second} magnitude for the \rc{1} chirp configuration. Clearly, the impact this has on usable return measurements depends greatly on the direction of the radar-frame velocity. In \cref{fig:evaluation:flight:doppler_limit:fov:forward} the beam width \ac{fov} is almost entirely consumed, whereas in \cref{fig:evaluation:flight:doppler_limit:fov:upRight} there is still significant, albeit limited, space for measurements.

The experiments conducted to support this study progress from \SI{4}{\meter\per\second} (the Doppler limit of \rc{1}) to \SI{\sim 11}{\meter\per\second} at \SI{1}{\meter\per\second} intervals. The commanded trajectory (including position, velocity, and acceleration setpoints) is linear after manual re-positioning, as seen in \cref{fig:evaluation:flight:trajectories:line}, and derived from the same minimum-jerk optimization used by \cref{sec:evaluation:flight:doppler_accuracy}. The overall results can be seen in \cref{tab:evaluation:flight:doppler_limit}. In these experiments, both \doer{} and \asl{} can be seen performing decently; however, they still exhibit worse performance than the proposed method.

\begin{table}[h]
    \centering
    \caption{Position \ac{rmse} for the Experiments from \cref{sec:evaluation:flight:doppler_limit}}
    \label{tab:evaluation:flight:doppler_limit}
    \begin{tabular}{@{}lccc@{}}
\toprule
\multirow{2}{*}{Sequence}	&\multicolumn{3}{c}{APE $|$ RPE ($\Delta:\SI{10}{\meter}$) [\si{\meter}]}\\
	&{\texttt{noise}}	&{\texttt{x-RIO}~\cite{doer2021xrio}}	&{\texttt{ASL RIO}~\cite{girod2024radar}}\\
\midrule
\Mine{}	&	&	&\\
\texttt{flight\_4mps}	&\textbf{0.327} $|$ 0.251	&1.020 $|$ 0.366	&0.593 $|$ \textbf{0.200}\\
\texttt{flight\_5mps}	&\textbf{0.208} $|$ \textbf{0.174}	&0.558 $|$ 0.342	&0.401 $|$ 0.251\\
\texttt{flight\_6mps}	&\textbf{0.213} $|$ 0.246	&0.552 $|$ 0.366	&0.384 $|$ \textbf{0.194}\\
\texttt{flight\_7mps}	&\textbf{0.199} $|$ \textbf{0.172}	&0.562 $|$ 0.351	&0.355 $|$ 0.268\\
\texttt{flight\_8mps}	&\textbf{0.167} $|$ 0.182	&0.351 $|$ 0.327	&0.309 $|$ \textbf{0.179}\\
\texttt{flight\_9mps}	&\textbf{0.202} $|$ \textbf{0.106}	&0.364 $|$ 0.287	&0.269 $|$ 0.241\\
\texttt{flight\_10mps}	&\textbf{0.193} $|$ 0.194	&0.442 $|$ 0.266	&0.309 $|$ \textbf{0.159}\\
\texttt{flight\_11mps}	&\textbf{0.185} $|$ \textbf{0.270}	&0.456 $|$ 0.438	&0.330 $|$ 0.326\\
\bottomrule
\end{tabular}

\end{table}

From the results presented in \cref{tab:evaluation:flight:doppler_limit}, no clear trend emerges of decreasing estimation accuracy as total speed increases, despite almost tripling the maximum Doppler of the configuration. This, of course, cannot be a universal result, as to some extent, this will depend on uncontrollable variables such as the environment (as well as controllable variables like the positioning of the radar). Furthermore, it also relates to the duration of the high-speed event; different maximum speeds will correlate with different reductions in the number of usable points. As can be seen in \cref{fig:evaluation:flight:doppler_limit:flight_6mps,fig:evaluation:flight:doppler_limit:flight_11mps}, violating the Doppler limit does not necessarily cause the usable point count to go exactly to zero. 
However, the usable region of the \ac{fov} is negatively impacted by the increasing speed, thereby increasing the likelihood of outliers. Long durations with completely aliased point clouds will negatively impact estimator performance, however this is in part mitigated here by the fact that the platform does not instantaneously reach the offending speed, and the fact that even at \SI{\sim 11}{\meter\per\second}, as shown in \cref{fig:evaluation:flight:doppler_limit:flight_11mps}, there are still occasional points prevailing. However, if the expected flight regime contains significant durations with speeds significantly above the Doppler limit, the configuration likely should be changed accordingly.

\begin{table}[h]
    \centering
    \caption{Radar Point Cloud Measurement Aliasing}
    \label{tab:evaluation:flight:doppler_limit:cloud_size}
    \newcommand{\vmove}{-0.675ex}
\begin{tabular}{
	l
	S[table-format=3.0]
	S[table-format=2.1(2.1)]
	S[table-format=1.1(2.1)]
}
\toprule
\multirow{2}{*}[\vmove]{Sequence}	&{\multirow{2}{*}[\vmove]{Measurements}}	&\multicolumn{2}{c}{Point Counts}\\\cmidrule(l){3-4}
	&	&{Nominal}	&{Aliased}\\
\midrule
\Mine{}\\
\texttt{flight\_4mps}	&116	&21.4 +- 8.7	&0.1 +- 0.5\\
\texttt{flight\_5mps}	&98	&21.3 +- 9.7	&4.4 +- 5.1\\
\texttt{flight\_6mps}	&88	&18.3 +- 10.5	&8.7 +- 8.5\\
\texttt{flight\_7mps}	&85	&16.0 +- 12.8	&8.8 +- 9.3\\
\texttt{flight\_8mps}	&78	&13.1 +- 13.8	&6.6 +- 8.2\\
\texttt{flight\_9mps}	&76	&15.3 +- 16.9	&7.0 +- 9.5\\
\texttt{flight\_10mps}	&61	&11.9 +- 16.3	&5.8 +- 9.9\\
\texttt{flight\_11mps}	&60	&12.2 +- 16.6	&7.1 +- 13.0\\
\bottomrule
\end{tabular}

\end{table}

Statistics covering the aliasing across the trajectory tracking portion of the high-speed experiments from the \mine{} environment are shown in \cref{tab:evaluation:flight:doppler_limit:cloud_size}. The table presents the point cloud size mean and standard deviation, showing that for higher speeds, the total point cloud size decreases with significant aliasing starting already at \SI{5}{\meter\per\second}.

\subsubsection{Free Flight}\label{sec:evaluation:flight:free_flight}
In addition to position control, trajectories were also captured in free-flight, manually piloted experiments conducted in the \gym{}, \mine{}, and \forest{} environments. A subset of these trajectories can be seen in \cref{fig:evaluation:flight:trajectories:manual1,fig:evaluation:flight:trajectories:manual2,fig:evaluation:flight:trajectories:manual3,fig:evaluation:flight:trajectories:manual4,fig:evaluation:flight:trajectories:manual8}. These experiments feature less repetitive and less structured trajectories, with high speeds and dynamic maneuvers. As they are less repeatable, such trajectories utilize the \rc{1} configuration, since the higher return count has been shown to result in more robust estimation across different environments. The results for these experiments are presented in \cref{tab:evaluation:flight:free_flight} and demonstrate the superior performance of the proposed method.

The higher point return count is particularly important in the \gym{} environment, where in general returns are more sparse due to the emptiness and size of the environment. Regardless, the proposed method demonstrates superior performance with respect to the state-of-the-art, as shown in \cref{tab:evaluation:flight:free_flight}. Some experiments, in particular \experiment{manual\_2}, pose challenges to the other methods; this difficulty originates from an extreme lack of points when the platform flies toward the center of the \gym{} environment, as the opposing walls can be difficult for the radar to perceive. A similar behavior was observed in the experiments from \cref{sec:evaluation:flight:doppler_accuracy}.
The \mine{} showcases a different behavior; here, the environment produces relatively denser measurements, potentially resulting from the material properties and increased clutter present in the environment. Such rich features enable the proposed method to, again, outperform the state-of-the-art.

\begin{table}[h!]
    \centering
    \caption{Position \ac{rmse} for the Experiments from \cref{sec:evaluation:flight:free_flight}}
    \label{tab:evaluation:flight:free_flight}
    \begin{tabular}{@{}lccc@{}}
\toprule
\multirow{2}{*}{Sequence}	&\multicolumn{3}{c}{APE $|$ RPE ($\Delta:\SI{10}{\meter}$) [\si{\meter}]}\\
	&{\texttt{noise}}	&{\texttt{x-RIO}~\cite{doer2021xrio}}	&{\texttt{ASL RIO}~\cite{girod2024radar}}\\
\midrule
\Gym{}	&	&	&\\
\texttt{manual\_1}	&\textbf{0.745} $|$ 0.247	&0.992 $|$ 0.380	&0.807 $|$ \textbf{0.238}\\
\texttt{manual\_2}	&0.854 $|$ \textbf{0.248}	&1.484 $|$ 0.911	&\textbf{0.450} $|$ 0.336\\
\midrule
\Mine{}	&	&	&\\
\texttt{manual\_3}	&\textbf{0.458} $|$ \textbf{0.136}	&1.394 $|$ 0.323	&1.161 $|$ 0.224\\
\midrule
\Forest{}	&	&	&\\
\texttt{manual\_4}	&\textbf{2.223} $|$ \textbf{0.217}	&7.204 $|$ 0.403	&7.439 $|$ 0.374\\
\texttt{manual\_5}	&\textbf{2.086} $|$ \textbf{0.283}	&9.180 $|$ 0.379	&9.171 $|$ 0.342\\
\texttt{manual\_6}	&\textbf{1.128} $|$ \textbf{0.152}	&6.695 $|$ 0.354	&7.695 $|$ 0.339\\
\texttt{manual\_7}	&\textbf{0.644} $|$ \textbf{0.279}	&5.224 $|$ 0.379	&6.859 $|$ 0.391\\
\texttt{manual\_8}	&\textbf{4.458} $|$ 0.367	&9.758 $|$ 0.534	&8.288 $|$ \textbf{0.320}\\
\midrule
\Basement{}	&	&	&\\
\texttt{manual\_9}	&\textbf{1.029} $|$ \textbf{0.462}	&5.981 $|$ 0.721	&5.108 $|$ 0.772\\
\texttt{manual\_10}	&\textbf{1.041} $|$ \textbf{0.431}	&6.991 $|$ 0.835	&40.875 $|$ 3.325\\
\texttt{manual\_11}	&\textbf{1.260} $|$ \textbf{0.332}	&7.426 $|$ 1.922	&5.882 $|$ 0.783\\
\bottomrule
\end{tabular}

\end{table}

Similar phenomena are visible in the trajectories from the \forest{} environment, where regions of the forest contain more sparse vegetation, resulting in low-density or even empty point cloud measurements. As a result, the estimators can show jumps between pre- and post-measurement estimates. This is due to a lack of measurements, causing the estimator to drift further from the true value, and thus necessitating a larger correction step when the measurement density improves. Nevertheless, the proposed method still demonstrates high accuracy in such scenarios.

Manual flights in the \basement{} environment utilize a new radar with a different configuration (\rc{3}), mounted pointing almost completely downwards. This benefits the vertical position accuracy, resulting in reduced \ac{ape} while maintaining \ac{rpe} accuracy. However, yaw drift persists in trajectories such as \cref{fig:evaluation:flight:trajectories:manual11}. Other implicit cons of this sensor positioning are the increased number of robot reflections and the high directionality of reflections. The former results in greater difficulty for methods that lack sufficient outlier rejection (e.g., \asl{}). Whereas the latter poses challenges for methods that rely on the least-squares velocity estimate (e.g., \doer{}).

\subsubsection{Long-Range}\label{sec:evaluation:flight:long_range}
Radar sensors exist in a wide array of sensing capabilities with respect to maxima and resolution. So far, the focus has been pronounced on shorter-range sensors; however, the question of how the proposed methods fare with longer-range units is also of interest. For this purpose, a third radar sensor, mounted on the handheld unit, is included. From \cref{tab:evaluation:chirp}, the maximum range is significantly higher, however, at the cost of reduced range and Doppler resolution. Furthermore, the antenna distribution is such that azimuth resolution is improved at the cost of elevation resolution.

Handheld experiments were conducted with this sensor in the \corridor{} environment, the results from which can be seen in \cref{tab:evaluation:flight:long_range1,tab:evaluation:flight:long_range2}. The increased range and azimuth resolution enable the \config{geometry} configuration to achieve large performance improvements over both \config{base} and \config{noise}. This helps offset the vertical drift caused by the poor elevation resolution, which is responsible for the high \ac{ape} errors in \cref{tab:evaluation:flight:long_range1}. The fact that this environment-radar combination yields a high-performing \config{geometry} configuration can also result in improved attitude errors. However, the \config{base} and \config{noise} configurations already show very little error in these experiments, which can be seen for \experiment{handheld\_2} in \cref{fig:evaluation:flight:trajectories:handheld2}. As a result, the numerical gains are not so substantial.

Furthermore, the relatively low speed of such experiments (\SI{\sim1.2}{\meter\per\second} on average) means there is little improvement from the proposed noise model over the \config{base} configuration. This is logical given the simulated findings from \cref{sec:evaluation:noise:correlation}, of the magnifying effect speed can have on the \ac{aoa} noise contribution. In addition, following \cref{sec:evaluation:noise:doppler_accuracy}, configurations with large Doppler resolutions, such as \rc{4}, will see a similar effect, where the Doppler uncertainty dominates over the \ac{aoa} noise. Comparisons with the state-of-the-art show similar findings, however, still with the proposed method as a top performer.

\begin{table}[h]
    \centering
    \caption{Ablation of Method Configurations for Position \ac{rmse}}
    \label{tab:evaluation:flight:long_range1}
    \begin{tabular}{@{}lcc@{}}
\toprule
\multirow{2}{*}{Sequence}	&\multicolumn{2}{c}{\texttt{base} -- \texttt{noise} -- \texttt{geometry}}\\
	&{APE [\si{\meter}]}	&{RPE ($\Delta:\SI{10}{\meter}$) [\si{\meter}]}\\
\midrule
\Corridor{}	&	&\\
\texttt{handheld\_1}	&5.276 -- 5.223 -- \textbf{0.782}	&0.450 -- 0.447 -- \textbf{0.243}\\
\texttt{handheld\_2}	&4.971 -- 4.969 -- \textbf{1.698}	&0.479 -- 0.480 -- \textbf{0.338}\\
\bottomrule
\end{tabular}

\end{table}

\begin{table}[h]
    \centering
    \caption{Position \ac{rmse} for the Experiments from \cref{sec:evaluation:flight:long_range}}
    \label{tab:evaluation:flight:long_range2}
    \begin{tabular}{@{}lccc@{}}
\toprule
\multirow{2}{*}{Sequence}	&\multicolumn{3}{c}{APE $|$ RPE ($\Delta:\SI{10}{\meter}$) [\si{\meter}]}\\
	&{\texttt{noise}}	&{\texttt{x-RIO}~\cite{doer2021xrio}}	&{\texttt{ASL RIO}~\cite{girod2024radar}}\\
\midrule
\Corridor{}	&	&	&\\
\texttt{handheld\_1}	&5.223 $|$ 0.447	&\textbf{4.830} $|$ \textbf{0.423}	&5.400 $|$ 0.445\\
\texttt{handheld\_2}	&\textbf{4.969} $|$ \textbf{0.480}	&6.601 $|$ 0.593	&7.133 $|$ 0.650\\
\bottomrule
\end{tabular}

\end{table}

\subsubsection{Barometry}\label{sec:evaluation:flight:barometry}
The characteristics of the \forest{} environment enable barometer sensors to be used more easily, due to the lack of closed spaces and the absence of continuous disturbances to the aerodynamics. As such, the proposed method (\config{noise}) is augmented with the differential barometry factors described in \cref{sec:method:diff_baro}.

\begin{figure}[ht!]
    \centering
    \includegraphics[width=\linewidth]{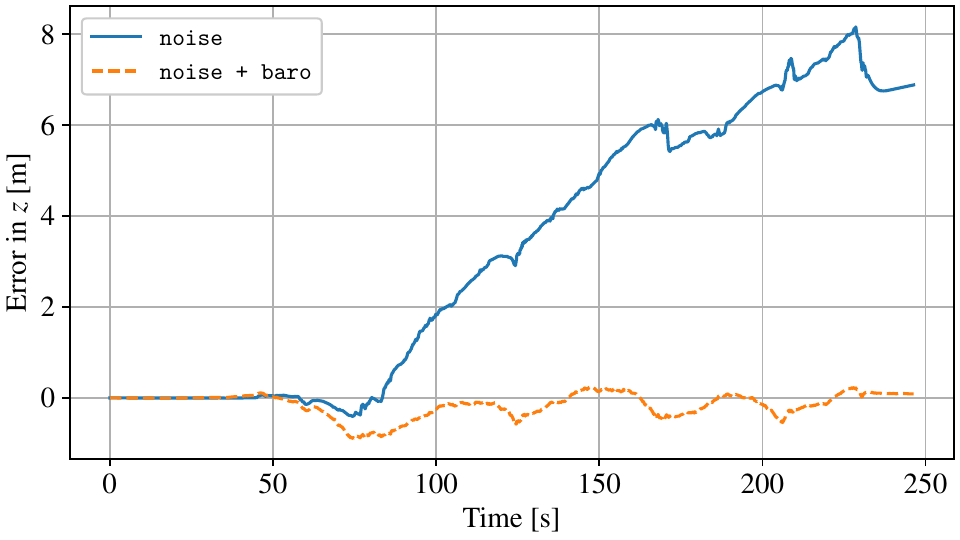}
    \spacingFigureCaption
    \caption{Vertical position estimate error for \experiment{manual\_8}, highlighting the positive impact of augmenting the \config{noise} configuration with barometer factors.}
    \label{fig:evaluation:flight:baro}
\end{figure}

An ablation of how this performs against the baseline and proposed methods can be seen in \cref{tab:evaluation:flight:baro}, where the augmentation clearly proves advantageous. The vertical estimation error from \experiment{manual\_8} is visualized in \cref{fig:evaluation:flight:baro}, where the vertical drift of \config{noise} can be seen, as well as the improved estimation performance of \config{noise + baro}.

\begin{table}[h!]
    \centering
    \caption{Ablation of Method Configurations for Position \ac{rmse}}
    \label{tab:evaluation:flight:baro}
    \begin{tabular}{@{}lcc@{}}
\toprule
\multirow{2}{*}{Sequence}	&\multicolumn{2}{c}{\texttt{base} -- \texttt{noise} -- \texttt{noise + baro}}\\
	&{APE [\si{\meter}]}	&{RPE ($\Delta:\SI{10}{\meter}$) [\si{\meter}]}\\
\midrule
\Forest{}	&	&\\
\texttt{manual\_4}	&2.194 -- 2.223 -- \textbf{1.081}	&\textbf{0.215} -- 0.217 -- 0.216\\
\texttt{manual\_5}	&2.072 -- 2.086 -- \textbf{1.059}	&0.283 -- 0.283 -- \textbf{0.267}\\
\texttt{manual\_6}	&\textbf{1.028} -- 1.128 -- 1.138	&0.150 -- 0.152 -- \textbf{0.147}\\
\texttt{manual\_7}	&0.618 -- 0.644 -- \textbf{0.491}	&0.277 -- 0.279 -- \textbf{0.259}\\
\texttt{manual\_8}	&4.534 -- 4.458 -- \textbf{0.863}	&0.359 -- 0.367 -- \textbf{0.335}\\
\bottomrule
\end{tabular}

\end{table}

\subsubsection{Real-time Feasibility}\label{sec:evaluation:flight:real_time}
\begin{figure}[h]
    \centering
    \includegraphics[width=\linewidth]{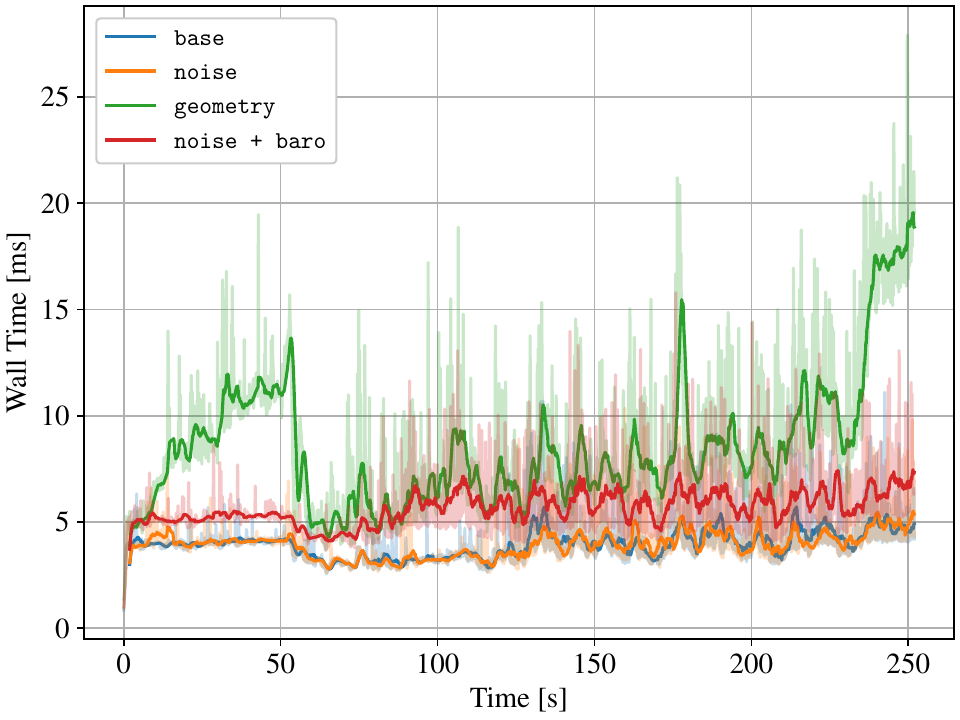}
    \spacingFigureCaption
    \caption{Raw and filtered durations for the radar callback from the \experiment{manual\_8} experiment. Note that, assuming a \SI{10}{\hertz} radar, the proposed method is within real-time bounds with a large margin, even in the \config{geometry} configuration, which includes expensive KD tree construction.}
    \label{fig:evaluation:flight:runtime}
\end{figure}

The aforementioned evaluations are conducted with an Intel Core i7-11850H processor, where the method demonstrates its performance also in terms of computational requirements. As shown in \cref{tab:evaluation:flight:runtime,fig:evaluation:flight:runtime}, the wall time of the radar point cloud callback, which handles all processing and publishing as described in \cref{sec:method:navigation}, is such that there is a large margin with respect to the radar update rate (typically \qtyrange{10}{30}{\hertz}). Furthermore, the impact of including the proposed noise model, which involves recalculation of the covariance upon relinearization, is marginal when comparing \config{noise} with \config{base}. The introduction of the barometer requires additional computation due to the added factors and an additional state to be estimated. Finally, including the distribution-distribution factor is seen to incur an increased, however still manageable, cost, resulting from the expensive KD tree construction and radius search.

\begin{table}[h]
    \centering
    \caption{Average Wall Time for Radar Callback}
    \label{tab:evaluation:flight:runtime}
    \begin{tabular}{
	l
	S[table-format=2.2]
	S[table-format=2.2]
	S[table-format=2.2]
	S[table-format=2.2]
}
\toprule
\multirow{2}{*}{Environment}	&\multicolumn{4}{c}{Average Wall Time [\si{\milli\second}]}\\
	&\texttt{base}	&\texttt{noise}	&\texttt{geometry}	&\texttt{noise + baro}\\
\midrule
\Corridor{}	&6.12	&6.70	&18.05	&{-}\\
\Gym{}	&5.85	&5.89	&12.72	&{-}\\
\Mine{}	&5.80	&5.64	&13.88	&{-}\\
\Forest{}	&4.10	&4.09	&10.19	&5.72\\
\Basement{}	&4.90	&5.18	&7.45	&{-}\\
\bottomrule
\end{tabular}

\end{table}

\subsection{Summary}\label{sec:evaluation:summary}
This section provides a brief summary of the core observations from the simulation studies presented in \cref{sec:evaluation:noise} and from the experimental studies presented in \cref{sec:evaluation:flight}. The proposed method was evaluated through extensive field experiments across five different environments, where it was found to provide robust and accurate estimates. Comparisons with other state-of-the-art radar-inertial odometry methods showcased its superior performance, especially in scenarios where the proposed method maintains accuracy while the other methods failed catastrophically.

Conditions in which \ac{aoa} measurement noise sources should have their greatest effect were identified and experimentally confirmed. Integrating the more complicated noise models to capture such phenomena resulted in improvements of up to \SI{47}{\percent} and \SI{14}{\percent} in \ac{ape} and \ac{rpe}, respectively. Such conditions relate clearly to how noise is injected into the factor residuals used by the estimator, as the effects are exaggerated by higher speeds. Furthermore, it was shown that the relative accuracy of the Doppler measurements themselves play a role in the importance of more advanced noise models. Unsurprisingly, considering more complicated-nonlinear phenomena may not yield such performance improvements if the simpler, linear noise contributions are dominating. Such circumstances can occur in radar configurations with high Doppler resolution or in trajectories with low speed. The former effect results from the Doppler noise dominating over other noise sources, following the results from \cref{sec:evaluation:noise:doppler_accuracy}. The latter behavior results in a damping of the \ac{aoa} noise contributions, following \cref{sec:evaluation:noise:correlation}.

The validity of \ac{fmcw} radar as an aiding sensor in circumstances exceeding its operating regime, especially with respect to Doppler, was also analyzed. This often-ignored shortcoming stems from the fundamental limitation imposed by the Nyquist theorem on the Fourier transform. We investigate such a scenario for Doppler and show how regions of the \ac{fov} can become invalid, subject to the magnitude and direction of the radar-frame velocity. Furthermore, we demonstrated the estimator's ability to maintain accuracy despite exceeding this limit through field flight experiments. However, depending on the speed, this is likely only a temporary effect, as the reduced point count over prolonged durations will eventually negatively impact the estimator's performance.

\section{CONCLUSION}\label{sec:conclusion}
This manuscript investigated the operational characteristics of typical measurements returned from an \ac{fmcw} radar, a sensing technology whose potential in robotics is of interest to the research community. Models were developed to capture the statistical properties of the range, Doppler, and \ac{aoa} measurement equations, and a factor graph-based estimator was developed to leverage these models. To validate our approach, simulation studies were conducted, grounded in the characteristics of the particular radar sensor in question. These simulations highlighted the significance of the hitherto largely unconsidered properties of uncertainty in the measurement relations, as well as analyzing behavior beyond the limits of the sensor's standard operating range. Furthermore, experiments were conducted specifically to examine the effects previously studied in simulation, as well as to extensively evaluate the overall performance of the proposed method, both in laboratory and field environments. Comparisons against relevant state-of-the-art methods demonstrated the effectiveness of the proposed method, as it outperforms the state-of-the-art.

\section{APPENDIX}

\subsection{Notations}
    \vspace{-11ex}
    \glsaddall
    \printglossary[title={}]

\subsection{Additional Results}
Comprehensive experimental results across different method configurations with state-of-the-art comparisons can be found in \cref{tab:additional_results:ape:trans,tab:additional_results:rpe:trans,tab:additional_results:ape:angle,tab:additional_results:rpe:angle}, where the proposed method can be seen consistently outperforming the state-of-the-art w.r.t. position error. Note that in some environments, the \config{geometry} configuration can lead to significant performance improvements; however, this is not always feasible. The unpredictable nature of the attitude performance can also be seen here. This behavior is dominated by the pervasive yaw drift, which results from a lack of observability. Finally, note also that, despite the state-of-the-art methods showing comparable performance in some experiments, they also occasionally demonstrate catastrophic failures, which the proposed method does not.

\begin{table*}[t!]
    \centering
    \caption{\ac{ape} Performance (Position \ac{rmse} $\pm$ Std. Dev.) Across Experiments}
    \label{tab:additional_results:ape:trans}
    \begin{tabular}{@{}lccccc@{}}
\toprule
\multirow{2}{*}{Sequence}	&\multicolumn{5}{c}{APE [\si{\meter}]}\\
	&{\texttt{base}}	&{\texttt{noise}}	&{\texttt{geometry}}	&{\texttt{x-RIO}~\cite{doer2021xrio}}	&{\texttt{ASL RIO}~\cite{girod2024radar}}\\
\midrule
\Corridor{}	&	&	&	&	&\\
\texttt{hel\_3mps\_align}	&0.544 $\pm$ 0.356	&0.529 $\pm$ 0.340	&\textbf{0.515 $\pm$ 0.320}	&1.111 $\pm$ 0.723	&1.171 $\pm$ 0.781\\
\texttt{hel\_4mps\_align}	&0.416 $\pm$ 0.218	&0.389 $\pm$ 0.201	&\textbf{0.084 $\pm$ 0.033}	&0.179 $\pm$ 0.083	&0.246 $\pm$ 0.132\\
\texttt{rec\_3mps\_align}	&3.031 $\pm$ 2.070	&\textbf{2.112 $\pm$ 1.325}	&2.122 $\pm$ 1.133	&6.613 $\pm$ 4.477	&6.351 $\pm$ 4.262\\
\texttt{rec\_3mps\_const}	&2.446 $\pm$ 1.706	&2.342 $\pm$ 1.627	&3.342 $\pm$ 2.181	&\textbf{2.338 $\pm$ 1.543}	&6.690 $\pm$ 4.296\\
\texttt{rec\_4mps\_align}	&1.239 $\pm$ 0.914	&1.147 $\pm$ 0.844	&\textbf{0.504 $\pm$ 0.273}	&1.024 $\pm$ 0.657	&0.831 $\pm$ 0.505\\
\texttt{rec\_4mps\_const}	&1.334 $\pm$ 0.904	&1.238 $\pm$ 0.835	&\textbf{0.447 $\pm$ 0.301}	&0.649 $\pm$ 0.398	&1.818 $\pm$ 1.215\\
\texttt{handheld\_1}	&5.276 $\pm$ 3.005	&5.223 $\pm$ 2.975	&\textbf{0.782 $\pm$ 0.499}	&4.830 $\pm$ 2.997	&5.400 $\pm$ 3.249\\
\texttt{handheld\_2}	&4.971 $\pm$ 2.813	&4.969 $\pm$ 2.811	&\textbf{1.698 $\pm$ 0.840}	&6.601 $\pm$ 3.929	&7.133 $\pm$ 4.141\\
\midrule
\Gym{}	&	&	&	&	&\\
\texttt{lem\_3mps\_align}	&1.671 $\pm$ 1.169	&\textbf{0.885 $\pm$ 0.626}	&1.752 $\pm$ 1.019	&22.918 $\pm$ 18.341	&30.365 $\pm$ 22.223\\
\texttt{lem\_3mps\_const}	&0.654 $\pm$ 0.478	&0.621 $\pm$ 0.445	&\textbf{0.349 $\pm$ 0.192}	&1.058 $\pm$ 0.652	&1.884 $\pm$ 1.296\\
\texttt{lem\_4mps\_align}	&\textbf{0.915 $\pm$ 0.616}	&0.955 $\pm$ 0.634	&1.246 $\pm$ 0.831	&4.281 $\pm$ 3.201	&2.569 $\pm$ 1.680\\
\texttt{lem\_4mps\_const}	&0.958 $\pm$ 0.687	&0.952 $\pm$ 0.683	&\textbf{0.607 $\pm$ 0.350}	&1.520 $\pm$ 0.938	&1.466 $\pm$ 0.942\\
\texttt{manual\_1}	&\textbf{0.460 $\pm$ 0.302}	&0.745 $\pm$ 0.509	&1.081 $\pm$ 0.635	&0.992 $\pm$ 0.631	&0.807 $\pm$ 0.496\\
\texttt{manual\_2}	&0.564 $\pm$ 0.442	&0.854 $\pm$ 0.701	&0.842 $\pm$ 0.641	&1.484 $\pm$ 1.222	&\textbf{0.450 $\pm$ 0.362}\\
\midrule
\Mine{}	&	&	&	&	&\\
\texttt{flight\_4mps}	&\textbf{0.324 $\pm$ 0.212}	&0.327 $\pm$ 0.214	&0.325 $\pm$ 0.175	&1.020 $\pm$ 0.757	&0.593 $\pm$ 0.423\\
\texttt{flight\_5mps}	&\textbf{0.205 $\pm$ 0.127}	&0.208 $\pm$ 0.129	&0.247 $\pm$ 0.172	&0.558 $\pm$ 0.397	&0.401 $\pm$ 0.293\\
\texttt{flight\_6mps}	&\textbf{0.207 $\pm$ 0.143}	&0.213 $\pm$ 0.147	&0.300 $\pm$ 0.199	&0.552 $\pm$ 0.401	&0.384 $\pm$ 0.281\\
\texttt{flight\_7mps}	&\textbf{0.197 $\pm$ 0.125}	&0.199 $\pm$ 0.126	&0.294 $\pm$ 0.170	&0.562 $\pm$ 0.392	&0.355 $\pm$ 0.229\\
\texttt{flight\_8mps}	&0.182 $\pm$ 0.124	&\textbf{0.167 $\pm$ 0.116}	&0.205 $\pm$ 0.100	&0.351 $\pm$ 0.259	&0.309 $\pm$ 0.219\\
\texttt{flight\_9mps}	&0.198 $\pm$ 0.130	&0.202 $\pm$ 0.133	&\textbf{0.159 $\pm$ 0.093}	&0.364 $\pm$ 0.244	&0.269 $\pm$ 0.181\\
\texttt{flight\_10mps}	&0.197 $\pm$ 0.139	&\textbf{0.193 $\pm$ 0.137}	&0.294 $\pm$ 0.134	&0.442 $\pm$ 0.320	&0.309 $\pm$ 0.227\\
\texttt{flight\_11mps}	&0.189 $\pm$ 0.133	&\textbf{0.185 $\pm$ 0.131}	&0.321 $\pm$ 0.217	&0.456 $\pm$ 0.338	&0.330 $\pm$ 0.247\\
\texttt{manual\_3}	&\textbf{0.436 $\pm$ 0.227}	&0.458 $\pm$ 0.239	&0.639 $\pm$ 0.318	&1.394 $\pm$ 0.796	&1.161 $\pm$ 0.649\\
\midrule
\Forest{}	&	&	&	&	&\\
\texttt{manual\_4}	&2.194 $\pm$ 1.309	&2.223 $\pm$ 1.319	&\textbf{1.800 $\pm$ 0.966}	&7.204 $\pm$ 3.999	&7.439 $\pm$ 4.439\\
\texttt{manual\_5}	&2.072 $\pm$ 1.024	&2.086 $\pm$ 1.017	&\textbf{1.156 $\pm$ 0.555}	&9.180 $\pm$ 4.870	&9.171 $\pm$ 5.206\\
\texttt{manual\_6}	&\textbf{1.028 $\pm$ 0.661}	&1.128 $\pm$ 0.709	&4.727 $\pm$ 2.903	&6.695 $\pm$ 3.867	&7.695 $\pm$ 4.864\\
\texttt{manual\_7}	&0.618 $\pm$ 0.249	&0.644 $\pm$ 0.268	&\textbf{0.365 $\pm$ 0.163}	&5.224 $\pm$ 2.575	&6.859 $\pm$ 3.849\\
\texttt{manual\_8}	&4.534 $\pm$ 2.949	&4.458 $\pm$ 2.891	&\textbf{2.819 $\pm$ 1.584}	&9.758 $\pm$ 5.931	&8.288 $\pm$ 5.382\\
\midrule
\Basement{}	&	&	&	&	&\\
\texttt{manual\_9}	&1.185 $\pm$ 0.731	&\textbf{1.029 $\pm$ 0.577}	&2.806 $\pm$ 1.560	&5.981 $\pm$ 3.543	&5.108 $\pm$ 2.765\\
\texttt{manual\_10}	&1.869 $\pm$ 1.085	&\textbf{1.041 $\pm$ 0.593}	&1.160 $\pm$ 0.523	&6.991 $\pm$ 4.135	&40.875 $\pm$ 25.169\\
\texttt{manual\_11}	&1.398 $\pm$ 0.860	&\textbf{1.260 $\pm$ 0.699}	&1.977 $\pm$ 1.088	&7.426 $\pm$ 4.740	&5.882 $\pm$ 3.719\\
\bottomrule
\end{tabular}

\end{table*}

\begin{table*}[t!]
    \centering
    \caption{\ac{rpe} Performance (Position \ac{rmse} $\pm$ Std. Dev.) Across Experiments}
    \label{tab:additional_results:rpe:trans}
    \begin{tabular}{@{}lccccc@{}}
\toprule
\multirow{2}{*}{Sequence}	&\multicolumn{5}{c}{RPE ($\Delta:\SI{10}{\meter}$) [\si{\meter}]}\\
	&{\texttt{base}}	&{\texttt{noise}}	&{\texttt{geometry}}	&{\texttt{x-RIO}~\cite{doer2021xrio}}	&{\texttt{ASL RIO}~\cite{girod2024radar}}\\
\midrule
\Corridor{}	&	&	&	&	&\\
\texttt{hel\_3mps\_align}	&0.221 $\pm$ 0.101	&0.201 $\pm$ 0.079	&\textbf{0.178 $\pm$ 0.055}	&0.314 $\pm$ 0.120	&0.267 $\pm$ 0.105\\
\texttt{hel\_4mps\_align}	&0.153 $\pm$ 0.076	&0.152 $\pm$ 0.080	&\textbf{0.068 $\pm$ 0.027}	&0.147 $\pm$ 0.061	&0.123 $\pm$ 0.045\\
\texttt{rec\_3mps\_align}	&0.292 $\pm$ 0.114	&0.257 $\pm$ 0.112	&\textbf{0.253 $\pm$ 0.100}	&0.565 $\pm$ 0.370	&1.995 $\pm$ 1.571\\
\texttt{rec\_3mps\_const}	&0.275 $\pm$ 0.120	&\textbf{0.266 $\pm$ 0.124}	&0.288 $\pm$ 0.141	&0.450 $\pm$ 0.227	&2.135 $\pm$ 1.938\\
\texttt{rec\_4mps\_align}	&0.181 $\pm$ 0.095	&0.180 $\pm$ 0.096	&\textbf{0.175 $\pm$ 0.086}	&0.229 $\pm$ 0.102	&0.264 $\pm$ 0.142\\
\texttt{rec\_4mps\_const}	&0.218 $\pm$ 0.086	&\textbf{0.215 $\pm$ 0.084}	&0.264 $\pm$ 0.116	&0.264 $\pm$ 0.095	&0.315 $\pm$ 0.152\\
\texttt{handheld\_1}	&0.450 $\pm$ 0.195	&0.447 $\pm$ 0.194	&\textbf{0.243 $\pm$ 0.115}	&0.423 $\pm$ 0.159	&0.445 $\pm$ 0.161\\
\texttt{handheld\_2}	&0.479 $\pm$ 0.269	&0.480 $\pm$ 0.270	&\textbf{0.338 $\pm$ 0.140}	&0.593 $\pm$ 0.223	&0.650 $\pm$ 0.274\\
\midrule
\Gym{}	&	&	&	&	&\\
\texttt{lem\_3mps\_align}	&0.397 $\pm$ 0.256	&\textbf{0.339 $\pm$ 0.211}	&0.355 $\pm$ 0.180	&3.104 $\pm$ 2.619	&5.408 $\pm$ 3.945\\
\texttt{lem\_3mps\_const}	&0.184 $\pm$ 0.070	&\textbf{0.178 $\pm$ 0.064}	&0.190 $\pm$ 0.068	&0.464 $\pm$ 0.348	&0.203 $\pm$ 0.095\\
\texttt{lem\_4mps\_align}	&0.264 $\pm$ 0.143	&\textbf{0.262 $\pm$ 0.147}	&0.278 $\pm$ 0.173	&0.452 $\pm$ 0.331	&0.315 $\pm$ 0.174\\
\texttt{lem\_4mps\_const}	&0.309 $\pm$ 0.114	&\textbf{0.308 $\pm$ 0.114}	&0.309 $\pm$ 0.118	&0.371 $\pm$ 0.153	&0.324 $\pm$ 0.158\\
\texttt{manual\_1}	&0.250 $\pm$ 0.141	&0.247 $\pm$ 0.140	&\textbf{0.206 $\pm$ 0.097}	&0.380 $\pm$ 0.165	&0.238 $\pm$ 0.074\\
\texttt{manual\_2}	&0.248 $\pm$ 0.103	&0.248 $\pm$ 0.104	&\textbf{0.177 $\pm$ 0.087}	&0.911 $\pm$ 0.332	&0.336 $\pm$ 0.127\\
\midrule
\Mine{}	&	&	&	&	&\\
\texttt{flight\_4mps}	&0.254 $\pm$ 0.062	&0.251 $\pm$ 0.061	&0.280 $\pm$ 0.088	&0.366 $\pm$ 0.087	&\textbf{0.200 $\pm$ 0.072}\\
\texttt{flight\_5mps}	&0.179 $\pm$ 0.060	&0.174 $\pm$ 0.057	&\textbf{0.170 $\pm$ 0.044}	&0.342 $\pm$ 0.072	&0.251 $\pm$ 0.062\\
\texttt{flight\_6mps}	&0.245 $\pm$ 0.097	&0.246 $\pm$ 0.101	&0.267 $\pm$ 0.077	&0.366 $\pm$ 0.094	&\textbf{0.194 $\pm$ 0.051}\\
\texttt{flight\_7mps}	&0.175 $\pm$ 0.063	&\textbf{0.172 $\pm$ 0.062}	&0.246 $\pm$ 0.076	&0.351 $\pm$ 0.113	&0.268 $\pm$ 0.101\\
\texttt{flight\_8mps}	&0.184 $\pm$ 0.096	&0.182 $\pm$ 0.095	&\textbf{0.177 $\pm$ 0.077}	&0.327 $\pm$ 0.119	&0.179 $\pm$ 0.061\\
\texttt{flight\_9mps}	&0.107 $\pm$ 0.055	&\textbf{0.106 $\pm$ 0.054}	&0.127 $\pm$ 0.044	&0.287 $\pm$ 0.103	&0.241 $\pm$ 0.091\\
\texttt{flight\_10mps}	&0.196 $\pm$ 0.080	&0.194 $\pm$ 0.081	&0.225 $\pm$ 0.081	&0.266 $\pm$ 0.093	&\textbf{0.159 $\pm$ 0.053}\\
\texttt{flight\_11mps}	&0.271 $\pm$ 0.087	&\textbf{0.270 $\pm$ 0.088}	&0.326 $\pm$ 0.104	&0.438 $\pm$ 0.109	&0.326 $\pm$ 0.156\\
\texttt{manual\_3}	&\textbf{0.136 $\pm$ 0.064}	&0.136 $\pm$ 0.064	&0.293 $\pm$ 0.136	&0.323 $\pm$ 0.146	&0.224 $\pm$ 0.090\\
\midrule
\Forest{}	&	&	&	&	&\\
\texttt{manual\_4}	&0.215 $\pm$ 0.094	&0.217 $\pm$ 0.095	&\textbf{0.206 $\pm$ 0.089}	&0.403 $\pm$ 0.133	&0.374 $\pm$ 0.136\\
\texttt{manual\_5}	&0.283 $\pm$ 0.115	&0.283 $\pm$ 0.115	&\textbf{0.281 $\pm$ 0.122}	&0.379 $\pm$ 0.118	&0.342 $\pm$ 0.106\\
\texttt{manual\_6}	&\textbf{0.150 $\pm$ 0.065}	&0.152 $\pm$ 0.063	&0.216 $\pm$ 0.055	&0.354 $\pm$ 0.111	&0.339 $\pm$ 0.102\\
\texttt{manual\_7}	&0.277 $\pm$ 0.111	&0.279 $\pm$ 0.111	&\textbf{0.268 $\pm$ 0.115}	&0.379 $\pm$ 0.141	&0.391 $\pm$ 0.137\\
\texttt{manual\_8}	&0.359 $\pm$ 0.184	&0.367 $\pm$ 0.191	&\textbf{0.312 $\pm$ 0.178}	&0.534 $\pm$ 0.271	&0.320 $\pm$ 0.162\\
\midrule
\Basement{}	&	&	&	&	&\\
\texttt{manual\_9}	&0.488 $\pm$ 0.197	&\textbf{0.462 $\pm$ 0.179}	&0.486 $\pm$ 0.147	&0.721 $\pm$ 0.196	&0.772 $\pm$ 0.301\\
\texttt{manual\_10}	&0.469 $\pm$ 0.192	&0.431 $\pm$ 0.176	&\textbf{0.340 $\pm$ 0.128}	&0.835 $\pm$ 0.360	&3.325 $\pm$ 1.575\\
\texttt{manual\_11}	&0.355 $\pm$ 0.164	&\textbf{0.332 $\pm$ 0.143}	&0.432 $\pm$ 0.204	&1.922 $\pm$ 1.595	&0.783 $\pm$ 0.472\\
\bottomrule
\end{tabular}

\end{table*}

\begin{table*}[t!]
    \centering
    \caption{\ac{ape} Performance (Attitude \ac{rmse} $\pm$ Std. Dev.) Across Experiments}
    \label{tab:additional_results:ape:angle}
    \begin{tabular}{@{}lccccc@{}}
\toprule
\multirow{2}{*}{Sequence}	&\multicolumn{5}{c}{APE [\si{\degree}]}\\
	&{\texttt{base}}	&{\texttt{noise}}	&{\texttt{geometry}}	&{\texttt{x-RIO}~\cite{doer2021xrio}}	&{\texttt{ASL RIO}~\cite{girod2024radar}}\\
\midrule
\Corridor{}	&	&	&	&	&\\
\texttt{hel\_3mps\_align}	&1.075 $\pm$ 0.810	&1.075 $\pm$ 0.810	&0.965 $\pm$ 0.690	&0.694 $\pm$ 0.317	&\textbf{0.517 $\pm$ 0.200}\\
\texttt{hel\_4mps\_align}	&0.258 $\pm$ 0.094	&\textbf{0.257 $\pm$ 0.094}	&0.502 $\pm$ 0.278	&1.738 $\pm$ 0.840	&1.685 $\pm$ 0.943\\
\texttt{rec\_3mps\_align}	&2.133 $\pm$ 1.347	&2.207 $\pm$ 1.405	&\textbf{1.072 $\pm$ 0.532}	&1.556 $\pm$ 0.867	&1.944 $\pm$ 1.233\\
\texttt{rec\_3mps\_const}	&2.046 $\pm$ 1.228	&2.134 $\pm$ 1.299	&\textbf{0.663 $\pm$ 0.283}	&0.828 $\pm$ 0.364	&1.460 $\pm$ 0.823\\
\texttt{rec\_4mps\_align}	&2.611 $\pm$ 1.673	&2.665 $\pm$ 1.721	&\textbf{1.073 $\pm$ 0.475}	&1.279 $\pm$ 0.771	&1.419 $\pm$ 0.781\\
\texttt{rec\_4mps\_const}	&1.221 $\pm$ 0.731	&1.229 $\pm$ 0.740	&\textbf{0.663 $\pm$ 0.394}	&1.299 $\pm$ 0.476	&1.517 $\pm$ 0.936\\
\texttt{handheld\_1}	&0.888 $\pm$ 0.434	&\textbf{0.887 $\pm$ 0.434}	&0.942 $\pm$ 0.467	&1.190 $\pm$ 0.741	&1.174 $\pm$ 0.662\\
\texttt{handheld\_2}	&1.404 $\pm$ 0.642	&1.403 $\pm$ 0.642	&\textbf{1.086 $\pm$ 0.448}	&1.390 $\pm$ 0.766	&1.640 $\pm$ 0.819\\
\midrule
\Gym{}	&	&	&	&	&\\
\texttt{lem\_3mps\_align}	&3.697 $\pm$ 2.540	&3.733 $\pm$ 2.565	&3.469 $\pm$ 2.137	&3.462 $\pm$ 2.408	&\textbf{2.674 $\pm$ 1.763}\\
\texttt{lem\_3mps\_const}	&1.842 $\pm$ 1.152	&1.821 $\pm$ 1.134	&\textbf{0.660 $\pm$ 0.265}	&1.444 $\pm$ 0.698	&1.445 $\pm$ 0.795\\
\texttt{lem\_4mps\_align}	&3.281 $\pm$ 2.224	&3.302 $\pm$ 2.241	&5.691 $\pm$ 3.816	&\textbf{1.842 $\pm$ 1.200}	&2.171 $\pm$ 1.432\\
\texttt{lem\_4mps\_const}	&3.181 $\pm$ 2.059	&3.212 $\pm$ 2.083	&\textbf{0.879 $\pm$ 0.435}	&2.739 $\pm$ 1.732	&3.415 $\pm$ 2.260\\
\texttt{manual\_1}	&1.188 $\pm$ 0.825	&1.178 $\pm$ 0.818	&\textbf{0.905 $\pm$ 0.673}	&1.070 $\pm$ 0.652	&0.956 $\pm$ 0.603\\
\texttt{manual\_2}	&2.287 $\pm$ 1.446	&2.302 $\pm$ 1.457	&\textbf{1.852 $\pm$ 1.174}	&2.079 $\pm$ 1.116	&2.678 $\pm$ 1.449\\
\midrule
\Mine{}	&	&	&	&	&\\
\texttt{flight\_4mps}	&0.871 $\pm$ 0.653	&0.871 $\pm$ 0.653	&0.862 $\pm$ 0.654	&1.397 $\pm$ 0.741	&\textbf{0.728 $\pm$ 0.392}\\
\texttt{flight\_5mps}	&0.463 $\pm$ 0.241	&0.462 $\pm$ 0.240	&\textbf{0.355 $\pm$ 0.176}	&1.202 $\pm$ 0.631	&0.943 $\pm$ 0.504\\
\texttt{flight\_6mps}	&0.751 $\pm$ 0.595	&\textbf{0.750 $\pm$ 0.594}	&0.775 $\pm$ 0.602	&1.758 $\pm$ 1.067	&1.165 $\pm$ 0.603\\
\texttt{flight\_7mps}	&0.656 $\pm$ 0.520	&0.656 $\pm$ 0.520	&\textbf{0.627 $\pm$ 0.532}	&1.870 $\pm$ 1.096	&1.302 $\pm$ 0.758\\
\texttt{flight\_8mps}	&0.648 $\pm$ 0.379	&0.649 $\pm$ 0.379	&\textbf{0.510 $\pm$ 0.326}	&1.039 $\pm$ 0.441	&0.884 $\pm$ 0.420\\
\texttt{flight\_9mps}	&0.235 $\pm$ 0.150	&\textbf{0.233 $\pm$ 0.149}	&0.329 $\pm$ 0.188	&0.688 $\pm$ 0.386	&0.634 $\pm$ 0.423\\
\texttt{flight\_10mps}	&0.420 $\pm$ 0.257	&0.420 $\pm$ 0.257	&\textbf{0.396 $\pm$ 0.251}	&1.276 $\pm$ 0.672	&0.624 $\pm$ 0.272\\
\texttt{flight\_11mps}	&0.833 $\pm$ 0.505	&0.833 $\pm$ 0.506	&\textbf{0.776 $\pm$ 0.476}	&1.550 $\pm$ 0.846	&1.178 $\pm$ 0.654\\
\texttt{manual\_3}	&0.425 $\pm$ 0.203	&0.425 $\pm$ 0.203	&\textbf{0.314 $\pm$ 0.176}	&1.954 $\pm$ 0.962	&1.819 $\pm$ 0.874\\
\midrule
\Forest{}	&	&	&	&	&\\
\texttt{manual\_4}	&4.190 $\pm$ 2.649	&4.192 $\pm$ 2.651	&4.714 $\pm$ 2.982	&\textbf{2.497 $\pm$ 1.392}	&3.782 $\pm$ 2.209\\
\texttt{manual\_5}	&4.717 $\pm$ 2.758	&4.721 $\pm$ 2.761	&4.913 $\pm$ 2.912	&\textbf{3.453 $\pm$ 2.146}	&3.574 $\pm$ 2.253\\
\texttt{manual\_6}	&4.299 $\pm$ 2.735	&4.307 $\pm$ 2.740	&4.298 $\pm$ 2.631	&\textbf{1.629 $\pm$ 0.973}	&3.426 $\pm$ 2.146\\
\texttt{manual\_7}	&2.620 $\pm$ 1.370	&2.622 $\pm$ 1.371	&\textbf{2.356 $\pm$ 1.258}	&2.902 $\pm$ 1.529	&2.996 $\pm$ 1.684\\
\texttt{manual\_8}	&2.284 $\pm$ 1.309	&2.283 $\pm$ 1.308	&1.890 $\pm$ 1.096	&1.494 $\pm$ 0.756	&\textbf{1.168 $\pm$ 0.597}\\
\midrule
\Basement{}	&	&	&	&	&\\
\texttt{manual\_9}	&1.908 $\pm$ 1.098	&1.910 $\pm$ 1.100	&1.633 $\pm$ 0.927	&1.250 $\pm$ 0.634	&\textbf{0.868 $\pm$ 0.346}\\
\texttt{manual\_10}	&2.119 $\pm$ 1.270	&2.131 $\pm$ 1.279	&1.799 $\pm$ 1.031	&\textbf{1.742 $\pm$ 0.884}	&1.964 $\pm$ 1.131\\
\texttt{manual\_11}	&2.526 $\pm$ 1.677	&2.524 $\pm$ 1.675	&\textbf{1.294 $\pm$ 0.848}	&2.312 $\pm$ 1.723	&2.324 $\pm$ 1.493\\
\bottomrule
\end{tabular}

\end{table*}

\begin{table*}[t!]
    \centering
    \caption{\ac{rpe} Performance (Attitude \ac{rmse} $\pm$ Std. Dev.) Across Experiments}
    \label{tab:additional_results:rpe:angle}
    \begin{tabular}{@{}lccccc@{}}
\toprule
\multirow{2}{*}{Sequence}	&\multicolumn{5}{c}{RPE ($\Delta:\SI{10}{\meter}$) [\si{\degree}]}\\
	&{\texttt{base}}	&{\texttt{noise}}	&{\texttt{geometry}}	&{\texttt{x-RIO}~\cite{doer2021xrio}}	&{\texttt{ASL RIO}~\cite{girod2024radar}}\\
\midrule
\Corridor{}	&	&	&	&	&\\
\texttt{hel\_3mps\_align}	&1.532 $\pm$ 0.940	&1.533 $\pm$ 0.940	&1.470 $\pm$ 0.903	&0.725 $\pm$ 0.330	&\textbf{0.352 $\pm$ 0.116}\\
\texttt{hel\_4mps\_align}	&0.379 $\pm$ 0.172	&0.378 $\pm$ 0.172	&\textbf{0.330 $\pm$ 0.115}	&1.282 $\pm$ 0.581	&1.581 $\pm$ 0.777\\
\texttt{rec\_3mps\_align}	&\textbf{0.953 $\pm$ 0.709}	&0.955 $\pm$ 0.711	&0.959 $\pm$ 0.708	&1.246 $\pm$ 0.868	&0.975 $\pm$ 0.723\\
\texttt{rec\_3mps\_const}	&0.501 $\pm$ 0.255	&0.503 $\pm$ 0.256	&0.493 $\pm$ 0.273	&0.501 $\pm$ 0.263	&\textbf{0.347 $\pm$ 0.197}\\
\texttt{rec\_4mps\_align}	&0.435 $\pm$ 0.213	&0.436 $\pm$ 0.213	&\textbf{0.401 $\pm$ 0.235}	&1.208 $\pm$ 0.830	&1.198 $\pm$ 0.881\\
\texttt{rec\_4mps\_const}	&0.648 $\pm$ 0.450	&\textbf{0.644 $\pm$ 0.448}	&0.712 $\pm$ 0.493	&0.878 $\pm$ 0.566	&0.738 $\pm$ 0.477\\
\texttt{handheld\_1}	&0.860 $\pm$ 0.394	&0.861 $\pm$ 0.395	&\textbf{0.818 $\pm$ 0.363}	&1.802 $\pm$ 0.787	&1.747 $\pm$ 0.745\\
\texttt{handheld\_2}	&0.535 $\pm$ 0.303	&0.535 $\pm$ 0.303	&\textbf{0.478 $\pm$ 0.257}	&1.441 $\pm$ 0.664	&1.451 $\pm$ 0.693\\
\midrule
\Gym{}	&	&	&	&	&\\
\texttt{lem\_3mps\_align}	&1.029 $\pm$ 0.719	&1.026 $\pm$ 0.716	&1.056 $\pm$ 0.754	&1.201 $\pm$ 0.695	&\textbf{0.962 $\pm$ 0.571}\\
\texttt{lem\_3mps\_const}	&0.505 $\pm$ 0.292	&0.502 $\pm$ 0.294	&\textbf{0.483 $\pm$ 0.289}	&0.844 $\pm$ 0.522	&0.559 $\pm$ 0.377\\
\texttt{lem\_4mps\_align}	&1.306 $\pm$ 1.007	&1.307 $\pm$ 1.008	&1.433 $\pm$ 1.075	&0.929 $\pm$ 0.461	&\textbf{0.618 $\pm$ 0.332}\\
\texttt{lem\_4mps\_const}	&0.701 $\pm$ 0.423	&0.701 $\pm$ 0.423	&0.646 $\pm$ 0.402	&0.891 $\pm$ 0.542	&\textbf{0.547 $\pm$ 0.309}\\
\texttt{manual\_1}	&1.111 $\pm$ 0.758	&1.111 $\pm$ 0.757	&\textbf{1.102 $\pm$ 0.711}	&1.299 $\pm$ 0.694	&1.106 $\pm$ 0.680\\
\texttt{manual\_2}	&1.543 $\pm$ 0.671	&1.541 $\pm$ 0.670	&\textbf{1.348 $\pm$ 0.670}	&2.224 $\pm$ 1.062	&1.814 $\pm$ 0.877\\
\midrule
\Mine{}	&	&	&	&	&\\
\texttt{flight\_4mps}	&1.231 $\pm$ 0.787	&1.248 $\pm$ 0.791	&1.225 $\pm$ 0.772	&1.162 $\pm$ 0.563	&\textbf{0.865 $\pm$ 0.484}\\
\texttt{flight\_5mps}	&0.260 $\pm$ 0.113	&0.260 $\pm$ 0.113	&\textbf{0.243 $\pm$ 0.106}	&0.857 $\pm$ 0.357	&0.802 $\pm$ 0.405\\
\texttt{flight\_6mps}	&1.086 $\pm$ 0.607	&1.086 $\pm$ 0.608	&1.548 $\pm$ 0.784	&0.762 $\pm$ 0.373	&\textbf{0.544 $\pm$ 0.298}\\
\texttt{flight\_7mps}	&1.730 $\pm$ 1.035	&1.734 $\pm$ 1.035	&\textbf{1.650 $\pm$ 0.998}	&2.548 $\pm$ 1.464	&1.903 $\pm$ 1.086\\
\texttt{flight\_8mps}	&0.709 $\pm$ 0.323	&0.713 $\pm$ 0.326	&0.730 $\pm$ 0.322	&0.572 $\pm$ 0.236	&\textbf{0.427 $\pm$ 0.193}\\
\texttt{flight\_9mps}	&0.339 $\pm$ 0.152	&0.339 $\pm$ 0.151	&\textbf{0.333 $\pm$ 0.146}	&0.749 $\pm$ 0.352	&0.823 $\pm$ 0.439\\
\texttt{flight\_10mps}	&0.510 $\pm$ 0.259	&0.507 $\pm$ 0.256	&0.516 $\pm$ 0.247	&0.549 $\pm$ 0.238	&\textbf{0.435 $\pm$ 0.198}\\
\texttt{flight\_11mps}	&1.161 $\pm$ 0.538	&1.156 $\pm$ 0.531	&\textbf{1.046 $\pm$ 0.518}	&1.169 $\pm$ 0.597	&1.235 $\pm$ 0.618\\
\texttt{manual\_3}	&\textbf{0.277 $\pm$ 0.142}	&0.277 $\pm$ 0.141	&0.283 $\pm$ 0.137	&0.711 $\pm$ 0.366	&0.771 $\pm$ 0.382\\
\midrule
\Forest{}	&	&	&	&	&\\
\texttt{manual\_4}	&1.915 $\pm$ 1.059	&1.915 $\pm$ 1.060	&\textbf{1.909 $\pm$ 1.097}	&3.002 $\pm$ 1.679	&3.413 $\pm$ 1.839\\
\texttt{manual\_5}	&3.274 $\pm$ 1.702	&3.273 $\pm$ 1.701	&3.278 $\pm$ 1.703	&2.085 $\pm$ 1.160	&\textbf{1.641 $\pm$ 0.956}\\
\texttt{manual\_6}	&1.285 $\pm$ 0.664	&\textbf{1.284 $\pm$ 0.663}	&1.305 $\pm$ 0.649	&2.554 $\pm$ 1.222	&2.864 $\pm$ 1.407\\
\texttt{manual\_7}	&3.643 $\pm$ 1.871	&\textbf{3.642 $\pm$ 1.872}	&3.646 $\pm$ 1.849	&4.144 $\pm$ 2.199	&3.779 $\pm$ 1.980\\
\texttt{manual\_8}	&2.088 $\pm$ 1.143	&2.089 $\pm$ 1.145	&2.100 $\pm$ 1.121	&1.392 $\pm$ 0.697	&\textbf{1.158 $\pm$ 0.605}\\
\midrule
\Basement{}	&	&	&	&	&\\
\texttt{manual\_9}	&1.038 $\pm$ 0.635	&1.034 $\pm$ 0.632	&1.032 $\pm$ 0.614	&0.811 $\pm$ 0.358	&\textbf{0.432 $\pm$ 0.204}\\
\texttt{manual\_10}	&0.967 $\pm$ 0.629	&0.962 $\pm$ 0.631	&\textbf{0.952 $\pm$ 0.624}	&1.493 $\pm$ 0.823	&1.263 $\pm$ 0.790\\
\texttt{manual\_11}	&0.324 $\pm$ 0.174	&\textbf{0.323 $\pm$ 0.168}	&0.325 $\pm$ 0.139	&1.147 $\pm$ 0.594	&1.026 $\pm$ 0.606\\
\bottomrule
\end{tabular}

\end{table*}

\section*{ACKNOWLEDGMENT}
The authors thank Mohit Singh and Philipp Weiss for their help in conducting the experiments.

\bibliographystyle{IEEEtran}
\bibliography{bib/cleaned}

% Generated by IEEEtran.bst, version: 1.14 (2015/08/26)
\begin{thebibliography}{10}
\providecommand{\url}[1]{#1}
\csname url@samestyle\endcsname
\providecommand{\newblock}{\relax}
\providecommand{\bibinfo}[2]{#2}
\providecommand{\BIBentrySTDinterwordspacing}{\spaceskip=0pt\relax}
\providecommand{\BIBentryALTinterwordstretchfactor}{4}
\providecommand{\BIBentryALTinterwordspacing}{\spaceskip=\fontdimen2\font plus
\BIBentryALTinterwordstretchfactor\fontdimen3\font minus \fontdimen4\font\relax}
\providecommand{\BIBforeignlanguage}[2]{{%
\expandafter\ifx\csname l@#1\endcsname\relax
\typeout{** WARNING: IEEEtran.bst: No hyphenation pattern has been}%
\typeout{** loaded for the language `#1'. Using the pattern for}%
\typeout{** the default language instead.}%
\else
\language=\csname l@#1\endcsname
\fi
#2}}
\providecommand{\BIBdecl}{\relax}
\BIBdecl

\bibitem{cadena2016future}
C.~Cadena, L.~Carlone, H.~Carrillo, Y.~Latif, D.~Scaramuzza, J.~Neira, I.~Reid, and J.~J. Leonard, ``Past, present, and future of simultaneous localization and mapping: Toward the robust-perception age,'' \emph{IEEE Transactions on Robotics}, vol.~32, no.~6, pp. 1309--1332, Dec. 2016.

\bibitem{ebadi2024present}
K.~Ebadi, L.~Bernreiter, H.~Biggie, G.~Catt, Y.~Chang, A.~Chatterjee, C.~E. Denniston, S.-P. Desch{\^{e}}nes, K.~Harlow, S.~Khattak, L.~Nogueira, M.~Palieri, P.~Petr{\'{a}}{\v{c}}ek, M.~Petrl{\'{i}}k, A.~Reinke, V.~Kr{\'{a}}tk{\'{y}}, S.~Zhao, A.-a. Agha-mohammadi, K.~Alexis, C.~Heckman, K.~Khosoussi, N.~Kottege, B.~Morrell, M.~Hutter, F.~Pauling, F.~Pomerleau, M.~Saska, S.~Scherer, R.~Siegwart, J.~L. Williams, and L.~Carlone, ``Present and future of {SLAM} in extreme environments: The {DARPA} {SubT} challenge,'' \emph{IEEE Transactions on Robotics}, vol.~40, pp. 936--959, 2024.

\bibitem{nissov2024degradation}
M.~Nissov, N.~Khedekar, and K.~Alexis, ``Degradation resilient lidar-radar-inertial odometry,'' in \emph{2024 IEEE International Conference on Robotics and Automation (ICRA)}.\hskip 1em plus 0.5em minus 0.4em\relax IEEE, May 2024, pp. 8587--8594.

\bibitem{hatleskog2024Probabilistic}
J.~Hatleskog and K.~Alexis, ``Probabilistic degeneracy detection for point-to-plane error minimization,'' \emph{IEEE Robotics and Automation Letters}, vol.~9, no.~12, pp. 11\,234--11\,241, Dec. 2024.

\bibitem{nissov2024robust}
M.~Nissov, J.~A. Edlund, P.~Spieler, C.~Padgett, K.~Alexis, and S.~Khattak, ``Robust high-speed state estimation for off-road navigation using radar velocity factors,'' \emph{IEEE Robotics and Automation Letters}, vol.~9, no.~12, pp. 11\,146--11\,153, Dec. 2024.

\bibitem{tuna2025Informed}
T.~Tuna, J.~Nubert, P.~Pfreundschuh, C.~Cadena, S.~Khattak, and M.~Hutter, ``Informed, constrained, aligned: A field analysis on degeneracy-aware point cloud registration in the wild,'' \emph{IEEE Transactions on Field Robotics}, vol.~2, pp. 485--515, 2025.

\bibitem{khattak2020Thermal}
S.~Khattak, C.~Papachristos, and K.~Alexis, ``Keyframe-based thermal-inertial odometry,'' \emph{Journal of Field Robotics}, vol.~37, no.~4, pp. 552--579, Dec. 2020.

\bibitem{tranzatto2022Cerberus}
M.~Tranzatto, T.~Miki, M.~Dharmadhikari, L.~Bernreiter, M.~Kulkarni, F.~Mascarich, O.~Andersson, S.~Khattak, M.~Hutter, R.~Siegwart, and K.~Alexis, ``{CERBERUS} in the {DARPA} subterranean challenge,'' \emph{Science Robotics}, vol.~7, no.~66, p. eabp9742, May 2022.

\bibitem{harlow2024new}
K.~Harlow, H.~Jang, T.~D. Barfoot, A.~Kim, and C.~Heckman, ``A new wave in robotics: Survey on recent mmwave radar applications in robotics,'' \emph{IEEE Transactions on Robotics}, vol.~40, pp. 4544--4560, 2024.

\bibitem{noh2025garlio}
C.~Noh, W.~Yang, M.~Jung, S.~Jung, and A.~Kim, ``{GaRLIO}: Gravity enhanced radar-lidar-inertial odometry,'' in \emph{2025 IEEE International Conference on Robotics and Automation (ICRA)}.\hskip 1em plus 0.5em minus 0.4em\relax IEEE, May 2025, pp. 9869--9875.

\bibitem{thormann2023bearing}
K.~Thormann and M.~Baum, ``Single-frame radar odometry incorporating bearing uncertainty,'' in \emph{2023 IEEE Symposium Sensor Data Fusion and International Conference on Multisensor Fusion and Integration (SDF-MFI)}.\hskip 1em plus 0.5em minus 0.4em\relax IEEE, Nov. 2023, pp. 1--7.

\bibitem{xu2025radarPointUncertainty}
Y.~Xu, Q.~Huang, S.~Shen, and H.~Yin, ``Incorporating point uncertainty in radar {SLAM},'' \emph{IEEE Robotics and Automation Letters}, vol.~10, no.~3, pp. 2168--2175, Mar. 2025.

\bibitem{zhu2025Robustradar}
J.~Zhu, J.~Hu, X.~Zhao, X.~Lang, Y.~Mao, and G.~Huang, ``Robust {4D} radar-aided inertial navigation for aerial vehicles,'' in \emph{2025 IEEE International Conference on Robotics and Automation (ICRA)}.\hskip 1em plus 0.5em minus 0.4em\relax IEEE, May 2025, pp. 9848--9854.

\bibitem{lisus2025spinningDoppler}
D.~Lisus, K.~Burnett, D.~J. Yoon, R.~Poulton, J.~Marshall, and T.~D. Barfoot, ``Are {D}oppler velocity measurements useful for spinning radar odometry?'' \emph{IEEE Robotics and Automation Letters}, vol.~10, no.~1, pp. 224--231, Jan. 2025.

\bibitem{cen2018Precise}
S.~H. Cen and P.~Newman, ``Precise ego-motion estimation with millimeter-wave radar under diverse and challenging conditions,'' in \emph{2018 IEEE International Conference on Robotics and Automation (ICRA)}.\hskip 1em plus 0.5em minus 0.4em\relax IEEE, May 2018, pp. 6045--6052.

\bibitem{cen2019Matching}
------, ``Radar-only ego-motion estimation in difficult settings via graph matching,'' in \emph{2019 IEEE International Conference on Robotics and Automation (ICRA)}.\hskip 1em plus 0.5em minus 0.4em\relax IEEE, May 2019, pp. 298--304.

\bibitem{lim2023orora}
H.~Lim, K.~Han, G.~Shin, G.~Kim, S.~Hong, and H.~Myung, ``{ORORA}: Outlier-robust radar odometry,'' in \emph{2023 IEEE International Conference on Robotics and Automation (ICRA)}.\hskip 1em plus 0.5em minus 0.4em\relax IEEE, May 2023, pp. 2046--2053.

\bibitem{burnett2025RadarLidar}
K.~Burnett, A.~P. Schoellig, and T.~D. Barfoot, ``Continuous-time radar-inertial and lidar-inertial odometry using a gaussian process motion prior,'' \emph{IEEE Transactions on Robotics}, vol.~41, pp. 1059--1076, 2025.

\bibitem{adolfsson2023radar}
D.~Adolfsson, M.~Magnusson, A.~Alhashimi, A.~J. Lilienthal, and H.~Andreasson, ``Lidar-level localization with radar? the {CFEAR} approach to accurate, fast, and robust large-scale radar odometry in diverse environments,'' \emph{IEEE Transactions on Robotics}, vol.~39, no.~2, pp. 1476--1495, Apr. 2023.

\bibitem{kellner2013instantaneous}
D.~Kellner, M.~Barjenbruch, J.~Klappstein, J.~Dickmann, and K.~Dietmayer, ``Instantaneous ego-motion estimation using {D}oppler radar,'' in \emph{16th International IEEE Conference on Intelligent Transportation Systems (ITSC 2013)}.\hskip 1em plus 0.5em minus 0.4em\relax IEEE, Oct. 2013, pp. 869--874.

\bibitem{kellner2014dual}
------, ``Instantaneous full-motion estimation of arbitrary objects using dual {D}oppler radar,'' in \emph{2014 IEEE Intelligent Vehicles Symposium Proceedings}.\hskip 1em plus 0.5em minus 0.4em\relax IEEE, Jun. 2014, pp. 324--329.

\bibitem{doer2020ekf}
C.~Doer and G.~F. Trommer, ``An {EKF} based approach to radar inertial odometry,'' in \emph{2020 IEEE International Conference on Multisensor Fusion and Integration for Intelligent Systems (MFI)}.\hskip 1em plus 0.5em minus 0.4em\relax IEEE, Sep. 2020, pp. 152--159.

\bibitem{doer2021yaw}
------, ``Yaw aided radar inertial odometry using manhattan world assumptions,'' in \emph{2021 28th Saint Petersburg International Conference on Integrated Navigation Systems (ICINS)}.\hskip 1em plus 0.5em minus 0.4em\relax IEEE, May 2021, pp. 1--9.

\bibitem{doer2021xrio}
------, ``{x-RIO}: Radar inertial odometry with multiple radar sensors and yaw aiding,'' \emph{Gyroscopy and Navigation}, vol.~12, no.~4, pp. 329--339, Dec. 2021.

\bibitem{zhuang2023iriom}
Y.~Zhuang, B.~Wang, J.~Huai, and M.~Li, ``4{D} {iRIOM}: 4{D} imaging radar inertial odometry and mapping,'' \emph{IEEE Robotics and Automation Letters}, vol.~8, no.~6, pp. 3246--3253, Jun. 2023.

\bibitem{kim2018scancontext}
G.~Kim and A.~Kim, ``Scan context: Egocentric spatial descriptor for place recognition within 3d point cloud map,'' in \emph{2018 IEEE/RSJ International Conference on Intelligent Robots and Systems (IROS)}, 2018, pp. 4802--4809.

\bibitem{li2023preint}
X.~Li, H.~Zhang, and W.~Chen, ``{4D} radar-based pose graph {SLAM} with ego-velocity pre-integration factor,'' \emph{IEEE Robotics and Automation Letters}, vol.~8, no.~8, pp. 5124--5131, Aug. 2023.

\bibitem{herraez2025rai}
D.~C. Herraez, M.~Zeller, D.~Wang, J.~Behley, M.~Heidingsfeld, and C.~Stachniss, ``Rai-slam: Radar-inertial slam for autonomous vehicles,'' \emph{IEEE Robotics and Automation Letters}, vol.~10, no.~6, pp. 5257--5264, 2025.

\bibitem{chen2023drio}
H.~Chen, Y.~Liu, and Y.~Cheng, ``{DRIO}: Robust radar-inertial odometry in dynamic environments,'' \emph{IEEE Robotics and Automation Letters}, vol.~8, no.~9, pp. 5918--5925, Sep. 2023.

\bibitem{wu2024efear}
X.~Wu, Y.~Chen, Z.~Li, Z.~Hong, and L.~Hu, ``{EFEAR-4D}: Ego-velocity filtering for efficient and accurate {4D} radar odometry,'' \emph{IEEE Robotics and Automation Letters}, vol.~9, no.~11, pp. 9828--9835, Nov. 2024.

\bibitem{kramer2024suburban}
A.~J. Kramer and C.~Heckman, ``Radar-based localization for autonomous ground vehicles in suburban neighborhoods,'' \emph{IEEE Transactions on Field Robotics}, vol.~1, pp. 161--169, 2024.

\bibitem{michalczyk2023tight}
J.~Michalczyk, R.~Jung, C.~Brommer, and S.~Weiss, ``Multi-state tightly-coupled {EKF}-based radar-inertial odometry with persistent landmarks,'' in \emph{2023 IEEE International Conference on Robotics and Automation (ICRA)}.\hskip 1em plus 0.5em minus 0.4em\relax IEEE, May 2023, pp. 4011--4017.

\bibitem{kramer2020visually}
A.~Kramer, C.~Stahoviak, A.~Santamaria-Navarro, A.-a. Agha-mohammadi, and C.~Heckman, ``Radar-inertial ego-velocity estimation for visually degraded environments,'' in \emph{2020 IEEE International Conference on Robotics and Automation (ICRA)}.\hskip 1em plus 0.5em minus 0.4em\relax IEEE, May 2020, pp. 5739--5746.

\bibitem{michalczyk2022tight}
J.~Michalczyk, R.~Jung, and S.~Weiss, ``Tightly-coupled {EKF}-based radar-inertial odometry,'' in \emph{2022 IEEE/RSJ International Conference on Intelligent Robots and Systems (IROS)}.\hskip 1em plus 0.5em minus 0.4em\relax IEEE, Oct. 2022, pp. 12\,336--12\,343.

\bibitem{mourikis2007MSCKF}
A.~I. Mourikis and S.~I. Roumeliotis, ``A multi-state constraint kalman filter for vision-aided inertial navigation,'' in \emph{Proceedings 2007 IEEE International Conference on Robotics and Automation}.\hskip 1em plus 0.5em minus 0.4em\relax IEEE, Apr. 2007, pp. 3565--3572.

\bibitem{girod2024radar}
R.~Girod, M.~Hauswirth, P.~Pfreundschuh, M.~Biasio, and R.~Siegwart, ``A robust baro-radar-inertial odometry m-estimator for multicopter navigation in cities and forests,'' in \emph{2024 IEEE International Conference on Multisensor Fusion and Integration for Intelligent Systems (MFI)}.\hskip 1em plus 0.5em minus 0.4em\relax IEEE, Sep. 2024, pp. 1--8.

\bibitem{wang2025dopplerslam}
D.~Wang, H.~Haag, D.~C. Herraez, S.~May, C.~Stachniss, and A.~N{\"{u}}chter, ``{D}oppler-{SLAM}: {D}oppler-aided radar-inertial and lidar-inertial simultaneous localization and mapping,'' \emph{IEEE Robotics and Automation Letters}, vol.~10, no.~9, pp. 9438--9445, Sep. 2025.

\bibitem{kubelka2024need}
V.~Kubelka, E.~Fritz, and M.~Magnusson, ``Do we need scan-matching in radar odometry?'' in \emph{2024 IEEE International Conference on Robotics and Automation (ICRA)}.\hskip 1em plus 0.5em minus 0.4em\relax IEEE, May 2024, pp. 13\,710--13\,716.

\bibitem{xiang2025vgc}
J.~Xiang, X.~He, Z.~Chen, L.~Zhang, X.~Luo, and J.~Mao, ``Vgc-rio: A tightly integrated radar-inertial odometry with spatial weighted doppler velocity and local geometric constrained rcs histograms,'' \emph{IEEE Robotics and Automation Letters}, vol.~10, no.~11, pp. 11\,642--11\,649, 2025.

\bibitem{bordonaro2012unbiased}
S.~V. Bordonaro, P.~Willett, and Y.~Bar-Shalom, ``Unbiased tracking with converted measurements,'' in \emph{2012 IEEE Radar Conference}.\hskip 1em plus 0.5em minus 0.4em\relax IEEE, May 2012, pp. 0741--0745.

\bibitem{wang2024riv}
D.~Wang, S.~May, and A.~Nuechter, ``{RIV-SLAM}: Radar-inertial-velocity optimization based graph {SLAM},'' in \emph{2024 IEEE 20th International Conference on Automation Science and Engineering (CASE)}.\hskip 1em plus 0.5em minus 0.4em\relax IEEE, Aug. 2024, pp. 774--781.

\bibitem{chan2025noise}
P.~H. Chan, S.~Shahbeigi~Roudposhti, X.~Ye, and V.~Donzella, ``A noise analysis of {4D} {RADAR}: Robust sensing for automotive?'' \emph{IEEE Sensors Journal}, vol.~25, no.~10, pp. 18\,291--18\,301, May 2025.

\bibitem{richards2013Fundamentals}
M.~A. Richards, \emph{Fundamentals of Radar Signal Processing}, 2nd~ed., ser. McGraw-Hill's AccessEngineering.\hskip 1em plus 0.5em minus 0.4em\relax Chicago, Ill.: McGraw-Hill Education LLC., 2014, "Second edition"-- Cover.

\bibitem{iovescu2020fundamentals}
C.~Iovescu and S.~Rao, ``The fundamentals of millimeter wave radar sensors,'' Texas Instruments Application Note, 2020.

\bibitem{veen1988beamforming}
B.~Van~Veen and K.~Buckley, ``Beamforming: a versatile approach to spatial filtering,'' \emph{IEEE ASSP Magazine}, vol.~5, no.~2, pp. 4--24, Apr. 1988.

\bibitem{capon1969Adaptive}
J.~Capon, ``High-resolution frequency-wavenumber spectrum analysis,'' \emph{Proceedings of the IEEE}, vol.~57, no.~8, pp. 1408--1418, 1969.

\bibitem{rao2017tiMIMO}
S.~Rao, ``{MIMO} radar,'' Texas Instruments, Application Note SWRA554A, May 2017.

\bibitem{parviainen2008barometry}
J.~Parviainen, J.~Kantola, and J.~Collin, ``Differential barometry in personal navigation,'' in \emph{2008 IEEE/ION Position, Location and Navigation Symposium}.\hskip 1em plus 0.5em minus 0.4em\relax IEEE, 2008, pp. 148--152.

\bibitem{gtsam}
\BIBentryALTinterwordspacing
F.~Dellaert and {GTSAM Contributors}, ``borglab/gtsam,'' May 2022. [Online]. Available: \url{https://github.com/borglab/gtsam}
\BIBentrySTDinterwordspacing

\bibitem{kaess2011isam2}
M.~Kaess, H.~Johannsson, R.~Roberts, V.~Ila, J.~Leonard, and F.~Dellaert, ``{iSAM2}: Incremental smoothing and mapping with fluid relinearization and incremental variable reordering,'' in \emph{2011 IEEE International Conference on Robotics and Automation}.\hskip 1em plus 0.5em minus 0.4em\relax IEEE, May 2011, pp. 3281--3288.

\bibitem{forster2017manifold}
C.~Forster, L.~Carlone, F.~Dellaert, and D.~Scaramuzza, ``On-manifold preintegration for real-time visual-inertial odometry,'' \emph{IEEE Transactions on Robotics}, vol.~33, no.~1, pp. 1--21, Feb. 2017.

\bibitem{huber2004robust}
P.~J. Huber, \emph{Robust Statistics}, ser. Wiley Series in Probability and Statistics - Applied Probability and Statistics Section Series.\hskip 1em plus 0.5em minus 0.4em\relax New York, NY [u.a.]: Wiley, 2004.

\bibitem{biber2003NDT}
P.~Biber and W.~Strasser, ``The normal distributions transform: a new approach to laser scan matching,'' in \emph{2003 IEEE/RSJ International Conference on Intelligent Robots and Systems (IROS)}, vol.~3.\hskip 1em plus 0.5em minus 0.4em\relax IEEE, 2003, pp. 2743--2748.

\bibitem{zhang20234DRadarSlam}
J.~Zhang, H.~Zhuge, Z.~Wu, G.~Peng, M.~Wen, Y.~Liu, and D.~Wang, ``{4DRadarSLAM}: A {4D} imaging radar slam system for large-scale environments based on pose graph optimization,'' in \emph{2023 IEEE International Conference on Robotics and Automation (ICRA)}.\hskip 1em plus 0.5em minus 0.4em\relax IEEE, May 2023, pp. 8333--8340.

\bibitem{blanco2014nanoflann}
J.~L. Blanco and P.~K. Rai, ``nanoflann: a {C}++ header-only fork of {FLANN}, a library for nearest neighbor ({NN}) with kd-trees,'' \url{https://github.com/jlblancoc/nanoflann}, 2014.

\bibitem{nissov2024roamer}
M.~Nissov, S.~Khattak, J.~A. Edlund, C.~Padgett, K.~Alexis, and P.~Spieler, ``{ROAMER}: Robust offroad autonomy using multimodal state estimation with radar velocity integration,'' in \emph{2024 IEEE Aerospace Conference}.\hskip 1em plus 0.5em minus 0.4em\relax IEEE, Mar. 2024, pp. 1--10.

\bibitem{nasa1976atmosphere}
{NASA}, {NOAA}, and {USAF}, ``{U.S. Standard Atmosphere, 1976},'' {National Aeronautics and Space Administration}, Tech. Rep. NASA-TM-X-74335, 1976, {T}echnical {M}emorandum.

\bibitem{nissov2025Sync}
M.~Nissov, N.~Khedekar, and K.~Alexis, ``Simultaneous triggering and synchronization of sensors and onboard computers,'' 2025.

\bibitem{meier2015px4}
L.~Meier, D.~Honegger, and M.~Pollefeys, ``{PX4}: A node-based multithreaded open source robotics framework for deeply embedded platforms,'' in \emph{2015 IEEE International Conference on Robotics and Automation (ICRA)}.\hskip 1em plus 0.5em minus 0.4em\relax IEEE, May 2015, pp. 6235--6240.

\bibitem{px4}
\BIBentryALTinterwordspacing
{PX4 Development Team}, ``{PX4} autopilot,'' Open-source software, 2024. [Online]. Available: \url{https://px4.io}
\BIBentrySTDinterwordspacing

\bibitem{buchanan2021Allan}
\BIBentryALTinterwordspacing
R.~Buchanan, ``{Allan Variance ROS},'' Nov. 2021. [Online]. Available: \url{https://github.com/ori-drs/allan_variance_ros}
\BIBentrySTDinterwordspacing

\bibitem{wang2021Trajectories}
Z.~Wang, H.~Ye, C.~Xu, and F.~Gao, ``Generating large-scale trajectories efficiently using double descriptions of polynomials,'' in \emph{2021 IEEE International Conference on Robotics and Automation (ICRA)}, 2021, pp. 7436--7442.

\bibitem{sturm2012benchmark}
J.~Sturm, N.~Engelhard, F.~Endres, W.~Burgard, and D.~Cremers, ``A benchmark for the evaluation of {{RGB-D SLAM}} systems,'' in \emph{2012 {{IEEE}}/{{RSJ International Conference}} on {{Intelligent Robots}} and {{Systems}}}.\hskip 1em plus 0.5em minus 0.4em\relax IEEE, Oct. 2012, pp. 573--580.

\bibitem{grupp2017evo}
M.~Grupp, ``evo: Python package for the evaluation of odometry and slam.'' \url{https://github.com/MichaelGrupp/evo}, 2017.

\end{thebibliography}

\vfill

\end{document}